\pdftrailerid{}
\pdfmapfile{+cm.map}
\pdfmapfile{+cmextra.map}
\pdfmapfile{+latxfont.map}
\pdfmapfile{+euler.map}
\pdfmapfile{+symbols.map}
\documentclass{article}
\makeatletter
\RequirePackage{fancyhdr}
\RequirePackage{natbib}

\setcitestyle{authoryear,round,citesep={;},aysep={,},yysep={;}}

\def\addcontentsline#1#2#3{}

\def\maketitle{\par
\begingroup
   \def\thefootnote{\fnsymbol{footnote}}
   \def\@makefnmark{\hbox to 0pt{$^{\@thefnmark}$\hss}} 
   \long\def\@makefntext##1{\parindent 1em\noindent
                            \hbox to1.8em{\hss $\m@th ^{\@thefnmark}$}##1}
   \@maketitle \@thanks
\endgroup
\setcounter{footnote}{0}
\let\maketitle\relax \let\@maketitle\relax
\gdef\@thanks{}\gdef\@author{}\gdef\@title{}\let\thanks\relax}

\def\@maketitle{\vbox{\hsize\textwidth
{\LARGE\sc \@title\par}
    \def\And{\end{tabular}\hfil\linebreak[0]\hfil
            \begin{tabular}[t]{l}\bf\rule{\z@}{24pt}\ignorespaces}%
  \def\AND{\end{tabular}\hfil\linebreak[4]\hfil
            \begin{tabular}[t]{l}\bf\rule{\z@}{24pt}\ignorespaces}%
    \begin{tabular}[t]{l}\bf\rule{\z@}{24pt}\@author\end{tabular}%
\vskip 0.3in minus 0.1in}}

\renewenvironment{abstract}{\vskip.075in\centerline{\large\sc
Abstract}\vspace{0.5ex}\begin{quote}}{\par\end{quote}\vskip 1ex}

\def\section{\@startsection {section}{1}{\z@}{-2.0ex plus
    -0.5ex minus -.2ex}{1.5ex plus 0.3ex
minus0.2ex}{\large\sc\raggedright}}

\def\subsection{\@startsection{subsection}{2}{\z@}{-1.8ex plus
-0.5ex minus -.2ex}{0.8ex plus .2ex}{\normalsize\sc\raggedright}}
\def\subsubsection{\@startsection{subsubsection}{3}{\z@}{-1.5ex
plus      -0.5ex minus -.2ex}{0.5ex plus
.2ex}{\normalsize\sc\raggedright}}
\def\paragraph{\@startsection{paragraph}{4}{\z@}{1.5ex plus
0.5ex minus .2ex}{-1em}{\normalsize\bf}}
\def\subparagraph{\@startsection{subparagraph}{5}{\z@}{1.5ex plus
  0.5ex minus .2ex}{-1em}{\normalsize\sc}}

\skip\footins 9pt plus 4pt minus 2pt
\def\footnoterule{\kern-3pt \hrule width 12pc \kern 2.6pt }
\leftmargini\leftmargin \leftmarginii 2em
\def\@listi{\leftmargin\leftmargini}
\def\@listii{\leftmargin\leftmarginii
   \labelwidth\leftmarginii\advance\labelwidth-\labelsep
   \topsep 2pt plus 1pt minus 0.5pt
   \parsep 1pt plus 0.5pt minus 0.5pt
   \itemsep \parsep}
\def\@listiii{\leftmargin\leftmarginiii
    \labelwidth\leftmarginiii\advance\labelwidth-\labelsep
    \topsep 1pt plus 0.5pt minus 0.5pt
    \parsep \z@ \partopsep 0.5pt plus 0pt minus 0.5pt
    \itemsep \topsep}
\def\@listiv{\leftmargin\leftmarginiv
     \labelwidth\leftmarginiv\advance\labelwidth-\labelsep}
\def\@listv{\leftmargin\leftmarginv
     \labelwidth\leftmarginv\advance\labelwidth-\labelsep}
\def\@listvi{\leftmargin\leftmarginvi
     \labelwidth\leftmarginvi\advance\labelwidth-\labelsep}

\belowdisplayskip \abovedisplayskip
\def\normalsize{\@setsize\normalsize{11pt}\xpt\@xpt}
\def\small{\@setsize\small{10pt}\ixpt\@ixpt}
\def\footnotesize{\@setsize\footnotesize{10pt}\ixpt\@ixpt}
\def\scriptsize{\@setsize\scriptsize{8pt}\viipt\@viipt}
\def\tiny{\@setsize\tiny{7pt}\vipt\@vipt}
\def\large{\@setsize\large{14pt}\xiipt\@xiipt}
\def\Large{\@setsize\Large{16pt}\xivpt\@xivpt}
\def\LARGE{\@setsize\LARGE{20pt}\xviipt\@xviipt}
\def\huge{\@setsize\huge{23pt}\xxpt\@xxpt}
\def\Huge{\@setsize\Huge{28pt}\xxvpt\@xxvpt}
\makeatother

\usepackage{times}
\usepackage[T1,OT1]{fontenc}
\pdfmapfile{+qtm.map}

\usepackage{amsmath,amsfonts,bm}

\newcommand{\modelid}[4]{{\rmfamily #1--#2--#3--#4}}

\newcommand{\papertablestyle}{%
  \normalfont\normalsize
  \setlength{\tabcolsep}{4pt}%
  \renewcommand{\arraystretch}{1.00}%
}

\newcommand{\effectci}[2]{%
  \begin{tabular}[c]{@{}c@{}}$#1$\\$[#2]$\end{tabular}}

\newcommand{\countpair}[2]{%
  \begin{tabular}[c]{@{}r@{}}#1 /\\#2\end{tabular}}

\def\eqref#1{equation~\ref{#1}}

\def\1{\bm{1}}

\newcommand{\valid}{\mathcal{D_{\mathrm{valid}}}}

\DeclareMathAlphabet{\mathsfit}{\encodingdefault}{\sfdefault}{m}{sl}
\SetMathAlphabet{\mathsfit}{bold}{\encodingdefault}{\sfdefault}{bx}{n}

\usepackage{amssymb}
\usepackage{hyperref}
\hypersetup{hypertexnames=false}
\usepackage{url}
\usepackage{graphicx}
\usepackage{wrapfig}
\usepackage{needspace}
\usepackage{placeins}
\usepackage{subcaption}
\usepackage{array}
\usepackage{pdflscape}
\usepackage{etoolbox}
\usepackage{listings}
\usepackage{algorithm}
\usepackage{algpseudocode}
\usepackage{booktabs}

\AtBeginEnvironment{tabular}{\papertablestyle}
\AtBeginEnvironment{tabular*}{\papertablestyle}

\title{Coverage Before Control: Route-Instruction Grounding and Steering for Controllable Retrosynthesis}

\author{%
Xuemin Chen, 
Xiaozhuang Song, 
Xinjian Zhao, 
Yaoyao Xu, 
Tianshu Yu\thanks{Corresponding author.}\\
{\normalfont The Chinese University of Hong Kong, Shenzhen}\\
{\normalfont Shanghai AI Lab}\\
{\normalfont\small\texttt{\{xueminchen,xiaozhuangsong1,xinjianzhao1,yaoyaoxu\}@link.cuhk.edu.cn}}\\
{\normalfont\small\texttt{yutianshu@cuhk.edu.cn}}\\
{\normalfont Code: \url{https://github.com/LOGO-CUHKSZ/RIGS}}
}
\hypersetup{pdfauthor={Xuemin Chen, Xiaozhuang Song, Xinjian Zhao, Yaoyao Xu, Tianshu Yu}}

\begin{document}

\maketitle

\begingroup
\setlength{\textfloatsep}{12pt plus 2pt minus 2pt}
\setlength{\floatsep}{8pt plus 2pt minus 2pt}

\begin{abstract}
Single-step retrosynthesis models are commonly evaluated by their ability to recover
recorded reactions. In practice, chemists may need to choose among several
precursor sets for the same product, for example to preserve a particular
motif. Recovering a recorded answer alone does not establish this ability to
follow a preference. Satisfying such requests requires both coverage of relevant
alternatives and control over which alternatives are favored. We introduce Route-Instruction
Grounding and Steering (RIGS), a two-stage framework for instruction-conditioned
retrosynthesis. Stage~A trains a language
projector, teaching it which alternatives an
instruction favors or discourages.  Stage~B uses the projector learned in Stage~A to steer a frozen
generative model through lightweight
residual adapters. We construct nested one-to-many training supports by pairing
each product with increasing numbers of candidate precursor sets. Extensive experiments demonstrate that broader support
helps the model generate a wider range of alternatives, and RIGS can learn to guide
generation according to instructions. The relationship between coverage and control
is consistent across model scales but non-monotone.
\end{abstract}

\section{Introduction}
\label{sec:introduction}

Retrosynthesis involves deciding which disconnection to pursue
\citep{thakkar2023disconnection}.
A target molecule can admit several plausible precursor sets
\citep{chen2019diverse}, and a chemist may prefer an alternative that preserves
a particular motif \citep{westerlund2025human}. Such preferences
call for a model that can generate alternatives for the same product and make
its output responsive to a natural-language request. We study this choice at
the single-step level; throughout this paper, a route denotes one
precursor-set alternative.

Top-$k$ exact-match accuracy is widely used to evaluate single-step
retrosynthesis models~\citep{igashov2024retrobridge,wang2025retrodiff,wang2026retrosynthesis}.
It measures recovery of a recorded precursor set, but can miss other
plausible alternatives~\citep{chen2019diverse,zagribelnyy2026single}.
It also does not directly test whether a model can select among
alternatives according to a user's request.
Preference-guided retrosynthesis helps align route choices with practical
goals such as material cost and yield~\citep{liu2024preference}.
Natural language lets chemists express requirements for each target, such
as which transformations to favor or avoid, as explored by Synthegy and
Synthelite~\citep{bran2026chemical,nguyen2025synthelite}.
Disconnection prompts and inference-time reward guidance
also provide ways to influence generated routes~\citep{thakkar2023disconnection,yadav2025retrosynflow}.
However, learning to ground free-form instructions in preferences among precursor
alternatives for the same product and transfer this grounding into direct
single-step generation remains less studied. We therefore ask how a generator
can learn such preferences without depending on a fixed candidate list at
inference.

This question depends on candidate coverage. In a motivating analysis, we
construct the USPTO multi-route collection (USPTO-MR) from
USPTO-Full~\citep{lowe2017uspto} by retaining products with at least two
distinct recorded precursor sets
(Appendix~\ref{app:uspto-mr-construction}).
Even this multi-route subset contains only 2.37 observed precursor sets
per product on average (Table~\ref{tab:main-data-summary}).
An instruction can promote an alternative only if the generator can produce
it. Synthetic expansion supplies additional candidates for base-model
training and preference supervision, but broader support alone does not
establish responsiveness to language. We therefore distinguish \emph{coverage},
the alternatives recovered under finite sampling, from \emph{control},
the instruction-dependent choice among those alternatives near the top of
the returned list. This motivates our organizing principle,
\emph{coverage before control}.

We introduce Route-Instruction Grounding and Steering (RIGS) to turn candidate
preferences into instruction-conditioned generation. RIGS contains two stages: Stage~A trains a language
projector using offline comparisons
among candidates for the same product, teaching it which alternatives an
instruction favors or discourages; Stage~B
transfers the projector to a lightweight instruction pathway connected through
residual adapters to the frozen GraphDiT module of RetroDiT
\citep{wang2026retrosynthesis}. At inference, RIGS
samples precursor sets directly.

We conduct a systematic experimental evaluation of RIGS to examine how
candidate coverage and instruction control interact. Broader training
support gives the generator more alternatives to choose from, while
instruction guidance can move preferred alternatives toward the top of
the returned list. The results reveal a consistent but non-monotone
interaction between coverage and control. For the 65M Top30 model, correct instructions improve
preferred-set Top-1 recovery by 18.52 percentage points over shuffled
instructions. Stage~A grounding further helps suppress alternatives that
conflict with the instruction, supporting the value of learning preferences
before steering generation.

Our contributions are threefold:
\begin{itemize}
  \item We introduce RIGS, which grounds language in product-local candidate
  preferences and transfers the learned representation to residual adapters on a
  frozen discrete-flow generator, enabling direct instruction-conditioned
  generation without test-time retrieval or reranking.
  \item We formulate and empirically examine the coverage-before-control
  principle. A shared-inventory study across four nested synthetic supports, an
  observed-reaction reference, and two model capacities separates candidate
  availability from instruction-dependent ranking.
  \item We provide controlled evaluations that test whether generation follows
  the instruction's content, isolate the contribution of Stage-A grounding, and
  assess instruction following with case-specific structural instructions and
  recorded reaction alternatives from Pistachio.
\end{itemize}

\section{Related Work}
\label{sec:related-work}

\paragraph{Single-step retrosynthesis and candidate coverage.}
RetroBridge models a product--reactant
Markov bridge \citep{igashov2024retrobridge}, while RetroDiff decomposes
generation into stages \citep{wang2025retrodiff} and DiffAlign addresses
permutation alignment \citep{laabid2025diffalign}.
Existing methods broaden candidate coverage through latent variables
\citep{chen2019diverse}, reaction-class prompts \citep{toniato2023diversity},
and training-data augmentation. RetroWISE uses synthetic reactions
\citep{zhang2024retrowise}, while the Triple Transformer Loop applies
templates with predictive filtering \citep{grandjean2026augmentation}.
\citet{tran2026failure} distinguish failures to propose candidates from
failures to rank them highly. We examine how nested training supports affect
instruction-dependent generation in RetroDiT's GraphDiT backbone
\citep{wang2026retrosynthesis}.

\paragraph{Grounding language in precursor preferences.}
MolT5 studies molecule--language translation \citep{edwards2022molt5},
MoleculeSTM learns joint structure--text representations \citep{liu2023moleculestm},
and Mol-Instructions develops biomolecular instruction tuning
\citep{fang2024molinstructions}.
In retrosynthesis, T-Rex uses generated descriptions to rank reaction centers
and precursors \citep{liu2024trex}, while RetroInText combines
descriptions with molecular and route context \citep{kang2025retrointext}.
RIGS grounds instructions in preferences among precursor sets for the same
product, learning which alternatives to favor or avoid.

\paragraph{Transferring preferences into generation.}
Multistep planners augment search \citep{segler2018planning,chen2020retrostar}
with Pareto objectives \citep{hastedt2026pareto}, while Ariadne conditions
route generation on depth and starting materials \citep{morgunov2026ariadne}.
More generally, discrete diffusion models support classifier-based
and classifier-free guidance \citep{schiff2025simple}.
Beyond Search steers autoregressive retrosynthesis with inference-time
classifier guidance \citep{laabid2026beyond}, while RetroSynFlow uses
Feynman--Kac reward steering
\citep{yadav2025retrosynflow}.
RIGS transfers learned precursor preferences to a frozen GraphDiT through
residual adapters, enabling direct instruction-conditioned generation.

\section{RIGS: Route-Instruction Grounding and Steering}
\label{sec:method}

RIGS learns which candidates an instruction favors, then transfers that
representation into a graph generator.  We train a separate no-instruction
Base model on the corresponding candidate support for each candidate budget
and model capacity.  Its backbone remains frozen during the two guidance
stages.  Stage~A (Sec.~\ref{sec:method-stage-a}) learns a language projector from offline preferences;
Stage~B (Sec.~\ref{sec:method-stage-b}) connects this projector to generation through residual adapters
(Figure~\ref{fig:architecture}).

\subsection{Candidate support and preference supervision}
\label{sec:method-support}

\noindent\begin{minipage}[t]{0.49\textwidth}
\vspace{0pt}
Retrosynthesis templates serve as reusable chemical reaction rules:
they encode local substructure patterns and associated bond changes.
Matching and applying these rules to a product $x$ with
RDChiral~\citep{coley2019rdchiral} yields complete precursor molecules
by carrying over molecular context outside the matched substructures
(Appendix~\ref{app:candidate-ranking}).
The resulting precursor sets form the candidate pool $\mathcal E(x)$.
We rank these candidates using a heuristic quality score and precursor
diversity. The rank $\operatorname{rank}_x(r)$ is candidate $r$'s first
position in this greedy ordering (Appendix~\ref{app:candidate-ranking},
Eq.~\ref{eq:candidate-selection-score}), with smaller ranks selected first.
For an integer budget $K\geq1$, we take the first $K$ ranks and apply shared
graph-construction checks and deduplication, denoted by
$\operatorname{Filter}$, to obtain the Base training support:
\end{minipage}\hfill
\begin{minipage}[t]{0.48\textwidth}
\vspace{0pt}
\centering
  \includegraphics[width=\linewidth]{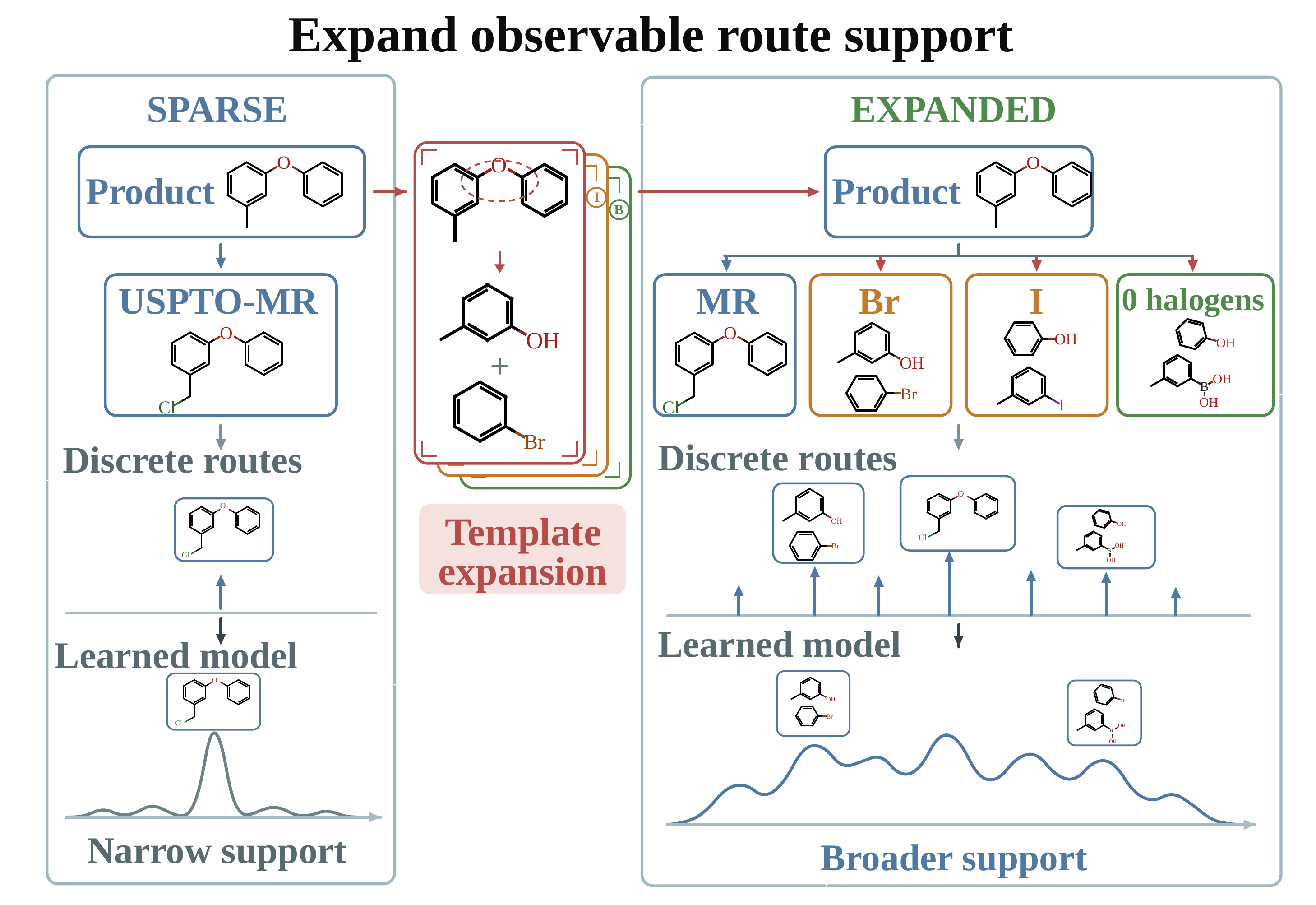}
  \captionsetup{labelfont=bf,hypcap=false,
    justification=raggedright,singlelinecheck=false}
  \captionof{figure}{\textbf{Support expansion.} Template-derived candidates
  broaden training support before instruction grounding.}
  \label{fig:support-expansion}
\end{minipage}
\par

\begin{equation}
  \begin{aligned}
    \mathcal A_K(x)
    &=\operatorname{Filter}\!\left(
      \{r\in\mathcal E(x):\operatorname{rank}_x(r)\leq K\}\right),\\
    \mathcal A_{K_1}(x)
    &\subseteq\mathcal A_{K_2}(x)\subseteq\cdots\subseteq\mathcal A_{K_m}(x),
      \qquad K_1<\cdots<K_m.
  \end{aligned}
  \label{eq:observable-support}
\end{equation}
Within each support $\mathcal A_K(x)$, we form preferred and avoided
candidate sets according to whether they follow a given guidance instruction.
These sets provide preference supervision for Stage~A grounding and
Stage~B instruction steering. Details of support construction and guidance
labeling are given in Appendices~\ref{app:candidate-ranking}
and~\ref{app:guidance-suite-construction}.

\begin{figure}[!htb]
  \centering
  \includegraphics[width=\textwidth]{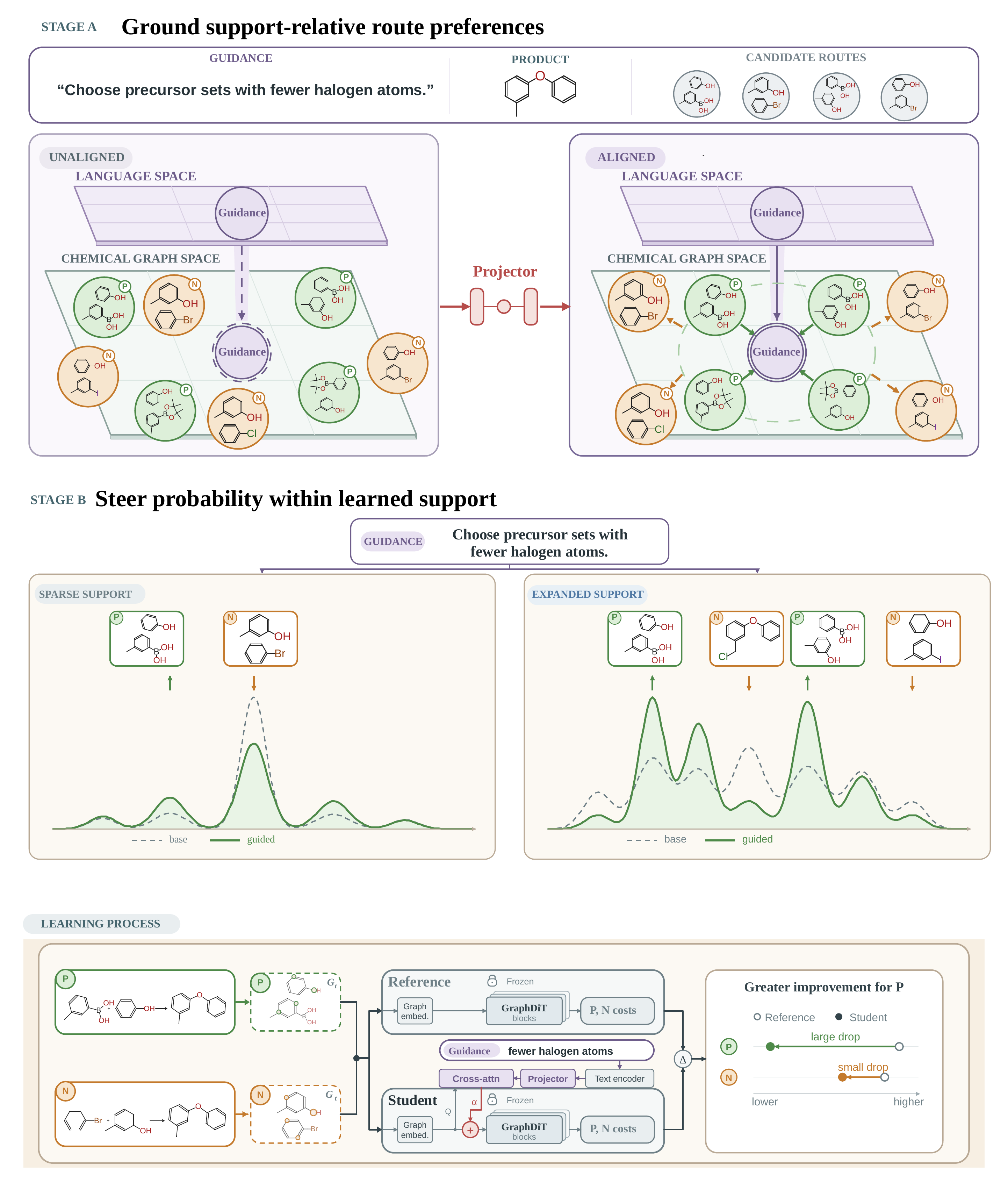}
  \caption{\textbf{RIGS-GraphDiT: grounding and steering.}
  Stage~A grounds natural-language guidance against preferred and avoided routes.
  Stage~B transfers this grounding into a residual instruction pathway that
  steers probability within the frozen generator's learned support.}
  \label{fig:architecture}
\end{figure}

\Needspace{6\baselineskip}
\subsection{Stage A: generator-free grounding}
\label{sec:method-stage-a}

For a target product $x$, let $r_1,\ldots,r_n$ denote alternative
precursor sets and $c$ a natural-language instruction. Stage~A learns to align instruction representations with
those of preferred candidates.
It uses frozen Qwen3-Embedding-8B~\citep{zhang2025qwen3embedding} as the
language encoder $E_T$ and a frozen reaction-pair encoder $E_R$ for each
product--precursor pair $(x,r_i)$ (Appendix~\ref{app:stage-a-inputs}).
The language encoder converts each instruction into a sequence of token
embeddings. We compute these text features offline and cache them for
reuse in both stages.
Stage~A trains only a projector $P_\phi$ to map the text features
into the reaction representation space.
We denote the instruction token embeddings by $H_c=E_T(c)$,
the candidate reaction embedding by $v_i=E_R(x,r_i)$, and
the projected instruction embeddings by $Z_c=P_\phi(H_c)$.
%

We use $Z_c$ and $\{v_i\}$ to predict a preference distribution
$\widehat q_\phi^c$ over candidates. We train the projector to match
this prediction to the target preference distribution $q^c$ constructed
from offline supervision.
The target distribution $q^c$ assigns higher weights to candidates
that better satisfy the instruction.
For instruction $c$, $\mathcal P_c$ contains the indices of positive
(preferred) candidates, and $\mathcal N_c$ contains those of negative
(avoided) candidates
(Appendix~\ref{app:guidance-suite-construction}).
The objective of Stage~A is:
\begin{equation}
  \begin{aligned}
    \mathcal L_A
    =\lambda_{\mathrm{list}}\,
      \operatorname{KL}(q^c\Vert\widehat q_\phi^c)
    \quad+\lambda_{\mathrm{pair}}\,
      \mathcal L_{\mathrm{pair}}(\mathcal P_c,\mathcal N_c).
  \end{aligned}
  \label{eq:method-stagea-objective}
\end{equation}
The KL term matches the predicted distribution to the target
preference weights, while the pairwise term encourages preferred
candidates to receive higher scores than avoided candidates.
Appendix~\ref{app:stage-a-details} specifies the candidate scoring
rule and the pairwise loss.
We minimize $\mathcal L_A$ by updating only $P_\phi$.
We select $P_{\phi^\star}$ on the validation set and
use its parameters to initialize Stage~B with a fresh optimizer.
Algorithm~\ref{alg:rigs-stage-a} summarizes this process.

\Needspace{6\baselineskip}
\subsection{Stage B: residual instruction steering}
\label{sec:method-stage-b}

Stage~B connects the instruction representation learned in Stage~A
to the frozen graph generator and reuses the cached text features
$H_c=E_T(c)$. We initialize the projectors from $P_{\phi^\star}$.
These projectors transform the cached text features into instruction
representations for node-to-text cross-attention adapters at selected
GraphDiT layers. Let $M_c^{(\ell)}$ denote the
projected text features supplied to layer $\ell$.
We denote the collection of these projected features by $M_c$.

At each selected layer $\ell$, the adapter $A_{\ell,\theta}$ uses
cross-attention to convert the projector's instruction features
$M_c^{(\ell)}$ into a residual update of the current graph-node states.
For node states $h_\ell$, frozen GraphDiT block
$B_\ell$, and steering strength $\alpha\geq0$, the update is
\begin{equation}
  h_{\ell+1}=B_\ell\!\left(
    h_\ell+\alpha A_{\ell,\theta}(h_\ell,M_c^{(\ell)})\right).
  \label{eq:method-alpha-residual}
\end{equation}
Setting $\alpha=0$ recovers the Base model.
The projector keeps a separate embedding for each instruction token,
so each graph node can use different parts of the instruction during
generation.
The adapters update only node representations. The frozen backbone
then uses these updates to adjust edge representations.
Appendix~\ref{app:stage-b-details} specifies projector sharing and
the full masked attention update.

\paragraph{Training objective.}
We optimize only the projector and adapter parameters $\theta$,
training the guided model to denoise preferred precursor sets while
regularizing its departure from the frozen Base model, which serves
as the reference. With steering strength fixed at
$\alpha_{\mathrm{train}}=1$, the core objective is:
\begin{equation}
  \mathcal L_{B,\mathrm{core}}
  =\lambda_{\mathrm{pos}}\mathcal L_{\mathrm{pos}}
   +\lambda_{\mathrm{rank}}\mathcal L_{\mathrm{rank}}
   +\lambda_{\mathrm{roll}}\mathcal L_{\mathrm{roll}}
   +\lambda_{\mathrm{guard}}\mathcal L_{\mathrm{guard}}.
  \label{eq:method-stageb-core}
\end{equation}
The positive flow-matching loss $\mathcal L_{\mathrm{pos}}$
directly trains denoising on preferred candidates.
Let $G_i$ be the clean precursor graph of candidate $r_i$ and
$G_{i,t}$ its noisy precursor graph at time $t$. The reference $f_0$
and guided model $f_{\theta,1}$ use the same $G_{i,t}$, with
denoising losses
\begin{equation}
  \begin{aligned}
    D_i^{\mathrm{ref}}
      &=\ell_{\mathrm{den}}\!\left(f_0(G_{i,t},x,t),G_i\right),\\
    D_i^{\mathrm{student}}
      &=\ell_{\mathrm{den}}\!\left(f_{\theta,1}(G_{i,t},x,t,M_c),G_i\right).
  \end{aligned}
  \label{eq:method-route-gain}
\end{equation}
Here, $\ell_{\mathrm{den}}$ sums the mean node and weighted mean edge
cross-entropies over valid positions. The gain
$d_i=D_i^{\mathrm{ref}}-D_i^{\mathrm{student}}$ is positive when guidance
reduces the denoising loss.
The ranking loss $\mathcal L_{\mathrm{rank}}$ penalizes violations
of the margin $d_i-d_j\geq m_{\mathrm{rank}}$ for
$i\in\mathcal P_c$ and $j\in\mathcal N_c$, encouraging preferred
candidates to benefit more from guidance than avoided candidates.
Ranking alone can satisfy this margin by worsening denoising on
avoided candidates; $\mathcal L_{\mathrm{pos}}$ additionally provides
a direct training signal to reconstruct preferred candidates.

Two KL penalties regularize deviations from the reference predictions.
Let $p_{\mathrm{ref}}$ and $p_{\mathrm{student}}$ denote the models'
categorical predictions at the same node or edge position of a
noisy precursor graph. For each product--instruction group $g$, $K_g$ is the
weighted mean node-and-edge KL across its noisy precursor graphs.
The penalties take the form
\begin{equation}
    \mathcal L_{\mathrm{roll}}
      =\left\langle
        \operatorname{KL}
        (p_{\mathrm{ref}}\Vert p_{\mathrm{student}})
        \right\rangle_{\mathrm{roll}},
    \mathcal L_{\mathrm{guard}}
      =\left\langle
        [K_g-\delta_{\mathrm{KL}}]_+
        \right\rangle_g.
  \label{eq:method-reference-regularization}
\end{equation}
$\mathcal L_{\mathrm{roll}}$ penalizes prediction drift on
noisy precursor graphs encountered during generation, while
$\mathcal L_{\mathrm{guard}}$ penalizes only the amount by which
a group's mean KL exceeds $\delta_{\mathrm{KL}}$.
For instructions about numerical properties, we add
$\lambda_{\mathrm{teach}}\mathcal L_{\mathrm{teach}}$ to the core
objective. This loss encourages larger denoising gains $d_i$ for
candidates whose property values better satisfy the instruction.
Appendix~\ref{app:stage-b-loss-details} gives the detailed loss
definitions, and Appendix~\ref{app:stageb-loss-ablation} reports
their ablation.
Algorithms~\ref{alg:rigs-stage-a} and~\ref{alg:rigs-stage-b}
summarize the two training stages.

\begin{figure*}[!htb]
\centering
\begin{minipage}[t]{0.485\textwidth}
  \vspace{0pt}
  \captionsetup{labelfont=bf,hypcap=false,
    justification=raggedright,singlelinecheck=false}
  \captionof{algorithm}{Stage A: generator-free grounding}
  \label{alg:rigs-stage-a}
  \hrule
  \vspace{2pt}
  \normalsize
  \begin{algorithmic}[1]
    \Require $\mathcal D_{\mathrm{tr}},\mathcal D_{\mathrm{val}}$; frozen $E_T,E_R$
    \State Cache $H_c\leftarrow E_T(c)$ and
      $v_i\leftarrow E_R(x,r_i)$ over $\mathcal D_{\mathrm{tr}}$
    \For{each training iteration}
      \State Sample groups with cached encoder outputs
      \State $Z_c\leftarrow P_\phi(H_c)$
      \State Form the predicted preference distribution
        $\widehat q_\phi^c$ from $Z_c,\{v_i\}$ via
        Eq.~(\ref{eq:grounding-score})
      \State Compute $\mathcal L_A$ from available labels via
        Eq.~(\ref{eq:method-stagea-objective})
      \State Update only $\phi$ using $\nabla_\phi\mathcal L_A$
    \EndFor
    \State \Return $P_{\phi^\star}$ selected on $\mathcal D_{\mathrm{val}}$
  \end{algorithmic}
  \vspace{2pt}
  \hrule
\end{minipage}\hfill
\begin{minipage}[t]{0.485\textwidth}
  \vspace{0pt}
  \captionsetup{labelfont=bf,hypcap=false,
    justification=raggedright,singlelinecheck=false}
  \captionof{algorithm}{Stage B: residual instruction steering}
  \label{alg:rigs-stage-b}
  \hrule
  \vspace{2pt}
  \normalsize
  \begin{algorithmic}[1]
    \Require $\mathcal D_{\mathrm{tr}},\mathcal D_{\mathrm{val}}$;
      frozen $f_0,E_T$; $P_{\phi^\star}$
    \State Initialize projectors from $P_{\phi^\star}$; zero the
      adapter output projections
    \For{each training iteration}
      \State Sample candidate groups and shared noisy precursor graphs
      \State Compute denoising gains $d_i$ from
        Eq.~(\ref{eq:method-route-gain})
      \State Compute $\mathcal L_{\mathrm{pos}},\mathcal L_{\mathrm{rank}},
        \mathcal L_{\mathrm{guard}}$
      \State Periodically sample guided states and replay them through
        both models to compute $\mathcal L_{\mathrm{roll}}$
      \State Form $\mathcal L_B$ via Eq.~(\ref{eq:method-stageb-core}); add
        $\lambda_{\mathrm{teach}}\mathcal L_{\mathrm{teach}}$
        for metric instructions
      \State Update only $\theta$ using $\nabla_\theta\mathcal L_B$
    \EndFor
    \State \Return $f_{\theta^\star,\alpha_{\mathrm{eval}}}$ selected on
      $\mathcal D_{\mathrm{val}}$
  \end{algorithmic}
  \vspace{2pt}
  \hrule
\end{minipage}
\end{figure*}

\paragraph{Inference.}
We reuse the Base continuous-time Markov chain (CTMC) sampler \citep{wang2026retrosynthesis}.
For product $x$ and instruction $c$, we cache the projected instruction
memory $M_c$ and use $f_{\theta^\star,\alpha_{\mathrm{eval}}}$ to supply
instruction-conditioned endpoint predictions at each sampling step.
To obtain a ranked list of distinct precursor candidates, we canonicalize
the valid outputs from $B$ raw draws, merge duplicates, and rank candidates
by how often they were sampled.
The CTMC rates and update are given in
Eqs.~(\ref{eq:method-guided-rate})--(\ref{eq:method-guided-sampling});
Appendix~\ref{app:test-protocol} specifies the sampling protocol.
For each model, we select the checkpoint $\theta^\star$ and
guidance strength $\alpha_{\mathrm{eval}}$ on the validation set,
using instruction-following performance and output validity.
We keep both fixed during test evaluation.
The evaluated scales and selection rules are detailed in
Appendix~\ref{app:common-top100-protocol} and Supplementary Section~S3.
\FloatBarrier

\section{Experiments}
\label{sec:experiments}

We evaluate how training support affects candidate coverage and instruction
alignment in RIGS. We then examine whether instruction grounding and
generation-time conditioning improve control, and assess RIGS's
applicability across datasets and guidance suites.

\subsection{Experimental Setup}
\label{sec:experimental-design}

\noindent\textbf{Datasets.}\quad
We compare recorded precursor alternatives in USPTO-MR with synthetic-only
template expansions, USPTO-MR-Top$K$ (Top$K$), for
$K\in\{15,30,45,60\}$. USPTO-MR groups reactions from
USPTO-Full~\citep{lowe2017uspto} by canonical product and retains products
with multiple distinct recorded precursor sets. All expansions use a fixed
retrosynthesis template library released with DESP~\citep{yu2024desp},
shared across disjoint product splits. Construction details and support
statistics appear in Appendices~\ref{app:uspto-mr-construction}--\ref{app:candidate-ranking}.

\noindent\textbf{Natural-language guidance.}\quad
We construct \emph{RDKit5}, a natural-language guidance dataset that pairs
instructions with preferred and avoided precursor sets.
Instructions express preferences such as \emph{Choose precursor sets
with fewer halogen atoms}. We assign these labels using property values
computed with RDKit~\citep{rdkit} within each product's candidate pool
(Appendix~\ref{app:guidance-suite-construction}).

\noindent\textbf{Evaluation setup.}\quad
We adopt the 8M (Large) and 65M (X-Large) backbone sizes from
RetroDiT~\citep{wang2026retrosynthesis}. For each support, we evaluate both
sizes on one shared Top100 inventory of 1,201 test products.
Reference is the support-matched unguided Base;
Matched and Shuffled use the same guided model with the case's instruction or
a different instruction, respectively. Each arm uses $B=100$ raw draws per group
without backfilling. We select checkpoints and guidance scales
$\alpha_{\mathrm{eval}}$ using validation data and keep them fixed for testing
(Appendix~\ref{app:common-top100-protocol}).

\noindent\textbf{Evaluation metrics.}\quad
Using the first $B$ samples from each product--instruction group, we
report two hit rates. Hit$_{\mathrm{all}}@B$ measures how often these
samples contain at least one precursor set from the product's Top100
inventory. Hit$_+@B$ measures how often they contain at least one
candidate from the instruction's preferred subset.
For each test product, we combine outputs from all instruction groups
and count how many distinct precursor sets from its Top100 inventory
were generated. Pooled recovery sums these counts across test products.
For each group, we canonicalize the 100 sampled outputs and rank the
distinct valid precursor sets by decreasing sampling frequency.
P@$k$ ($\uparrow$) and N@$k$ ($\downarrow$) measure how often the top $k$
candidates in this ranking contain at least one preferred or avoided
candidate, respectively.
Raw validity is the percentage of all sampled outputs that pass RDKit's
molecular structure checks, counting repeated outputs each time.
For unique-valid count, we remove duplicate valid outputs within each
group, then average the remaining counts across groups.
Further metric definitions and sampling settings appear in
Appendices~\ref{app:evaluation-metrics} and~\ref{app:test-protocol}, respectively.

\begin{figure*}[!htb]
  \centering
  \includegraphics[width=\textwidth]{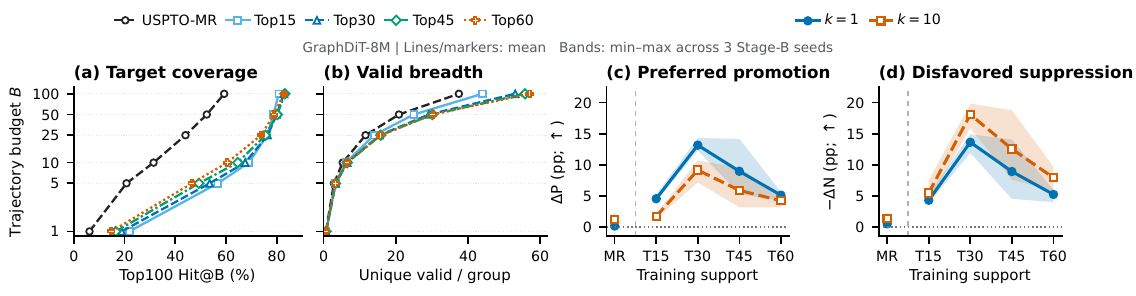}
  \par
  \includegraphics[width=\textwidth]{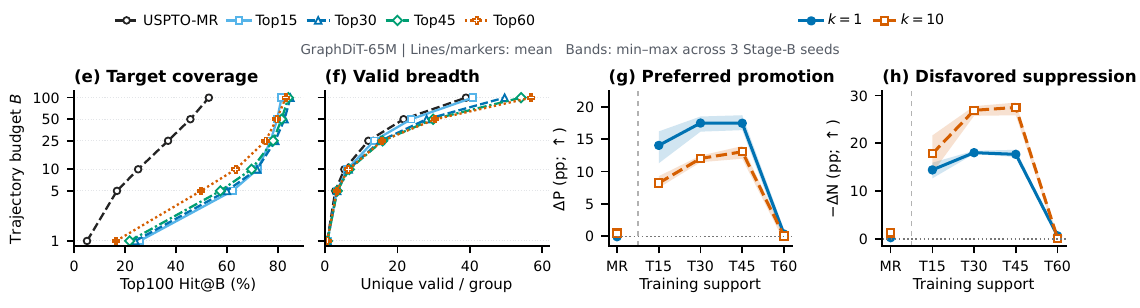}
  \caption{\textbf{Candidate coverage and instruction control across training
  supports.} GraphDiT-8M (top) and 65M (bottom) on the same Top100 test
  inventory, using equal-axis, case-balanced means. Left panels show Reference:
  (a,e) the rate of sampling at least one candidate from this inventory and
  (b,f) the number of distinct valid outputs per group, with raw-draw budget
  $B$ on the vertical axis. Right panels compare Matched with Shuffled
  instructions at $B=100$: (c,g) increases in preferred-set hit rate,
  $\Delta\mathrm{P}@k$, and (d,h) reductions in avoided-set hit rate,
  $-\Delta\mathrm{N}@k$, for $k=1,10$. Positive values indicate better
  instruction alignment. Control markers show means over three Stage-B
  training seeds; shaded bands span their minimum and maximum.
  Each seed uses its validation-selected checkpoint and guidance scale;
  Reference is fixed for each support. Result summaries and seed analyses
  can be found in Appendices~\ref{app:common-top100-complete}
  and~\ref{app:all-support-seed-robustness}, respectively.}
  \label{fig:common-top100-coverage-control}
\end{figure*}

\subsection{Broader training support improves candidate coverage}
\label{sec:coverage-control-results}

With the same sampling budget, unguided Reference models trained on Top15
recover more distinct precursor alternatives from the shared Top100 test
inventory than those trained on USPTO-MR. Pooled recovery increases from
2,348 to 12,880 for the 8M model and from 1,723 to 13,661 for the 65M model
(Table~\ref{tab:main-coverage-control}). Expanding support from Top15 to Top30 and Top45 further increases pooled
recovery, with smaller gains at each step. From Top45 to Top60, pooled
recovery is nearly unchanged for the 8M model and decreases for the 65M
model, although both models generate more distinct valid outputs
(Figure~\ref{fig:common-top100-coverage-control}~(b,f);
Table~\ref{tab:main-coverage-control}). Thus, generating more different
structures does not always mean recovering more candidates in the test inventory.
We next examine whether RIGS can use instructions to make preferred
alternatives appear earlier in the returned list.

\begin{table*}[!htb]
  \caption{\textbf{Candidate coverage and instruction following at
  validation-selected settings.}
  Results use $B=100$ per group and Stage-B seed 42. Pooled recovery counts
  distinct product--candidate pairs recovered by Reference from the Top100
  inventory across all instruction groups. Effects are in percentage points;
  positive values indicate improvement. Bold marks maxima in the effect
  columns within each model size. Absolute rates and Matched--Shuffled
  intervals appear in Tables~\ref{tab:common-top100-absolute-results}
  and~\ref{tab:common-top100-control-summary}, respectively.
  Matched--Reference intervals for P@1, P@10, and N@10 appear in
  Table~\ref{tab:common-top100-unguided-contrasts}.
  Three-seed summaries can be found in Appendix~\ref{app:all-support-seed-robustness}.}
  \label{tab:main-coverage-control}
  \centering
  \appto\papertablestyle{\scriptsize\setlength{\tabcolsep}{2.5pt}%
    \renewcommand{\arraystretch}{1.15}}
  \begin{tabular*}{\linewidth}{@{\extracolsep{\fill}}lr|rr|rrrr|rrrr@{}}
    \toprule
    & Reference & \multicolumn{2}{c|}{Absolute P@1 (\%)}
      & \multicolumn{4}{c|}{Matched--Shuffled (pp)}
      & \multicolumn{4}{c}{Matched--Reference (pp)} \\
    \cline{3-4}\cline{5-8}\cline{9-12}
    Support & \shortstack{Pooled\\recovery} & Reference & Matched
      & $\Delta\mathrm{P}@1$ & $\Delta\mathrm{P}@10$
      & $-\Delta\mathrm{N}@1$ & $-\Delta\mathrm{N}@10$
      & $\Delta\mathrm{P}@1$ & $\Delta\mathrm{P}@10$
      & $-\Delta\mathrm{N}@1$ & $-\Delta\mathrm{N}@10$ \\
    \midrule
    \multicolumn{12}{@{}l}{\textbf{GraphDiT-8M}} \\
    USPTO-MR & 2,348 & 8.48 & 8.98 & +0.08 & +2.03 & +0.74 & +1.37 & +0.51 & +1.91 & +0.86 & +2.54 \\
    \midrule
    Top15 & 12,880 & 30.16 & 34.10 & +4.10 & +1.37 & +3.83 & +4.65 & +3.95 & +0.27 & +4.06 & +5.16 \\
    Top30 & 16,922 & 30.20 & 40.20 & +12.89 & \textbf{+10.70} & \textbf{+14.06} & +18.52 & +10.00 & -0.66 & \textbf{+18.24} & \textbf{+27.73} \\
    Top45 & 17,370 & 29.84 & 40.63 & \textbf{+14.18} & +9.53 & \textbf{+14.06} & \textbf{+18.79} & \textbf{+10.78} & \textbf{+2.62} & +15.20 & +24.92 \\
    Top60 & 17,378 & 29.22 & 33.05 & +5.63 & +4.88 & +5.94 & +8.36 & +3.83 & +1.17 & +5.55 & +10.08 \\
    \midrule
    \multicolumn{12}{@{}l}{\textbf{GraphDiT-65M}} \\
    USPTO-MR & 1,723 & 7.70 & 8.16 & +0.00 & +0.23 & -0.20 & +1.60 & +0.47 & +2.30 & -0.08 & +0.70 \\
    \midrule
    Top15 & 13,661 & 30.23 & 41.25 & +11.29 & +7.50 & +12.66 & +13.83 & +11.02 & +2.42 & +15.00 & +19.57 \\
    Top30 & 20,075 & 32.93 & 48.83 & \textbf{+18.52} & \textbf{+12.19} & \textbf{+17.89} & \textbf{+27.19} & \textbf{+15.90} & \textbf{+4.57} & +16.56 & \textbf{+34.57} \\
    Top45 & 21,562 & 30.59 & 45.74 & +17.30 & +11.99 & +17.11 & +25.86 & +15.16 & +3.79 & \textbf{+18.32} & +32.97 \\
    Top60 & 19,241 & 30.20 & 29.14 & +1.09 & +0.35 & +0.39 & +0.63 & -1.05 & -0.43 & -0.23 & +1.72 \\
    \bottomrule
  \end{tabular*}
\end{table*}

\subsection{Instructions help the model return candidates that match the request}
\label{sec:topk-control}

To test whether instructions help preferred candidates rank higher, we
compare three conditions: Reference generates without instructions,
Matched uses the intended instruction, and Shuffled uses one expressing
a different preference.
We first compare Matched with Reference to measure the
benefit of instruction guidance. At 65M Top30, P@1 increases from 32.93\%
to 48.83\%, while N@10 decreases by 34.57 percentage points
(Table~\ref{tab:main-coverage-control}). RIGS more often ranks a preferred
candidate first. Its top 10 outputs are also less likely to include an
avoided candidate.
We then compare Matched with Shuffled to test whether this benefit
depends on the instruction's content.
At 65M Top30, Matched improves P@1 by 18.52 percentage points over Shuffled.
Compared with Shuffled, Matched generates about five more preferred and
five fewer avoided outputs per 100 samples on average
(Table~\ref{tab:empirical-mass-reallocation}). These comparisons show that
RIGS can steer precursor generation according to the instruction's content.

Instruction-following gains vary with training support in a consistent
but non-monotone pattern across model capacities. On observed USPTO-MR
support, Matched--Shuffled P@1 gains are near zero. They increase at Top15
and are largest at Top30/Top45 (Table~\ref{tab:main-coverage-control}).
At Top30, the gains over unguided generation are concentrated at Top-1:
the 65M model improves by 15.90 percentage points at Top-1 and 4.57 at
Top-10, while the 8M model improves at Top-1 with little change at Top-10
(Appendix~\ref{app:topk-reference-effects}).
At Top60, Matched--Shuffled gains decline at both capacities, with little
improvement for the 65M model across all three Stage-B seeds
(Appendix~\ref{app:all-support-seed-robustness}). Additional analyses
examine how guidance strength affects instruction control and output
validity and compare supports under alternative training budgets
(Appendices~\ref{app:validation-alpha-tradeoff}--\ref{app:top45-top60-alpha-pareto}
and~\ref{app:compute-matched-support-scaling}).

\subsection{Roles of grounding and generation-time guidance}
\label{sec:grounding-baseline}

We show what Stage~A grounding adds to direct Stage-B training and
demonstrate the benefits of guiding generation over simply reranking
unguided outputs.

\paragraph{Stage-A grounding.}
We compare 8M Top30 models with Stage~A grounding (RIGS) and without
it (Direct Stage-B), using the same Stage-B training and inference settings.
On a separate test cohort, Stage~A grounding increases the
Matched--Shuffled reduction in N@10 by 6.82 percentage points
(Table~\ref{tab:main-grounding}). Stage~A trains the projector to score
preferred candidates above avoided ones for the same instruction.
Consistent with this objective, grounding mainly helps suppress
candidates that conflict with the request
(Table~\ref{tab:stagea-initialization-frozen-test-did}).

\paragraph{Guidance during generation.}
We compare guidance during generation with post-hoc reranking of
unguided Reference outputs, using 100 samples per group for both.
RIGS ranks its guided outputs by sampling frequency; the reranker uses
a scalar scorer trained on signed RDKit utility.
A preferred candidate appears among the samples in about 64\% of groups
for both methods, but RIGS achieves higher P@1: 32.73\% versus 22.23\%
(Table~\ref{tab:budgeted-proposal-readout}). The reranker lowers N@1,
yet its P@1 also falls below the unguided baseline's 24.57\%.
Thus, this reranker suppresses avoided candidates at the cost of selecting
fewer known preferred candidates. RIGS improves both outcomes over
unguided generation.
Appendix~\ref{app:budgeted-proposal-selection} gives the selector details
and full comparison.

\begingroup
\setlength{\intextsep}{4pt}
\begin{table}[!htb]
  \captionsetup{skip=4pt}
  \centering
  \begin{minipage}[t]{0.48\textwidth}
    \vspace{0pt}
    \caption{\textbf{Effect of Stage-A grounding.}
    On a separate test cohort, we compare 8M Top30 models with and without
    Stage-A pretraining. Each row shows Matched--Shuffled gains in
    percentage points; higher is better
    (Appendix~\ref{app:stagea-initialization-ablation}).}
    \label{tab:main-grounding}
    \label{tab:main-controlled-alternatives}
    \centering
    \appto\papertablestyle{\setlength{\tabcolsep}{3pt}}
    \begin{tabular}{@{}lrrr@{}}
      \toprule
      Variant & \shortstack{$\Delta\mathrm{P}$\\@1}
        & \shortstack{$-\Delta\mathrm{N}$\\@1}
        & \shortstack{$-\Delta\mathrm{N}$\\@10} \\
      \midrule
      RIGS & +12.77 & +10.69 & +21.21 \\
      Direct Stage-B & +11.96 & +7.67 & +14.39 \\
      \bottomrule
    \end{tabular}
  \end{minipage}\hfill
  \begin{minipage}[t]{0.48\textwidth}
    \vspace{0pt}
    \caption{\textbf{Generation-time guidance versus reranking.}
    We use 100 raw draws per group for Top30. Hit$_+@100$ reports how
    often they include a preferred candidate.
    Appendix~\ref{app:budgeted-proposal-selection} details the separate cohort.}
    \label{tab:budgeted-proposal-readout}
    \centering
    \appto\papertablestyle{\setlength{\tabcolsep}{3pt}}
    \begin{tabular}{@{}lrrr@{}}
      \toprule
      Method & \shortstack{Hit$_+@100$\\(\%)}
        & \shortstack{P@1 $\uparrow$\\(\%)}
        & \shortstack{N@1 $\downarrow$\\(\%)} \\
      \midrule
      Reference & 64.80 & 24.57 & 25.39 \\
      Reranking & 64.80 & 22.23 & \textbf{3.32} \\
      \midrule
      \textbf{RIGS} & 63.87 & \textbf{32.73} & 8.75 \\
      \bottomrule
    \end{tabular}
  \end{minipage}
\end{table}
\endgroup

\FloatBarrier
\subsection{Applicability across datasets and guidance suites}
\label{sec:free-form-language}


We assess the generality of RIGS across reaction datasets and guidance
types. We also present a case study illustrating how RIGS supports
multistep retrosynthetic planning.


\begin{table}[!htb]
  \captionsetup{skip=4pt}
  \caption{\textbf{Control across datasets and guidance suites.}
  We evaluate separately trained Top30 models for each dataset and guidance
  suite. Positive values indicate better instruction alignment. Cohort and
  aggregation details for Pistachio, Structure, and Metric4 can be found in
  Appendices~\ref{app:pistachio-replication},
  \ref{app:structure-embedding-ablation}, and~\ref{app:metric4-generalization}, respectively.}
  \label{tab:main-dataset-guidance}
  \label{tab:main-pistachio}
  \label{tab:main-structure}
  \centering
  \begin{tabular*}{\linewidth}{@{\extracolsep{\fill}}lllrrrr@{}}
    \toprule
    & & & \multicolumn{2}{c}{Matched--Shuffled (pp)}
      & \multicolumn{2}{c}{Matched--Reference (pp)} \\
    \cmidrule(lr){4-5}\cmidrule(lr){6-7}
    Reaction data & Guidance & Backbone & $\Delta$P@1 $\uparrow$
      & $-\Delta$N@10 $\uparrow$ & $\Delta$P@1 $\uparrow$
      & $-\Delta$N@10 $\uparrow$ \\
    \midrule
    Pistachio & RDKit5 & 65M & +1.52 & +7.19 & +0.98 & +13.59 \\
    \midrule
    USPTO & Structure & 8M & +3.55 & +3.43 & +5.54 & +10.57 \\
          &           & 65M & +8.09 & +18.44 & +14.74 & +35.79 \\
    \midrule
    USPTO & Metric4 & 8M & +5.37 & +7.41 & +4.07 & +8.04 \\
          &         & 65M & +15.78 & +21.92 & +13.44 & +27.52 \\
    \bottomrule
  \end{tabular*}
\end{table}


\paragraph{Cross-dataset evaluation on Pistachio.}
To assess the generality of RIGS across reaction datasets, we separately
train a 65M Top30 model on Pistachio, a commercial reaction database
\citep{mayfield2021pistachio}. We construct the Top30 training support using
the same template library as in the USPTO experiments. We evaluate on a
fixed cohort of held-out Pistachio products with recorded precursor
alternatives. Details of cohort construction can be found in
Appendix~\ref{app:pistachio-replication}.
Matched improves P@1 by 1.52 percentage points and reduces N@10 by 7.19
points relative to Shuffled, with improvements over Reference as well
(Table~\ref{tab:main-dataset-guidance}).
These results support the generality of RIGS across reaction datasets.

\paragraph{Case-specific structural instructions.}
We use the \emph{Structure} guidance dataset to test whether RIGS can follow
free-form natural-language instructions.
We generate case-specific instructions from structural evidence,
describing molecular features to retain or avoid.
On the test guidance set of 4,606 distinct instructions
(Appendix~\ref{app:structure-embedding-ablation}), Matched instructions
promote preferred candidates and suppress avoided candidates relative to
Shuffled at both model capacities
(Table~\ref{tab:main-dataset-guidance}). At 65M, P@1 improves by 8.09 percentage
points and N@10 decreases by 18.44 points. These gains support RIGS's
ability to follow case-specific structural requests expressed in free-form
natural language.

\paragraph{Synthesis-related guidance.}
We use the \emph{Metric4} guidance dataset to test whether RIGS can follow
instructions about synthesis-related properties.
Metric4 combines four metrics: reduction in synthetic complexity,
procurement burden, halogen count, and ring count
(Appendix~\ref{app:guidance-suite-construction}).
The Top30 results show preferred-candidate
promotion and avoided-candidate suppression at both capacities
(Table~\ref{tab:main-dataset-guidance}). Across all synthetic supports,
we observe improved instruction following in seven of eight settings
(Appendix~\ref{app:metric4-generalization}).

\paragraph{Multistep planning.}
To examine how RIGS can support multistep retrosynthetic planning, we use it
as the single-step precursor generator in a custom value-guided AND--OR
tree-search planner inspired by Retro*~\citep{chen2020retrostar}.
The planner recursively expands precursor molecules to construct a
synthesis route. In this case study, conditioning
each expansion on the instruction to use fewer halogen atoms yields a
complete route with halogen-free starting materials. This illustrates how
RIGS can guide precursor selection across multiple retrosynthetic steps
(Figure~\ref{fig:multistep-selected-proofs}B in
Appendix~\ref{app:multistep-planning-witnesses}).

\section{Conclusion}
\label{sec:conclusion}

We studied how single-step retrosynthesis can generate precursor alternatives
that respond to a chemist's preferences. RIGS grounds instructions in
preferences among alternatives for the same product, then transfers the
learned representation to a frozen graph generator through lightweight
residual adapters. Across nested training supports and two model capacities,
synthetic expansion improves recovery of known alternatives over
observed-reaction training, but further expansion does not monotonically
improve coverage or instruction control. Matched-versus-shuffled comparisons
show that RIGS can promote preferred candidates and suppress conflicting ones,
with Stage-A grounding providing its clearest additional benefit in conflict
suppression. Evaluations with case-specific structural instructions further
support control beyond predefined molecular descriptors. The results
reveal a consistent but non-monotone interaction between coverage and
control across model capacities.

\endgroup

\begingroup
\fontencoding{T1}\fontfamily{qtm}\selectfont
\bibliography{references}
\bibliographystyle{references}
\endgroup

\clearpage
\appendix

\raggedbottom
\setlength{\textfloatsep}{12pt plus 2pt minus 2pt}
\setlength{\floatsep}{10pt plus 2pt minus 2pt}
\setlength{\intextsep}{10pt plus 2pt minus 2pt}
\captionsetup{skip=6pt}
\renewcommand{\bottomfraction}{0.9}
\renewcommand{\floatpagefraction}{0.9}
\setcounter{topnumber}{3}
\setcounter{bottomnumber}{3}
\setcounter{totalnumber}{5}
\makeatletter
\setlength{\@fptop}{0pt}
\setlength{\@fpsep}{12pt plus 2pt minus 2pt}
\setlength{\@fpbot}{0pt plus 1fil}
\makeatother

\pdfbookmark[0]{Appendix contents}{appendix-contents}
\section*{Appendix contents}
\makeatletter
\begingroup
  \c@tocdepth=2
  \hypersetup{linktoc=all}
  \setlength{\parskip}{0pt}
  \renewcommand*\l@section[2]{%
    \addpenalty\@secpenalty
    \addvspace{0.45em}%
    \begingroup\bfseries
      \@dottedtocline{1}{0em}{1.8em}{#1}{#2}%
    \endgroup}
  \renewcommand*\l@subsection{\@dottedtocline{2}{1.8em}{2.6em}}
  \@starttoc{apc}
\endgroup
\let\appendix@addcontentsline\addcontentsline
\renewcommand*\addcontentsline[3]{%
  \appendix@addcontentsline{#1}{#2}{#3}%
  \ifstrequal{#1}{toc}{%
    \addtocontents{apc}{%
      \protect\contentsline{#2}{#3}{\thepage}{\@currentHref}%
      \protected@file@percent}}{}}
\makeatother
\clearpage

\section{Common-inventory protocol and frozen coordinates}
\label{app:common-top100-protocol}

\paragraph{Controlled support sequence.}
Top15/30/45/60 are nested synthetic-only prefixes of one product-local
candidate ordering.  USPTO-MR is a separate natural-data reference.
The primary training recipe uses within-family data passes, so realized
update counts differ with support size.  Exact budgets appear in
Appendix~\ref{app:implementation}; the common-update-ceiling sensitivity
appears in Appendix~\ref{app:compute-matched-support-scaling}.

\paragraph{Validation calibration.}
Calibration targets high Matched P@1, with trajectory validity informing
scale and checkpoint choice.  The primary study retains each run's recorded
validation-selected coordinate (Table~\ref{tab:common-top100-coordinates}).
Supplementary Section~S3 documents the support-specific selection rules,
including validation gates, fallback decisions, and recorded amendments.

\begin{table*}[!htbp]
  \caption{Frozen coordinates for the ten
  \modelid{USPTO}{Guide(RDKit5)}{8M}{[Support]} and
  \modelid{USPTO}{Guide(RDKit5)}{65M}{[Support]} models in the Common Top100
  study. We selected all coordinates on validation before opening the common
  test; every row uses Reference, Matched, and Shuffled arms.}
  \label{tab:common-top100-coordinates}
  \centering
  \begin{tabular}{lrr}
    \toprule
    Guided model ID & Stage-B epoch & $\alpha_{\mathrm{eval}}$ \\
    \midrule
    \modelid{USPTO}{Guide(RDKit5)}{8M}{USPTO-MR} & 14 & .50 \\
    \modelid{USPTO}{Guide(RDKit5)}{8M}{Top15}    &  6 & .20 \\
    \modelid{USPTO}{Guide(RDKit5)}{8M}{Top30}    & 12 & .50 \\
    \modelid{USPTO}{Guide(RDKit5)}{8M}{Top45}    & 14 & .50 \\
    \modelid{USPTO}{Guide(RDKit5)}{8M}{Top60}    &  6 & .25 \\
    \midrule
    \modelid{USPTO}{Guide(RDKit5)}{65M}{USPTO-MR} & 14 & .50 \\
    \modelid{USPTO}{Guide(RDKit5)}{65M}{Top15}    &  1 & .50 \\
    \modelid{USPTO}{Guide(RDKit5)}{65M}{Top30}    & 14 & .50 \\
    \modelid{USPTO}{Guide(RDKit5)}{65M}{Top45}    &  8 & .50 \\
    \modelid{USPTO}{Guide(RDKit5)}{65M}{Top60}    &  4 & .05 \\
    \bottomrule
  \end{tabular}
\end{table*}

\paragraph{Frozen inventory and sampling.}
The common test contains 1,201 cases, 1,280 paired semantic families, and
2,560 signed groups balanced over five RDKit axes and both polarities.
All arms use the shared sampler in Appendix~\ref{app:test-protocol}.
The ten-model, three-arm study contains 7,680,000 trajectories.
Table~\ref{tab:top100-common-support-cohort} gives construction-stage accounting.

\paragraph{Evaluation metrics.}
\label{app:evaluation-metrics}
For each product--instruction group, the raw budget $B$ counts every draw,
including invalid outputs. Hit$_{\mathrm{all}}@B$ is the rate of groups
whose first $B$ draws recover any candidate in the frozen inventory;
Hit$_+@B$ restricts this event to the preferred set. Reference coverage tables
abbreviate Hit$_{\mathrm{all}}@B$ as Hit@$B$; the proposal--selection
diagnostic uses Hit$_+@B$. Budget curves use $B\in\{1,5,10,25,50,100\}$.
For P@$k$ and N@$k$, we use all 100 raw draws per group, canonicalize and
deduplicate valid outputs, and rank the resulting precursor sets by
decreasing draw frequency. P@$k$ and N@$k$ are the rates of groups whose
top $k$ ranked candidates contain at least one preferred or avoided candidate,
respectively; they are set-hit rates, not the fraction of matching candidates
in the list.

Raw validity $V$ is the fraction of draws yielding a precursor set that
passes RDKit parsing and canonicalization. If $n_g$ draws are valid and
$u_g$ are distinct, the group's duplicate rate is $(n_g-u_g)/B$; unique-valid
count averages $u_g$ with the stated cohort weights. Pooled recovery counts distinct
product--candidate pairs recovered from the known inventory across all groups.
Validity concerns molecular representation, not experimental feasibility.

\paragraph{Contrasts and uncertainty.}
We report rates as percentages and their differences as percentage points;
counts retain their natural units. In three-arm comparisons, $\Delta$ subtracts
the named comparator from Matched: Shuffled tests instruction identity, while Reference measures
the net guidance effect. Positive $\Delta$P and negative $\Delta$N are
favorable; $-\Delta$N reverses the suppression sign for readability.
$\Delta V$ retains the signed validity difference. Aggregation and case-cluster
confidence intervals (CIs) follow Appendix~\ref{app:test-protocol};
Stage-B seeds 42--44 form a separate robustness analysis.

\paragraph{Data provenance and test overlap.}
The candidate audit confirms product-wise nesting of the synthetic supports.
We retain a 2.03\% Top60 catalog-version extension in the provenance
records. We inspected earlier Axis4 test reports during protocol
development. The named earlier cohort contains 988 products and shares
154 with the 1,201-product Common Top100 cohort (12.82\% of Common Top100).
Thus the later inventory and generated outputs are new, but the test products
are not wholly independent of that exploratory evaluation. We fixed the
reported model coordinates before inspecting the Common Top100 outputs.

\section{Common Top100 results and diagnostics}
\label{app:common-top100-complete}

A more diverse sampler need not recover known alternatives more often in
its first draw:
Top60 produces the most distinct valid outputs at $B=100$ but has the lowest
Hit@1 among synthetic-support models at each capacity
(Table~\ref{tab:reference-absorption-summary}).

\begin{table}[!htbp]
  \caption{Reference coverage at the endpoints of the Common Top100 budget
  curves. Hit, validity and duplicate rate are percentages; pooled recovery
  counts distinct case--candidate pairs. Rates and mean counts use equal-axis,
  case-balanced aggregation at both capacities. All columns except Hit@1 use
  $B=100$, without backfilling.}
  \label{tab:reference-absorption-summary}
  \centering
  \begin{tabular}{@{}lrrrrrr@{}}
    \toprule
    Support & Hit@1 & Hit@100 & Pooled & \shortstack{Unique valid\\/group}
      & Validity & \shortstack{Duplicate\\rate} \\
    \midrule
    \multicolumn{7}{@{}l}{\textbf{GraphDiT-8M}} \\
    USPTO-MR & 6.17 & 59.30 & 2,348 & 37.475 & 63.49 & 26.01 \\
    Top15 & 21.91 & 80.82 & 12,880 & 44.044 & 68.38 & 24.34 \\
    Top30 & 18.91 & 82.89 & 16,922 & 53.074 & 68.09 & 15.01 \\
    Top45 & 16.56 & 83.24 & 17,370 & 55.828 & 67.36 & 11.53 \\
    Top60 & 14.77 & 82.77 & 17,378 & 57.067 & 66.14 & 9.07 \\
    \midrule
    \multicolumn{7}{@{}l}{\textbf{GraphDiT-65M}} \\
    USPTO-MR & 5.08 & 52.93 & 1,723 & 38.938 & 63.47 & 24.53 \\
    Top15 & 25.74 & 81.48 & 13,661 & 40.836 & 69.56 & 28.72 \\
    Top30 & 24.02 & 84.96 & 20,075 & 49.615 & 69.21 & 19.59 \\
    Top45 & 21.84 & 84.18 & 21,562 & 54.159 & 69.18 & 15.02 \\
    Top60 & 16.48 & 83.16 & 19,241 & 56.956 & 67.10 & 10.14 \\
    \bottomrule
  \end{tabular}
\end{table}

Table~\ref{tab:common-top100-control-summary} retains the early- and
late-rank instruction-content contrasts. Top30 and Top45 show clear
preferred-route promotion and avoided-route suppression at both capacities;
65M Top60 remains statistically unresolved at these operating points.
For the primary 8M comparisons, Top30--Top45 paired intervals include zero,
whereas Top30 improves over Top15 and Top60 weakens relative to Top45.
Per-axis effects are not uniform, so the primary estimand gives each axis
equal weight. Matched--Shuffled validity intervals span zero across all
supports; at $B=100$, intervals for any-inventory availability and
unique-valid breadth also span zero for every expanded support.
Together, these results locate the instruction effect mainly in the
composition and ranking of sampled outputs.

\begin{table}[!htbp]
  \caption{Matched--Shuffled effects at returned ranks 1 and 10 on Common
  Top100. Entries are percentage-point estimates with 95\% paired
  case-cluster bootstrap intervals, using equal-axis, case-balanced
  aggregation. Positive $\Delta$P and negative $\Delta$N are favorable.
  The registered primary coordinates use Stage-B seed 42.}
  \label{tab:common-top100-control-summary}
  \centering
  \begin{tabular}{@{}lcccc@{}}
    \toprule
    Support & $\Delta$P@1 & $\Delta$P@10 & $\Delta$N@1 & $\Delta$N@10 \\
    \midrule
    \multicolumn{5}{@{}l}{\textbf{GraphDiT-8M}} \\
    USPTO-MR & \effectci{+0.08}{-0.66,+0.82} & \effectci{+2.03}{+1.05,+3.05} & \effectci{-0.74}{-1.48,0.00} & \effectci{-1.37}{-2.42,-0.35} \\
    Top15 & \effectci{+4.10}{+2.85,+5.31} & \effectci{+1.37}{+0.55,+2.19} & \effectci{-3.83}{-4.92,-2.70} & \effectci{-4.65}{-5.70,-3.59} \\
    Top30 & \effectci{+12.89}{+11.02,+14.77} & \effectci{+10.70}{+9.22,+12.19} & \effectci{-14.06}{-15.70,-12.42} & \effectci{-18.52}{-20.27,-16.80} \\
    Top45 & \effectci{+14.18}{+12.46,+15.94} & \effectci{+9.53}{+8.12,+10.94} & \effectci{-14.06}{-15.70,-12.42} & \effectci{-18.79}{-20.47,-17.11} \\
    Top60 & \effectci{+5.63}{+4.06,+7.15} & \effectci{+4.88}{+3.67,+6.13} & \effectci{-5.94}{-7.46,-4.45} & \effectci{-8.36}{-9.73,-6.95} \\
    \midrule
    \multicolumn{5}{@{}l}{\textbf{GraphDiT-65M}} \\
    USPTO-MR & \effectci{+0.00}{-0.63,+0.63} & \effectci{+0.23}{-0.63,+1.09} & \effectci{+0.20}{-0.39,+0.78} & \effectci{-1.60}{-2.46,-0.78} \\
    Top15 & \effectci{+11.29}{+9.49,+13.13} & \effectci{+7.50}{+6.21,+8.79} & \effectci{-12.66}{-14.30,-11.02} & \effectci{-13.83}{-15.31,-12.31} \\
    Top30 & \effectci{+18.52}{+16.56,+20.43} & \effectci{+12.19}{+10.78,+13.59} & \effectci{-17.89}{-19.61,-16.21} & \effectci{-27.19}{-28.91,-25.47} \\
    Top45 & \effectci{+17.30}{+15.39,+19.22} & \effectci{+11.99}{+10.47,+13.56} & \effectci{-17.11}{-18.79,-15.43} & \effectci{-25.86}{-27.62,-24.06} \\
    Top60 & \effectci{+1.09}{-0.23,+2.42} & \effectci{+0.35}{-0.70,+1.45} & \effectci{-0.39}{-1.72,+0.94} & \effectci{-0.63}{-1.76,+0.55} \\
    \bottomrule
  \end{tabular}
\end{table}

\subsection{Absolute arm results}
\label{app:absolute-results}

The absolute arms in Table~\ref{tab:common-top100-absolute-results} expose
costs shared by Matched and Shuffled: similar validity between those arms
can coexist with a substantial loss relative to Reference. Entries use the
primary equal-axis, case-balanced aggregation and the sampling protocol in
Appendix~\ref{app:test-protocol}. Native Structure results appear in
Appendix~\ref{app:structure-embedding-ablation}; Metric4 results appear in
Appendix~\ref{app:metric4-generalization}.

\begin{table}[!htbp]
  \caption{Key absolute-arm Common Top100 results (\%) at the coordinates
  in Table~\ref{tab:common-top100-coordinates}. Each support lists all three
  conditions at the same raw-draw budget $B=100$.}
  \label{tab:common-top100-absolute-results}
  \centering
  \begin{tabular}{@{}llrrrrr@{}}
    \toprule
    Support & Arm & P@1 & P@10 & N@1 & N@10 & Validity \\
    \midrule
    \multicolumn{7}{@{}l}{\textbf{GraphDiT-8M}} \\
    USPTO-MR & Reference & 8.477 & 25.547 & 8.555 & 25.391 & 63.488 \\
     & Matched & 8.984 & 27.461 & 7.695 & 22.852 & 60.816 \\
     & Shuffled & 8.906 & 25.430 & 8.438 & 24.219 & 60.913 \\
    \midrule
    Top15 & Reference & 30.156 & 58.789 & 30.117 & 58.242 & 68.382 \\
     & Matched & 34.102 & 59.062 & 26.055 & 53.086 & 67.084 \\
     & Shuffled & 30.000 & 57.695 & 29.883 & 57.734 & 67.148 \\
    \midrule
    Top30 & Reference & 30.195 & 58.281 & 29.844 & 56.484 & 68.086 \\
     & Matched & 40.195 & 57.617 & 11.602 & 28.750 & 52.605 \\
     & Shuffled & 27.305 & 46.914 & 25.664 & 47.266 & 52.712 \\
    \midrule
    Top45 & Reference & 29.844 & 56.328 & 28.789 & 55.664 & 67.356 \\
     & Matched & 40.625 & 58.945 & 13.594 & 30.742 & 62.611 \\
     & Shuffled & 26.445 & 49.414 & 27.656 & 49.531 & 62.457 \\
    \midrule
    Top60 & Reference & 29.219 & 55.312 & 27.539 & 53.867 & 66.138 \\
     & Matched & 33.047 & 56.484 & 21.992 & 43.789 & 64.696 \\
     & Shuffled & 27.422 & 51.602 & 27.930 & 52.148 & 64.700 \\
    \midrule
    \multicolumn{7}{@{}l}{\textbf{GraphDiT-65M}} \\
    USPTO-MR & Reference & 7.695 & 21.562 & 7.422 & 21.602 & 63.470 \\
     & Matched & 8.164 & 23.867 & 7.500 & 20.898 & 63.163 \\
     & Shuffled & 8.164 & 23.633 & 7.305 & 22.500 & 63.104 \\
    \midrule
    Top15 & Reference & 30.234 & 61.055 & 31.406 & 61.367 & 69.555 \\
     & Matched & 41.250 & 63.477 & 16.406 & 41.797 & 60.367 \\
     & Shuffled & 29.961 & 55.977 & 29.063 & 55.625 & 60.230 \\
    \midrule
    Top30 & Reference & 32.930 & 63.281 & 29.727 & 62.734 & 69.209 \\
     & Matched & 48.828 & 67.852 & 13.164 & 28.164 & 68.012 \\
     & Shuffled & 30.312 & 55.664 & 31.055 & 55.352 & 67.897 \\
    \midrule
    Top45 & Reference & 30.586 & 61.211 & 30.273 & 60.234 & 69.182 \\
     & Matched & 45.742 & 65.000 & 11.953 & 27.266 & 65.178 \\
     & Shuffled & 28.438 & 53.008 & 29.063 & 53.125 & 65.086 \\
    \midrule
    Top60 & Reference & 30.195 & 55.586 & 27.695 & 55.391 & 67.099 \\
     & Matched & 29.141 & 55.156 & 27.930 & 53.672 & 63.466 \\
     & Shuffled & 28.047 & 54.805 & 28.320 & 54.297 & 63.362 \\
    \bottomrule
  \end{tabular}
\end{table}

\subsection{Returned-list contrasts with the unguided model}
\label{app:unguided-topk}
\label{app:topk-reference-effects}

Table~\ref{tab:common-top100-unguided-contrasts} expands the
Matched--Reference columns of Table~\ref{tab:main-coverage-control} with
paired intervals at the frozen primary coordinates, using the aggregation
and sampling protocol in Appendix~\ref{app:test-protocol}. Early-rank gains
need not extend to Top-10: guidance can move an already available preferred
route to the front. Avoided-route suppression can remain strong across the
list even when later-rank promotion is small.

\begin{table}[!htbp]
  \caption{\textbf{Returned Top-$k$ changes relative to unguided generation.}
  Cells give Matched--Reference effects in percentage points with 95\%
  paired case-bootstrap intervals. We reverse the sign of avoided-set recovery;
  larger values are favorable in every column.  Both capacities show
  early-rank preferred gains at Top15/30/45, while higher-rank effects vary.}
  \label{tab:common-top100-unguided-contrasts}
  \centering
  \begin{tabular}{lccc}
    \toprule
    Support & $\Delta$P@1 & $\Delta$P@10 & $-\Delta$N@10 \\
    \midrule
    \multicolumn{4}{l}{\textbf{GraphDiT-8M}} \\
    USPTO-MR & \effectci{+0.51}{-0.23,+1.25}
      & \effectci{+1.91}{+0.86,+2.97} & \effectci{+2.54}{+1.45,+3.63} \\[2pt]
    \midrule
    Top15 & \effectci{+3.95}{+2.66,+5.16}
      & \effectci{+0.27}{-0.55,+1.09} & \effectci{+5.16}{+4.14,+6.17} \\[2pt]
    Top30 & \effectci{+10.00}{+8.09,+11.87}
      & \effectci{-0.66}{-2.03,+0.74} & \effectci{+27.73}{+25.90,+29.53} \\[2pt]
    Top45 & \effectci{+10.78}{+9.06,+12.50}
      & \effectci{+2.62}{+1.33,+3.91} & \effectci{+24.92}{+23.24,+26.72} \\[2pt]
    Top60 & \effectci{+3.83}{+2.27,+5.39}
      & \effectci{+1.17}{-0.08,+2.42} & \effectci{+10.08}{+8.71,+11.48} \\[2pt]
    \midrule
    \multicolumn{4}{l}{\textbf{GraphDiT-65M}} \\
    USPTO-MR & \effectci{+0.47}{-0.27,+1.25}
      & \effectci{+2.30}{+1.25,+3.40} & \effectci{+0.70}{-0.31,+1.72} \\[2pt]
    \midrule
    Top15 & \effectci{+11.02}{+9.41,+12.66}
      & \effectci{+2.42}{+1.29,+3.52} & \effectci{+19.57}{+18.01,+21.09} \\[2pt]
    Top30 & \effectci{+15.90}{+14.14,+17.66}
      & \effectci{+4.57}{+3.40,+5.78} & \effectci{+34.57}{+32.77,+36.37} \\[2pt]
    Top45 & \effectci{+15.16}{+13.20,+17.11}
      & \effectci{+3.79}{+2.38,+5.16} & \effectci{+32.97}{+31.09,+34.80} \\[2pt]
    Top60 & \effectci{-1.05}{-2.50,+0.39}
      & \effectci{-0.43}{-1.64,+0.78} & \effectci{+1.72}{+0.47,+2.93} \\[2pt]
    \bottomrule
  \end{tabular}
\end{table}


\subsection{Common-update-ceiling support sensitivity}
\label{app:compute-matched-support-scaling}

To examine sensitivity to support-dependent training budgets, we retrained
Top15, Top30, and Top45 with common Stage-A/B training ceilings
of 125,850/58,730 optimizer updates (effective batch 32, seed 42).
We selected each deployment checkpoint and residual scale on validation
before inspecting the corresponding Common Top100 outputs. Top60 reuses its
primary checkpoint.  This sensitivity equalizes adapter-training ceilings,
not the selected update counts or the costs of base training and historical
reaction-pair pretraining.

\begin{figure*}[!htbp]
  \centering
  \includegraphics[width=0.96\textwidth]{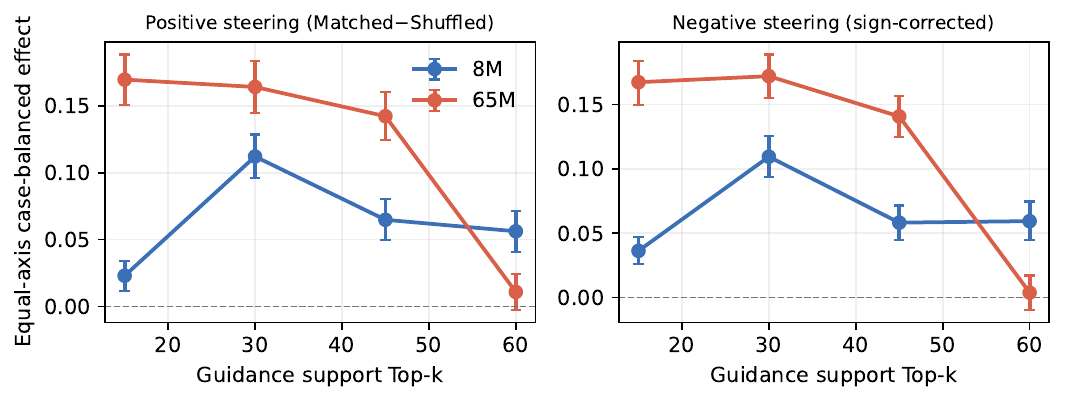}
  \caption{\textbf{Common-update-ceiling support-scaling sensitivity.}  Points are
  equal-axis, case-balanced Matched--Shuffled effects on the frozen Common
  Top100 test, shown as probability differences ($0.10=10$ percentage points).
  Error bars are 95\% shared case-cluster bootstrap intervals.
  The right panel plots $-\Delta\mathrm{N}@1$ so that larger values are
  favorable.  Top15/30/45 use the common update ceilings above;
  Top60 reuses its primary checkpoint.}
  \label{fig:compute-matched-support-scaling}
\end{figure*}

\begin{table*}[!htbp]
  \caption{Common-update-ceiling deployment coordinates and effects.
  Selected B gives optimizer updates (Top60-equivalent epoch index);
  \emph{reused} denotes the primary Top60 coordinates in
  Table~\ref{tab:common-top100-coordinates}.
  Effects are Matched minus Shuffled in percentage points [95\% shared
  case-cluster CI]; larger $\Delta$P@1 and smaller $\Delta$N@1 are favorable.
  $V$ denotes raw-draw validity.}
  \label{tab:compute-matched-effects}
  \label{tab:compute-matched-coordinates}
  \centering
  \begin{tabular}{@{}lcrccc@{}}
    \toprule
    Support & \shortstack{Selected B\\(coord.)} & $\alpha$
      & $\Delta$P@1 & $\Delta$N@1 & $\Delta V$ \\
    \midrule
    \multicolumn{6}{@{}l}{\textbf{8M}} \\
    Top15 & \shortstack{58,730\\(14)} & $.15$
      & \effectci{+2.30}{+1.17,+3.44}
      & \effectci{-3.63}{-4.73,-2.62}
      & \effectci{+0.07}{-0.06,+0.20} \\
    Top30 & \shortstack{33,560\\(8)} & $.50$
      & \effectci{+11.21}{+9.61,+12.85}
      & \effectci{-10.94}{-12.54,-9.34}
      & \effectci{-0.05}{-0.30,+0.20} \\
    Top45 & \shortstack{50,340\\(12)} & $.25$
      & \effectci{+6.48}{+4.96,+8.01}
      & \effectci{-5.82}{-7.15,-4.45}
      & \effectci{+0.07}{-0.11,+0.26} \\
    Top60 & reused & $.25$
      & \effectci{+5.62}{+4.06,+7.15}
      & \effectci{-5.94}{-7.46,-4.45}
      & \effectci{-0.00}{-0.21,+0.20} \\
    \midrule
    \multicolumn{6}{@{}l}{\textbf{65M}} \\
    Top15 & \shortstack{8,390\\(2)} & $.50$
      & \effectci{+16.95}{+15.08,+18.83}
      & \effectci{-16.72}{-18.40,-14.96}
      & \effectci{+0.10}{-0.26,+0.46} \\
    Top30 & \shortstack{58,730\\(14)} & $.50$
      & \effectci{+16.41}{+14.49,+18.32}
      & \effectci{-17.19}{-18.87,-15.51}
      & \effectci{+0.16}{-0.18,+0.49} \\
    Top45 & \shortstack{8,390\\(2)} & $.50$
      & \effectci{+14.22}{+12.42,+16.05}
      & \effectci{-14.06}{-15.66,-12.46}
      & \effectci{+0.09}{-0.30,+0.49} \\
    Top60 & reused & $.05$
      & \effectci{+1.09}{-0.23,+2.42}
      & \effectci{-0.39}{-1.72,+0.94}
      & \effectci{+0.10}{-0.06,+0.27} \\
    \bottomrule
  \end{tabular}
\end{table*}

Positive control survives the common training ceilings
(Table~\ref{tab:compute-matched-effects}), but the 8M peak shifts to Top30.
Support choice therefore depends on training and selection; candidate count
alone does not determine control strength.
Reference and Top60 use the primary
coordinates summarized in Table~\ref{tab:common-top100-absolute-results}.

\subsection{Relaxed round-trip coverage and accuracy across candidate ranks}
\label{app:reactiont5-relaxed-roundtrip}

This auxiliary analysis evaluates the Top30 65M reference, matched-guidance,
and shuffled-guidance arms on the same 200 frozen test groups.  The guided
arms use the validation-selected epoch-14 checkpoint with $\alpha=0.5$.
For each group and arm, we reuse 100 reverse samples generated with 50 sampling
steps, canonicalize and deduplicate valid precursor sets, and rank them by
sample frequency, breaking ties by canonical SMILES.  We evaluate the first
$K\in\{1,3,5,10\}$ available reverse candidates with a fixed ReactionT5
forward model \citep{sagawa2025reactiont5}, using our locally fine-tuned
checkpoint.

For candidate precursor set $r_{g,i}$ in group $g$, let $x_g$ be the target
product and $\mathcal R_g^{\mathrm{ref}}$ the frozen collection of canonical
positive-reference precursor sets associated with that semantic group.
Let $F_1(r)$ be the top-ranked product from the fixed five-beam forward
decoder.  We accept only this Forward Top-1 product, so the relaxed indicator
for an available reverse candidate is
\begin{equation}
  \chi_{g,i}
  =\mathbf{1}\!\left[
    F_1(r_{g,i})=x_g
    \ \lor\ r_{g,i}\in\mathcal R_g^{\mathrm{ref}}
  \right].
  \label{eq:relaxed-roundtrip-pooled}
\end{equation}
We use the same forward checkpoint, empty-reagent interface, and deterministic
\texttt{num\_beams=5}, \texttt{num\_return\_sequences=5} decoding throughout,
and ignore beams below rank one. We fix the reverse-candidate ranking
before forward evaluation.

For each arm, let $n_g(K)$ be the number of available candidates among the
first $K$ ranks in group $g$, and define $z_{g,i}=\chi_{g,i}$ when rank $i$
is available and $z_{g,i}=0$ otherwise.  Table~\ref{tab:reactiont5-relaxed-pooled}
reports fixed-slot RT-Accuracy and group-level RT-Coverage across the $G=200$
groups:
\begin{equation}
  A_{\mathrm{RT}}^{\mathrm{slot}}(K)
  =100\,\frac{\sum_{g=1}^{G}\sum_{i=1}^{K}z_{g,i}}{GK},
  \qquad
  C_{\mathrm{RT}}(K)
  =100\,\frac{\sum_{g=1}^{G}\mathbf{1}[\sum_{i=1}^{K}z_{g,i}>0]}{G}.
  \label{eq:relaxed-roundtrip-accuracy}
\end{equation}
We count missing or invalid candidate ranks as failures and keep them in
the accuracy denominator. At $K=1$, RT-Accuracy and
RT-Coverage are necessarily identical.
Every entry retains the reference-precursor shortcut in
Eq.~(\ref{eq:relaxed-roundtrip-pooled}).

\begin{table*}[!htbp]
  \caption{Relaxed Forward Top-1 round-trip results (\%) for Top30 65M on
  200 frozen test groups.  Reverse $K$ is the candidate-rank cutoff.
  Coverage is the fraction of groups with at least one round-trip hit;
  Accuracy uses the fixed $200K$ candidate-slot denominator, counting missing
  or invalid ranks as failures.  Guidance denotes matched guidance, and
  Shuffled is the shuffled-guidance control.  $\Delta$ is Guidance minus
  Reference in percentage points.  Bold marks the highest point estimate for
  each metric and row.  All results include the reference shortcut in
  Eq.~(\ref{eq:relaxed-roundtrip-pooled}).}
  \label{tab:reactiont5-relaxed-pooled}
  \centering
  \setlength{\tabcolsep}{3.5pt}
  \begin{tabular}{r rr rr rr rr}
    \toprule
    & \multicolumn{2}{c}{Reference}
    & \multicolumn{2}{c}{Guidance}
    & \multicolumn{2}{c}{Shuffled}
    & \multicolumn{2}{c}{$\Delta$ (pp)} \\
    \cline{2-3}\cline{4-5}\cline{6-7}\cline{8-9}
    Reverse $K$ & Cov. & Acc. & Cov. & Acc. & Cov. & Acc. & Cov. & Acc. \\
    \midrule
    1  & 63.00 & 63.00 & \textbf{71.00} & \textbf{71.00}
       & 57.50 & 57.50 & $+8.00$ & $+8.00$ \\
    3  & \textbf{76.50} & 59.50 & 75.00 & \textbf{62.83}
       & 73.50 & 56.83 & $-1.50$ & $+3.33$ \\
    5  & \textbf{79.00} & \textbf{57.90} & 76.50 & \textbf{57.90}
       & 76.00 & 52.60 & $-2.50$ & $0.00$ \\
    10 & \textbf{80.00} & \textbf{51.25} & \textbf{80.00} & 47.85
       & 77.00 & 42.90 & $0.00$ & $-3.40$ \\
    \bottomrule
  \end{tabular}
\end{table*}


Guidance concentrates relaxed round-trip hits at the first returned candidate:
the $K=1$ gain has a paired stratified bootstrap 95\% interval of
$[3.00,13.50]$ points (10,000 resamples; seed 1042). This advantage fades
deeper in the list. At $K=10$, equal group coverage but lower fixed-slot
accuracy means that guidance reaches as many groups with a hit while
returning fewer successful candidates per group. This interpretation remains
conditional on the relaxed rule, whose reference-precursor shortcut can
accept a candidate without forward-model closure.


\section{Calibration and Stage-B seed robustness}
\label{app:sensitivity-diagnostics}

We examine calibration at fixed checkpoints
(Appendices~\ref{app:validation-alpha-tradeoff}--\ref{app:top45-top60-alpha-pareto})
and report the full support--capacity matrix across Stage-B seeds
(Appendix~\ref{app:all-support-seed-robustness}).
Supplementary Section~S3 records the selection rules and validation
constraints; Appendix~\ref{app:common-top100-protocol} lists
the frozen evaluation coordinates.

\subsection{Validation guidance-strength trade-off}
\label{app:validation-alpha-tradeoff}

\begin{figure}[!htbp]
  \centering
  \includegraphics[width=0.56\textwidth]{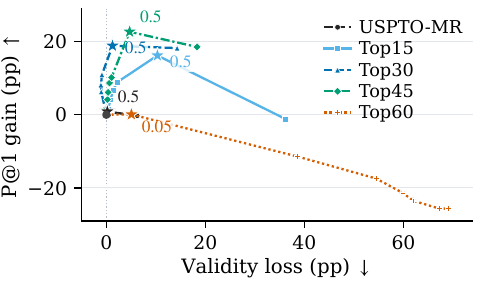}
  \caption{\textbf{Guidance-strength trade-off at fixed checkpoints.}
  GraphDiT-65M on each support's original validation cohort at epoch~14.
  P@1 gain and raw-validity loss are relative to $\alpha=0$.
  Stars mark final selected scales evaluated at this checkpoint;
  deployed checkpoints may differ.}
  \label{fig:alpha-tradeoff}
\end{figure}

At inference, $\alpha=\alpha_{\mathrm{eval}}$ scales the instruction residual
applied to the frozen backbone; $\alpha=0$ recovers the unguided Base model.
Stronger intervention can favor preferred routes while disrupting valid
generation (Figure~\ref{fig:alpha-tradeoff}), motivating calibration using
both preference recovery and output validity.

These 65M sweeps use each support's original epoch-14 validation cohort
and aggregates. Stars annotate final deployment scales evaluated here at
epoch~14; deployed checkpoints may differ. The curves characterize each
model's sensitivity to guidance strength, but do not isolate a causal effect
of training support. They are separate from the common-inventory test
diagnostics below.

\subsection{Matched residual-scale sensitivity}
\label{app:matched-alpha}

We compare all five 8M supports at $\alpha_{\mathrm{eval}}=.25$, retaining
their seed-42 validation-selected checkpoints. This controls inference
strength while preserving the support-specific trained models.

\begin{table*}[!htbp]
  \caption{Matched-scale 8M sensitivity at $\alpha_{\mathrm{eval}}=.25$.
  Values are equal-axis, case-balanced Matched--Shuffled effects in
  percentage points. Positive $\Delta\mathrm{P}$ and negative
  $\Delta\mathrm{N}$ indicate better instruction alignment;
  $\Delta V$ is the Matched--Shuffled trajectory-validity difference.}
  \label{tab:matched-alpha-results}
  \centering
  \begin{tabular}{lrrrrr}
    \toprule
    Support & $\Delta$P@1 & $\Delta$P@10 & $\Delta$N@1
      & $\Delta$N@10 & $\Delta V$ \\
    \midrule
    USPTO-MR & +0.04 & +0.63 & -0.16 & -1.25 & -0.04 \\
    Top15    & +4.73 & +2.27 & -4.69 & -5.00 & -0.05 \\
    Top30    & +4.80 & +3.28 & -4.22 & -8.67 & -0.08 \\
    Top45    & +6.60 & +4.41 & -6.37 & -9.61 & +0.01 \\
    Top60    & +5.63 & +4.88 & -5.94 & -8.36 & +0.00 \\
    \bottomrule
  \end{tabular}
\end{table*}

Matching $\alpha$ narrows the primary Top45--Top60 contrast: their paired
95\% intervals include zero at all four recovery endpoints. The selected
support ranking therefore depends partly on deployment strength; this
comparison alone does not establish an intrinsic advantage for either support.

\subsection{Three-seed frozen-checkpoint inference-scale response}
\label{app:top45-top60-alpha-pareto}

For 8M Top45 and Top60, we hold each seed's validation-selected checkpoint
fixed and vary $\alpha\in\{0,.05,.10,.15,.20,.25,.50,1\}$, using seeds
42--44 and the paired test schedule from
Appendix~\ref{app:common-top100-protocol}. The grid is a post-hoc diagnostic
and does not revise validation-selected operating points. Its archived
estimand weights distinct cases equally after averaging their signed groups;
the selected-point tables instead give each axis equal weight.

\begin{figure*}[!htbp]
  \centering
  \includegraphics[width=\textwidth]{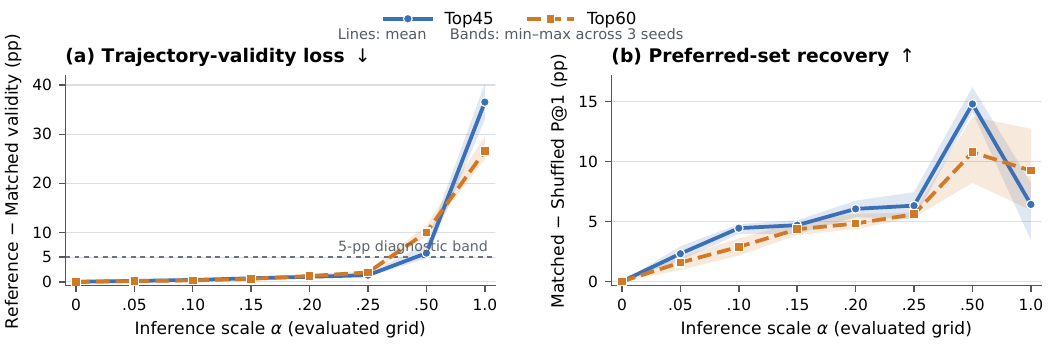}
  \caption{\textbf{Fixed-checkpoint scale response across three Stage-B seeds.}
  GraphDiT-8M Top45 and Top60: (a) Reference-minus-Matched trajectory-validity
  loss; (b) Matched--Shuffled P@1 gain, both in percentage points.
  Lines are seed means and bands span the seedwise minimum and maximum,
  not confidence intervals. The dashed five-point line is a diagnostic
  validity-loss reference.}
  \label{fig:top45-top60-alpha-pareto}
\end{figure*}

Weak guidance yields similar responses. Raising $\alpha$ from $.25$ to $.50$
costs Top60 more validity without a larger P@1 gain, indicating a narrower
calibration margin at these checkpoints (Figure~\ref{fig:top45-top60-alpha-pareto}).
Both supports lose substantial validity under strong guidance.

\subsection{Complete three-seed analysis across supports and capacities}
\label{app:all-support-seed-robustness}

\paragraph{Protocol and scope.}
All ten capacity--support settings in Table~\ref{tab:main-coverage-control}
have Stage-B seeds 42, 43, and 44. We fix the base generator, Stage-A projector,
training corpus, and paired test sampling schedule; each run
retains its own archived validation-selected epoch and $\alpha$ under the
pipeline's recorded rule, with no test-based reselection.
Table~\ref{tab:all-support-seed-summary} uses the same equal-axis,
case-balanced aggregation as Table~\ref{tab:main-coverage-control}.
Sample SDs describe conditional Stage-B training-and-selection variation;
they are not confidence intervals or estimates of full-pipeline uncertainty.
Validity loss is $L_{\mathrm{val}}=V_R-V_M$, where $R$ and $M$ denote
Reference and Matched; positive values indicate a guidance cost.

\begin{table*}[!htbp]
  \caption{All-support Stage-B replication of Table~\ref{tab:main-coverage-control}, using seeds 42--44. Cells show the mean above $\pm$ sample SD ($n=3$, denominator $n-1$), in percentage points. All alignment contrasts are higher-is-better; validity loss $L_{\mathrm{val}}=V_R-V_M$ is lower-is-better. $^{\mathrm F}$ marks 65M Top15, where all three runs use fallback selections that failed the original validation hard gate. Reference is fixed across Stage-B seeds.}
  \label{tab:all-support-seed-summary}
  \centering
  \renewcommand{\papertablestyle}{\normalfont\normalsize\setlength{\tabcolsep}{3pt}\renewcommand{\arraystretch}{1.08}}
  \begin{tabular}{@{}lrrrrrr@{}}
    \toprule
    & \multicolumn{2}{c}{Matched--Shuffled} & \multicolumn{3}{c}{Matched--Reference} & Validity \\
    \cline{2-3}\cline{4-6}
    Support & $\Delta\mathrm{P}@1$ & $-\Delta\mathrm{N}@10$ & $\Delta\mathrm{P}@1$ & $\Delta\mathrm{P}@10$ & $-\Delta\mathrm{N}@10$ & $L_{\mathrm{val}}$ \\
    \midrule
    \multicolumn{7}{@{}l}{\textbf{GraphDiT-8M}} \\
    USPTO-MR & \shortstack{$0.14$\\[-1pt]{$\pm 0.18$}} & \shortstack{$1.35$\\[-1pt]{$\pm 0.92$}} & \shortstack{$0.40$\\[-1pt]{$\pm 0.40$}} & \shortstack{$1.16$\\[-1pt]{$\pm 0.93$}} & \shortstack{$1.74$\\[-1pt]{$\pm 1.44$}} & \shortstack{$2.08$\\[-1pt]{$\pm 1.59$}} \\
    Top15 & \shortstack{$4.54$\\[-1pt]{$\pm 0.61$}} & \shortstack{$5.46$\\[-1pt]{$\pm 1.78$}} & \shortstack{$4.30$\\[-1pt]{$\pm 0.58$}} & \shortstack{$0.69$\\[-1pt]{$\pm 0.36$}} & \shortstack{$6.41$\\[-1pt]{$\pm 1.84$}} & \shortstack{$2.14$\\[-1pt]{$\pm 0.81$}} \\
    Top30 & \shortstack{$13.16$\\[-1pt]{$\pm 1.02$}} & \shortstack{$18.09$\\[-1pt]{$\pm 2.01$}} & \shortstack{$10.70$\\[-1pt]{$\pm 1.29$}} & \shortstack{$0.26$\\[-1pt]{$\pm 1.16$}} & \shortstack{$25.81$\\[-1pt]{$\pm 2.42$}} & \shortstack{$12.85$\\[-1pt]{$\pm 2.53$}} \\
    Top45 & \shortstack{$8.96$\\[-1pt]{$\pm 4.64$}} & \shortstack{$12.55$\\[-1pt]{$\pm 5.70$}} & \shortstack{$7.30$\\[-1pt]{$\pm 3.24$}} & \shortstack{$2.96$\\[-1pt]{$\pm 0.95$}} & \shortstack{$14.82$\\[-1pt]{$\pm 8.93$}} & \shortstack{$2.46$\\[-1pt]{$\pm 1.99$}} \\
    Top60 & \shortstack{$5.12$\\[-1pt]{$\pm 0.81$}} & \shortstack{$7.93$\\[-1pt]{$\pm 2.01$}} & \shortstack{$3.74$\\[-1pt]{$\pm 0.19$}} & \shortstack{$1.09$\\[-1pt]{$\pm 0.28$}} & \shortstack{$9.10$\\[-1pt]{$\pm 2.36$}} & \shortstack{$1.63$\\[-1pt]{$\pm 0.31$}} \\
    \midrule
    \multicolumn{7}{@{}l}{\textbf{GraphDiT-65M}} \\
    USPTO-MR & \shortstack{$-0.07$\\[-1pt]{$\pm 0.06$}} & \shortstack{$1.25$\\[-1pt]{$\pm 0.40$}} & \shortstack{$0.43$\\[-1pt]{$\pm 0.22$}} & \shortstack{$2.04$\\[-1pt]{$\pm 0.78$}} & \shortstack{$0.25$\\[-1pt]{$\pm 0.57$}} & \shortstack{$-0.02$\\[-1pt]{$\pm 0.35$}} \\
    Top15$^{\mathrm F}$ & \shortstack{$14.05$\\[-1pt]{$\pm 2.52$}} & \shortstack{$17.81$\\[-1pt]{$\pm 3.91$}} & \shortstack{$13.13$\\[-1pt]{$\pm 1.99$}} & \shortstack{$3.19$\\[-1pt]{$\pm 0.73$}} & \shortstack{$23.32$\\[-1pt]{$\pm 4.07$}} & \shortstack{$8.95$\\[-1pt]{$\pm 0.85$}} \\
    Top30 & \shortstack{$17.50$\\[-1pt]{$\pm 1.20$}} & \shortstack{$26.90$\\[-1pt]{$\pm 1.05$}} & \shortstack{$15.57$\\[-1pt]{$\pm 0.44$}} & \shortstack{$5.00$\\[-1pt]{$\pm 0.41$}} & \shortstack{$33.49$\\[-1pt]{$\pm 2.11$}} & \shortstack{$1.76$\\[-1pt]{$\pm 0.82$}} \\
    Top45 & \shortstack{$17.47$\\[-1pt]{$\pm 1.16$}} & \shortstack{$27.50$\\[-1pt]{$\pm 1.43$}} & \shortstack{$15.53$\\[-1pt]{$\pm 0.69$}} & \shortstack{$4.57$\\[-1pt]{$\pm 0.69$}} & \shortstack{$34.78$\\[-1pt]{$\pm 1.58$}} & \shortstack{$3.90$\\[-1pt]{$\pm 0.20$}} \\
    Top60 & \shortstack{$0.35$\\[-1pt]{$\pm 0.91$}} & \shortstack{$0.10$\\[-1pt]{$\pm 0.60$}} & \shortstack{$-1.65$\\[-1pt]{$\pm 0.67$}} & \shortstack{$-0.25$\\[-1pt]{$\pm 0.53$}} & \shortstack{$0.86$\\[-1pt]{$\pm 0.75$}} & \shortstack{$2.37$\\[-1pt]{$\pm 1.67$}} \\
    \bottomrule
  \end{tabular}
\end{table*}

\paragraph{Control across supports.}
At 8M, all synthetic supports improve P@1 against both Shuffled and Reference
in all three seeds, but Top45's primary maximum varies with training and
scale selection. At 65M, Top30/Top45 retain strong gains, whereas Top60's
net P@1 effect is negative in every seed. Thus the repeats support reliable
control at intermediate supports without fixing a universal best support.
The 8M Top30 results also distinguish moving preferred routes to the front
from improving their recovery anywhere in the Top-10. USPTO-MR retains
near-zero P@1 instruction-content effects at both capacities.

\paragraph{Validity costs.}
Stable rank gains can still be costly: 8M Top30 loses substantially more
validity than Top45. All three 65M Top15 runs use the archived fallback rule
after failing the original validation hard gate, and lose more validity
than Top30/Top45. The support comparisons therefore reflect different
validity costs and selection rules, not a shared validity budget.

\section{Controlled method alternatives}
\label{app:controlled-training-ablations}

\subsection{Direct Stage-B: effect of Stage-A-aligned initialization}
\label{app:stagea-initialization-ablation}

We isolate the contribution of Stage-A-aligned initialization.  RIGS (called
Joint in the original logs) copies the validation-selected epoch-26 Stage-A
projector into each Stage-B projector group; Direct Stage-B (Scratch-B in the
logs) uses the same constructor and seed at epoch~0, before any Stage-A update.
The arms otherwise share the base checkpoint, adapter topology, trainable
scope, Stage-B data and objective, optimizer, update budget, checkpoint
coordinates, cohorts, and sampling protocol.  The separately reported exact
cosine and scalar-head controls in
Appendix~\ref{app:budgeted-proposal-selection} instead test frozen-output
selection, not Stage-B initialization.

Table~\ref{tab:stagea-initialization-validation-summary} retains the matched
validation coordinates needed to interpret the ablation.  Differences are
RIGS minus Direct Stage-B in percentage points.  For 8M, both variants select
$\alpha_{\mathrm{eval}}=.50$ and share the P@1-selected epoch~12 coordinate.
For 65M, we fix epoch~14 and compare both variants at the RIGS operating scale
$\alpha=.50$; comparing their independently selected scales would confound
initialization with deployment selection.

\begin{table*}[!htbp]
  \caption{Matched-validation summary for the Stage-A initialization
  ablation.  $\Delta$ is RIGS minus Direct Stage-B in percentage points.
  Larger $\Delta$P@$k$, smaller $\Delta$N@$k$, and positive $\Delta$Validity
  favor RIGS.  The 65M row is a matched-scale validation comparison, not a
  test result.}
  \label{tab:stagea-initialization-validation-summary}
  \centering
  \begin{tabular}{lcrrrrr}
    \toprule
    Capacity & $(e,\alpha_{\mathrm{eval}})$ & $\Delta$P@1 & $\Delta$P@10
      & $\Delta$N@1 & $\Delta$N@10 & $\Delta$Validity \\
    \midrule
    8M  & $(12,.50)$ & $+1.25$ & $+1.48$ & $-2.19$ & $-5.39$ & $+1.12$ \\
    65M & $(14,.50)$ & $+1.17$ & $+0.86$ & $-0.63$ & $-3.28$ & $-1.16$ \\
    \bottomrule
  \end{tabular}
\end{table*}

Avoided-route suppression is the most consistent benefit in the 8M
validation sweep: N@10 improves at every checkpoint, whereas preferred-route
recovery does not improve uniformly across ranks. The 65M matched-scale
sweep also strengthens control, with a modest validity cost. These
single-seed diagnostics motivate checking whether the suppression advantage
persists on held-out data.

We additionally evaluate the validation-frozen epoch-12/$\alpha=.50$ RIGS and
Direct Stage-B checkpoints on exactly the same frozen test cohort and sampling
schedule.  The cohort contains 1,185 cases and 2,560 signed groups, with 100
trajectories per condition.  Here ``Top60'' denotes the frozen evaluation
candidate upper bound; we train both models with the Top30 RDKit5
recipe. For each model, the guidance effect is Matched minus Shuffled, and the
difference-in-differences (DiD) is Direct Stage-B effect minus RIGS effect.

\begin{table*}[!htbp]
  \caption{Frozen-test Stage-A initialization difference-in-differences for
  \modelid{USPTO}{Guide(RDKit5)}{8M}{Top30}.  RIGS and Direct Stage-B use the same
epoch-12 checkpoint coordinate, $\alpha_{\mathrm{eval}}=.50$, test cohort,
and sampling schedule; M/S denote Matched/Shuffled.  Absolute arms are
  case-balanced percentages; effects and DiD are percentage points.  Intervals
  use 10,000 paired case-cluster
  bootstrap replicates.  For P@$k$ and validity, a negative Direct-minus-RIGS
  DiD favors RIGS; for N@$k$, a positive DiD favors RIGS because lower recovery
  is better.}
  \label{tab:stagea-initialization-frozen-test-did}
  \centering
  \begin{tabular}{lrrrrrrr}
    \toprule
    Metric & \shortstack{RIGS\\M} & \shortstack{RIGS\\S}
      & \shortstack{Direct\\M} & \shortstack{Direct\\S}
      & \shortstack{RIGS\\M$-$S} & \shortstack{Direct\\M$-$S}
      & \shortstack{Direct$-$RIGS\\DiD [95\% CI]} \\
    \midrule
    P@1  & 24.80 & 12.03 & 23.26 & 11.31 & $+12.77$ & $+11.96$ & $-0.82$ [$-3.47,+1.89$] \\
    P@3  & 37.09 & 21.56 & 34.70 & 21.27 & $+15.53$ & $+13.43$ & $-2.10$ [$-4.90,+0.70$] \\
    P@5  & 39.83 & 25.11 & 37.74 & 25.36 & $+14.73$ & $+12.38$ & $-2.35$ [$-5.01,+0.37$] \\
    P@10 & 42.60 & 29.37 & 41.39 & 28.73 & $+13.24$ & $+12.66$ & $-0.58$ [$-3.25,+2.14$] \\
    N@1  &  2.93 & 13.62 &  5.08 & 12.74 & $-10.69$ & $-7.67$ & \textbf{$+3.02$ [$+0.89,+5.19$]} \\
    N@3  &  5.46 & 22.42 &  9.76 & 22.49 & $-16.96$ & $-12.73$ & \textbf{$+4.23$ [$+1.91,+6.62$]} \\
    N@5  &  6.68 & 26.50 & 12.24 & 26.33 & $-19.82$ & $-14.09$ & \textbf{$+5.72$ [$+3.39,+8.19$]} \\
    N@10 &  9.00 & 30.21 & 15.61 & 30.00 & $-21.21$ & $-14.39$ & \textbf{$+6.82$ [$+4.36,+9.42$]} \\
    Validity & 53.27 & 53.28 & 52.43 & 52.45 & $-0.01$ & $-0.03$ & $-0.02$ [$-0.31,+0.30$] \\
    \bottomrule
  \end{tabular}
\end{table*}

The frozen 8M test supports Stage A chiefly as an initialization that helps
the adapter reject instruction-incompatible routes. All four N@$k$ DiD
intervals favor Stage A, while preferred-route and validity intervals cross
zero (Table~\ref{tab:stagea-initialization-frozen-test-did}). The evidence
does not establish a promotion gain. Intervals are per metric, without
multiplicity correction or training-seed uncertainty; the 65M evidence
remains validation-only.

\subsection{Stage-B loss ablation}
\label{app:stageb-loss-ablation}

We evaluate the five Stage-B objective terms in a single-seed ablation for
\modelid{USPTO}{Guide(RDKit5)}{65M}{Top30}.  A fresh complete
$P/R/T/G/K$ arm and five leave-one-out arms share the same base checkpoint,
the selected Stage-A A0 projector, zero-output adapter initialization, data
order, optimizer, 14-epoch / 57,260-update budget, and fixed
$\alpha_{\mathrm{train}}=1$.  The complete objective is
\begin{equation}
  \mathcal L_{B,\mathrm{metric}}
  = 0.10\mathcal L_{\mathrm{pos}}
  + 0.05\mathcal L_{\mathrm{rank}}
  + 0.05\mathcal L_{\mathrm{teach}}
  + 0.05\mathcal L_{\mathrm{guard}}
  + 1.00\mathcal L_{\mathrm{roll}}.
  \label{eq:app-stageb-loss-objective}
\end{equation}
The arm labels $P/R/T/G/K$ abbreviate positive DFM, route ranking,
metric-teacher KL, full-reference KL guard, and rollout-reference KL,
respectively; they are not new loss symbols.  Each
leave-one-out arm changes exactly one coefficient to zero.

\begin{figure}[!htbp]
  \centering
  \includegraphics[width=\textwidth]{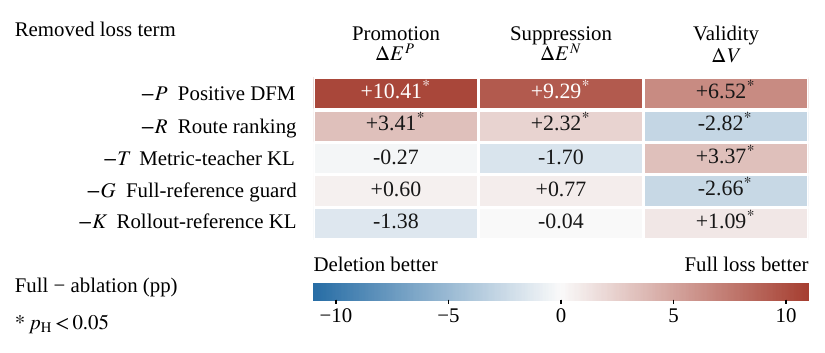}
  \caption{\textbf{Stage-B loss ablation.} Full-minus-ablation effects (pp)
  for \modelid{USPTO}{Guide(RDKit5)}{65M}{Top30} at epoch~14 and
  $\alpha_{\mathrm{eval}}=.50$. Red favors the full loss; blue favors deletion.
  Asterisks mark $p_{\mathrm H}<.05$, Holm-adjusted over the five deletions
  separately for each endpoint. Complete 95\% paired-bootstrap intervals
  and adjusted $p$-values appear in the supplementary statistical table.
  This is a single-seed, validation-only comparison.}
  \label{fig:stageb-loss-ablation-heatmap}
\end{figure}

We evaluate only on a frozen validation cohort of 1,162 case clusters and
2,560 signed groups (256 per axis/polarity cell), at epoch~14 and
$\alpha_{\mathrm{eval}}=.50$. Each Matched or Shuffled condition contains
100 trajectories per group; we reuse the protocol-audited Reference surface.
For ablation variant $v$, we define
$E^P_v=(\mathrm{P}@1)_{v,\mathrm{Matched}}-(\mathrm{P}@1)_{v,\mathrm{Shuffled}}$ and
$E^N_v=(\mathrm{N}@1)_{v,\mathrm{Shuffled}}-(\mathrm{N}@1)_{v,\mathrm{Matched}}$, so larger values
mean stronger instruction-specific promotion or suppression, respectively.
Figure~\ref{fig:stageb-loss-ablation-heatmap} reports
$\Delta E^P=E^P_{\mathrm{full}}-E^P_v$,
$\Delta E^N=E^N_{\mathrm{full}}-E^N_v$, and
$\Delta V=V_{\mathrm{full,Matched}}-V_{v,\mathrm{Matched}}$ in percentage
points.  Positive $\Delta V$ therefore means that deleting the term reduces
validity.  The full arm has $E^P=14.89$, $E^N=14.40$, and Matched validity
of $68.97\%$ (Reference: $70.22\%$), corresponding to a $1.25$-point
Reference--Matched validity loss. The reported $p_{\mathrm H}$ adjusts the
five deletion comparisons by Holm's method separately for each endpoint.
The supplementary document reports the full confidence intervals and adjusted
$p$-values under ``Stage-B loss ablation: full statistical results.''

Positive DFM and route ranking provide the clearest control signal: removing
either weakens promotion and suppression, with paired intervals excluding zero.
Positive DFM also supports validity, whereas ranking trades validity for
control. Metric-teacher and rollout-reference KL support validity without a
resolved control gain. Deleting the guard improves validity with unresolved
control changes, so this single-seed validation ablation does not establish
its intended protective role.


\section{Proposal and selection counterfactuals}
\label{app:proposal-selection-counterfactuals}

\subsection{Budgeted proposal--selection counterfactual}
\label{app:budgeted-proposal-selection}

The USPTO Top30 RDKit5 diagnostic uses the 8M backbone on 1,182 test cases
and 2,560 signed groups. Reference generates without instruction guidance;
Matched uses the correct instruction at Stage-B epoch~12 and
$\alpha=.50$. Each arm contributes
the literal first $B\in\{1,5,10,25,50,100\}$ trajectories without invalid or
duplicate backfilling. The following selectors reorder the valid distinct
candidates in each realized pool, leaving Hit@$B$ and Unique valid unchanged.

\paragraph{Selectors.}
\textbf{Frequency} ranks distinct candidates by descending occurrence count
across the $B$ raw draws.

\noindent\textbf{Scalar head (s42)} applies a width-256 MLP (GELU,
dropout 0.1) to normalized instruction and route embeddings and their
elementwise product, absolute difference, and cosine. The score also includes
log frequency with a learned nonnegative weight. Here s42 denotes training
seed~42.

\noindent\textbf{Exact Stage-A cosine} ranks candidates by the frozen Stage-A
cosine similarity between projected instruction and route embeddings, without
additional learned calibration or a frequency term.

\noindent\textbf{RDKit oracle} directly orders candidates by the instructed
RDKit descriptor: larger values for ``more'' and smaller values for ``fewer.''
It attains the best descriptor utility in the pool and hence zero RDKit regret;
this need not maximize recovery of candidates in the frozen preferred set.

\paragraph{Readout fitting.}
The scalar head learns the requested RDKit preferences from
Reference-generated training pools, using frozen instruction and route
embeddings. We select it by P@1 on a separate validation set, then apply
the same head across proposal pools. Table~\ref{tab:scalar-head-seed-sensitivity}
reports sensitivity to the training seed.
Scalar-head and oracle readouts are evaluation diagnostics; primary inference
uses frequency ordering.

\paragraph{Selector metrics.}
Here Hit@$B$ means preferred-set Hit$_+@B$. P@1/Hit is
$100\,\mathrm{P}@1/\mathrm{Hit}_+@B$, the percentage of groups selecting a
preferred candidate first among those whose pool contains one.
For signed raw RDKit utility $u_g(r)=s_g m_g(r)$, with polarity
$s_g\in\{-1,+1\}$ and descriptor count $m_g$, regret is
$\max_{r\in\mathcal U_g(B)}u_g(r)-u_g(r_{g,1})$:
the gap between the best utility in the realized valid pool $\mathcal U_g(B)$
and the first selected candidate. Lower is better. We average regret over
groups with nonempty pools and metric values for every candidate, using raw
descriptor units rather than the standardized training utility.

\begin{table*}[!htbp]
  \caption{Budgeted Top30 proposal and selector results.  Panel (a) reports
  proposal support at representative budgets.  Panel (b) reports all selectors
  used for the primary Reference and Matched pools at $B=100$.
  Hit, P@1, N@1, P@1/Hit, and duplicate rate are percentages;
  Unique valid is a mean count, and RDKit regret uses raw descriptor units.}
  \label{tab:budgeted-proposal-selection-full}
  \centering
  \begin{tabular}{lrrrr}
    \toprule
    \multicolumn{5}{l}{\textbf{(a) Literal-prefix proposal support}} \\
    Proposal & $B$ & Hit@$B$ & Unique valid & Duplicate rate \\
    \midrule
    Reference  &   5 & 24.38 &  3.40 &  1.55 \\
    Reference  &  25 & 50.98 & 15.68 &  6.08 \\
    Reference  & 100 & 64.80 & 53.53 & 15.23 \\
    Matched   &   5 & 18.44 &  2.64 &  0.62 \\
    Matched   &  25 & 45.35 & 12.59 &  2.42 \\
    Matched   & 100 & 63.87 & 46.95 &  6.10 \\
    \bottomrule
  \end{tabular}

  \vspace{3pt}
  \begin{tabular}{llrrrrrr}
    \toprule
    \multicolumn{8}{l}{\textbf{(b) $B=100$ proposal--selector combinations}} \\
    Proposal & Selector & Hit@100 & \shortstack{Unique\\valid} & P@1 & N@1
      & P@1/Hit & \shortstack{RDKit\\regret} \\
    \midrule
    Reference  & Frequency           & 64.80 & 53.53 & 24.57 & 25.39 & 37.91 & 1.332 \\
    Reference  & Scalar head (s42)   & 64.80 & 53.53 & 22.23 &  3.32 & 34.30 & 0.510 \\
    Reference  & Exact Stage-A cosine& 64.80 & 53.53 &  5.35 & 12.89 &  8.26 & 1.627 \\
    Reference  & RDKit oracle        & 64.80 & 53.53 & 26.99 &  0.35 & 41.65 & 0.000 \\
    Matched & Frequency           & 63.87 & 46.95 & 32.73 &  8.75 & 51.25 & 1.342 \\
    Matched & Scalar head (s42)   & 63.87 & 46.95 & 19.06 &  0.62 & 29.85 & 0.588 \\
    Matched & Exact Stage-A cosine& 63.87 & 46.95 &  7.70 &  6.48 & 12.05 & 1.693 \\
    Matched & RDKit oracle        & 63.87 & 46.95 & 18.48 &  0.12 & 28.93 & 0.000 \\
    \bottomrule
  \end{tabular}
\end{table*}

Guidance improves which proposal reaches the front without a resolved increase
in preferred-set availability. At $B=100$, paired 95\% intervals resolve higher
P@1, lower N@1, and fewer unique-valid candidates under frequency ordering,
while the Hit@100 interval includes zero. This pattern supports prioritization
of preferred routes within the sampled pool.

Terminal selection exposes a different trade-off. All three scalar heads
reduce N@1 and metric regret on both pools while lowering high-budget P@1
(Table~\ref{tab:scalar-head-seed-sensitivity}); exact Stage-A cosine is weaker
at selecting known positives. Even the RDKit oracle can miss those positives
because descriptor optimality and frozen-set membership are different
objectives. The trade-off's magnitude varies across head seeds.

\begin{table*}[!htbp]
  \caption{Scalar-head seed sensitivity at $B=100$ on the frozen test pools.
  Seed 42 is the registered primary head; seeds 43 and 44 are sensitivity runs.
  We select checkpoints independently on Reference validation and reuse them
  without adaptation across proposal sources. P@1 and N@1 are percentages.}
  \label{tab:scalar-head-seed-sensitivity}
  \centering
  \begin{tabular}{llrrrr}
    \toprule
    Proposal & Head seed & Selected epoch & P@1 & N@1 & RDKit regret \\
    \midrule
    Reference & 42 (primary) & 3 & 22.23 & 3.32 & 0.510 \\
    Reference & 43 & 1 & 23.05 & 9.22 & 0.895 \\
    Reference & 44 & 2 & 22.38 & 5.94 & 0.677 \\
    Matched (e12) & 42 (primary) & 3 & 19.06 & 0.62 & 0.588 \\
    Matched (e12) & 43 & 1 & 21.37 & 1.13 & 0.814 \\
    Matched (e12) & 44 & 2 & 19.88 & 0.86 & 0.689 \\
    \bottomrule
  \end{tabular}
\end{table*}


\section{Language and chemistry evidence}
\label{app:language-chemistry-evidence}

\subsection{Free-form structural instructions and the frozen language representation}
\label{app:structure-embedding-ablation}

Structure supplies case-specific retain/avoid instructions rather than a fixed
descriptor registry or scalar metric teacher.  Offline annotation combines
product-local RDKit evidence with language-model judgments
(Appendix~\ref{app:structure-annotation-recipe}).  The frozen test split contains
2,229 cases, 6,752 semantic groups, and 4,606 distinct instruction strings.
Qwen3-Embedding-8B \citep{zhang2025qwen3embedding} represents these strings
as frozen token features; it does not generate their annotations
(Appendix~\ref{app:stage-a-inputs}).
Sampling and aggregation follow Appendix~\ref{app:test-protocol}.

We selected the following verbatim test instructions without consulting
Matched--Shuffled outcomes. They span different constraint types:
\begin{itemize}
  \item \textbf{Core retention and stereochemistry:} ``Prefer precursors that
  retain the intact pyrrolidine core with the correct stereochemistry.''
  \item \textbf{Bond connectivity:} ``Prefer precursors that already contain
  the biaryl bond between the phenanthrene core and the pendant phenyl ring.''
  \item \textbf{Substitution relation:} ``Prefer precursors with a single
  halogen substituent ortho or para to the amino group.''
  \item \textbf{Functional form and route preference:} ``Prefer precursors
  that incorporate the tert-butyl carbamate protected amine directly rather
  than requiring azide or isocyanate intermediates.''
  \item \textbf{Ring-system exclusion:} ``Avoid precursors that introduce
  additional fused or spiro ring systems beyond the target cyclopropane.''
  \item \textbf{Fragment/source integration:} ``Avoid precursors that
  introduce the bromine atom as a salt or separate fragment.''
\end{itemize}

Each instruction combines a case-specific molecular entity with a relation,
polarity, or route-form preference.  The learned projector transfers its
frozen-language features to route scoring and node-to-text memory
(Eqs.~\ref{eq:grounding-score} and \ref{eq:method-token-xattn}).
Matched and Shuffled share the checkpoint and sampling schedule, changing only
which instruction enters this pathway.

\begin{table}[!htbp]
  \caption{Native-cohort Structure absolute-arm results (\%) on 2,229 cases and 6,752 groups. Values average groups; paired case-balanced effects appear in Table~\ref{tab:structure-embedding-summary}.}
  \label{tab:structure-native-absolute}
  \centering
  \begin{minipage}[t]{0.50\linewidth}
    \centering
    \textbf{(a) Preferred-route recovery}\par\smallskip
  \begin{tabular}{@{}l@{\hspace{2.5pt}}r@{\hspace{2.5pt}}r@{\hspace{2.5pt}}r@{\hspace{2.5pt}}r@{}}
      \toprule
      Arm & P@1 & P@3 & P@5 & P@10 \\
      \midrule
      \multicolumn{5}{@{}l}{\shortstack[l]{\textbf{\modelid{USPTO}{Guide(Struct)}{8M}{Top30}}\\e22, $\alpha=.50$}} \\
      Reference & 30.554 & 49.674 & 57.079 & 64.114 \\
      Matched   & 35.382 & 53.525 & 58.960 & 63.181 \\
      Shuffled  & 32.065 & 49.348 & 54.902 & 59.775 \\
      \midrule
      \multicolumn{5}{@{}l}{\shortstack[l]{\textbf{\modelid{USPTO}{Guide(Struct)}{65M}{Top30}}\\e26, $\alpha=1$}} \\
      Reference & 31.117 & 51.466 & 59.064 & 66.558 \\
      Matched   & 45.053 & 62.115 & 66.780 & 70.172 \\
      Shuffled  & 37.456 & 55.213 & 60.930 & 65.744 \\
      \bottomrule
      \end{tabular}
  \end{minipage}\hfill
  \begin{minipage}[t]{0.49\linewidth}
    \centering
    \textbf{(b) Disfavored-route recovery and validity}\par\smallskip
  \begin{tabular}{@{}l@{\hspace{2.5pt}}r@{\hspace{2.5pt}}r@{\hspace{2.5pt}}r@{\hspace{2.5pt}}r@{\hspace{2.5pt}}r@{}}
      \toprule
      Arm & N@1 & N@3 & N@5 & N@10 & Validity \\
      \midrule
      \multicolumn{6}{@{}l}{\shortstack[l]{\textbf{\modelid{USPTO}{Guide(Struct)}{8M}{Top30}}\\e22, $\alpha=.50$}} \\
      Reference & 19.535 & 36.893 & 44.283 & 51.555 & 69.982 \\
      Matched   & 12.678 & 26.052 & 33.797 & 41.410 & 60.567 \\
      Shuffled  & 16.321 & 31.398 & 38.315 & 44.787 & 60.500 \\
      \midrule
      \multicolumn{6}{@{}l}{\shortstack[l]{\textbf{\modelid{USPTO}{Guide(Struct)}{65M}{Top30}}\\e26, $\alpha=1$}} \\
      Reference & 24.097 & 42.210 & 49.985 & 58.175 & 71.231 \\
      Matched   &  6.680 & 14.662 & 18.824 & 23.978 & 64.555 \\
      Shuffled  & 13.907 & 27.636 & 34.864 & 41.484 & 64.706 \\
      \bottomrule
      \end{tabular}
  \end{minipage}
\end{table}

\begin{table}[!htbp]
  \caption{Case-specific structural control: case-balanced Matched--Shuffled
  effects in percentage points with 95\% case-cluster intervals.
  Positive $\Delta$P and $-\Delta$N indicate promotion and suppression,
  respectively.}
  \label{tab:structure-embedding-summary}
  \centering
  \begin{tabular}{lcc}
    \toprule
    Metric & 8M, e22, $\alpha=.50$ & 65M, e26, $\alpha=1$ \\
    \midrule
    $\Delta$P@1 & +3.554 [2.491, 4.598] & +8.089 [6.910, 9.243] \\
    $\Delta$P@10 & +3.769 [2.939, 4.573] & +5.003 [4.155, 5.853] \\
    $-\Delta$N@1 & +3.908 [3.082, 4.755] & +7.799 [6.878, 8.753] \\
    $-\Delta$N@10 & +3.434 [2.525, 4.341] & +18.444 [17.245, 19.651] \\
    $\Delta$Validity & +0.051 [-0.086, +0.190] & -0.035 [-0.278, +0.212] \\
    \bottomrule
  \end{tabular}
\end{table}

Both capacities promote preferred routes and suppress avoided routes at
Top-1 and Top-10, with intervals excluding zero in the requested direction.
These effects support case-specific structural control beyond the fixed
metric registry.
Using the same case-balanced aggregation, the Matched--Reference P@1 gains
and N@10 reductions reported in Table~\ref{tab:main-dataset-guidance} are
5.54 and 10.57 percentage points at 8M, and 14.74 and 35.79 at 65M,
respectively. These point estimates average within-case contrasts;
subtracting the group-averaged absolute values in
Table~\ref{tab:structure-native-absolute} gives a different weighting.

\subsection{A chemistry-audited paired case: fewer precursor halogens}
\label{app:qualitative-halogen-case}

\begin{figure*}[!htbp]
  \centering
  \includegraphics[width=\textwidth]{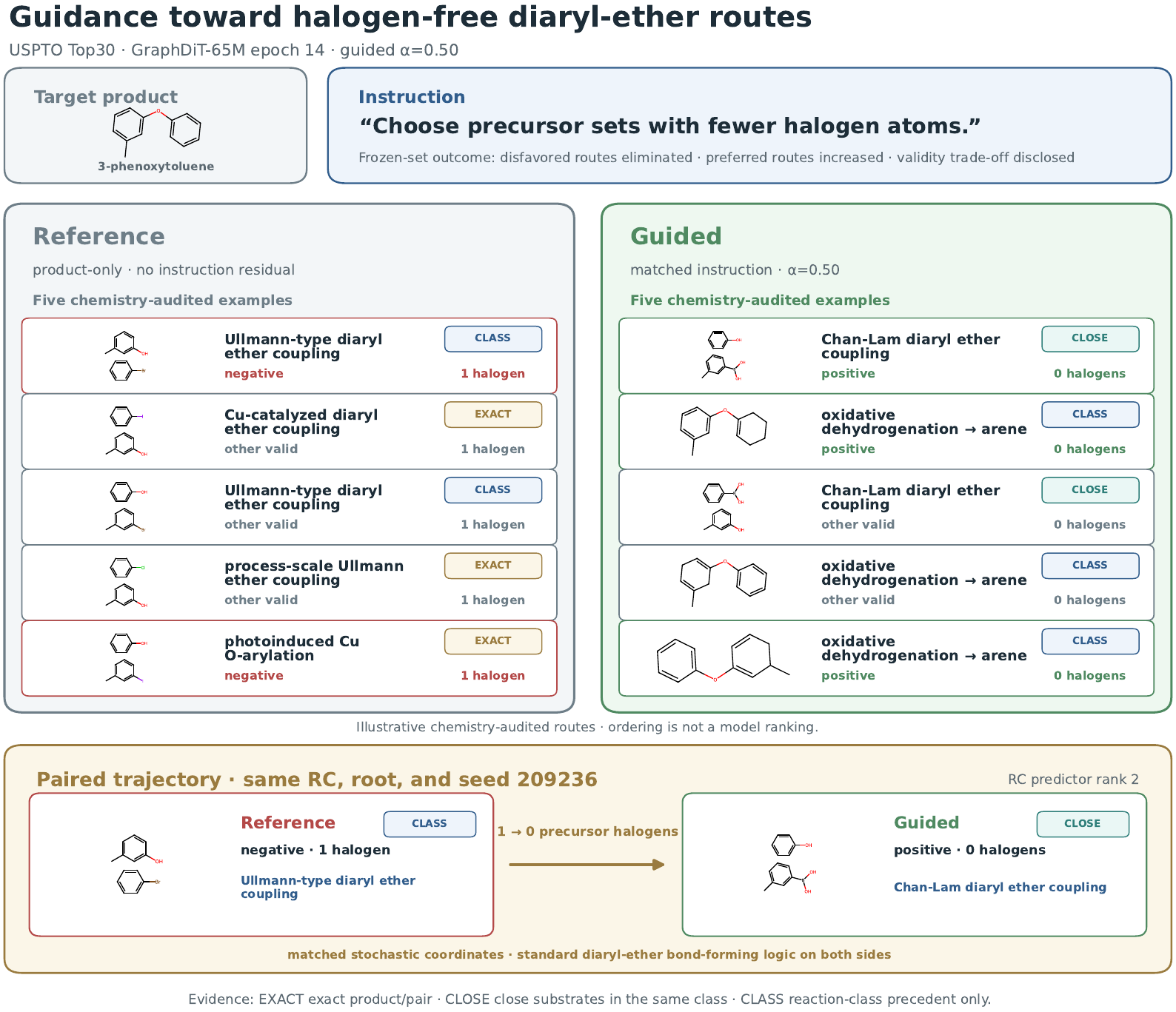}
  \caption{Chemistry-audited qualitative witness for ``Choose precursor sets
    with fewer halogen atoms.''  Each column shows five manually reviewed
    examples selected from frequency-ranked valid unique outputs, not the
    unfiltered model Top-5.  Evidence badges denote exact-product experiments
    (Exact), close-substrate experiments (Close), and reaction-class-only
    precedent (Class). The bottom row pairs outputs at fixed RC schedule and
    seed. Positive/negative labels denote frozen-set membership,
    independently of the evidence badges.  Across all 100 draws, Reference has
    21 negative and 19 positive hits at 91\% validity; Guided has 0 and 24 at
    85\% validity.}
  \label{fig:qualitative-halogen-case}
\end{figure*}

Figure~\ref{fig:qualitative-halogen-case} illustrates how a compositional
instruction can change the chosen disconnection for a fixed product. The
manually audited USPTO Top30 case uses GraphDiT-65M at epoch~14 and the
instruction ``Choose precursor sets with fewer halogen atoms'' for
3-phenoxytoluene, \path{Cc1cccc(Oc2ccccc2)c1}. Guidance shifts draws toward
the preferred set; the caption reports the validity cost.

We rank non-invalid unique outputs by empirical frequency (ties use first
trajectory index) and show five examples per arm that pass manual checks for
atom-source consistency, meta regiochemistry, an identifiable reaction class,
and primary-literature precedent.  These are curated chemistry witnesses, not
the unfiltered model Top-5; curation does not alter any quantitative count.
The two Reference bromoarene routes have reaction-class precedent for
ultrasound-assisted Ullmann coupling \citep{smith1992ultrasound}.
The remaining three Reference routes have exact-product precedent
\citep{ling2012supported_cuo,chandnani1999process,tan2014photoarylation}.
The Guided examples comprise two close-substrate Chan--Lam precedents and three
reaction-class-only oxidative-dehydrogenation precedents
\citep{chan1998arylations,evans1998diaryl,iosub2015dehydrogenation}.

We fix RC-predictor rank 2, the root choice, and trajectory seed 209236. The
bottom row then changes from a one-halogen Ullmann disconnection with reaction-class precedent to a
zero-halogen, close-substrate Chan--Lam disconnection.  It is a qualitative
witness of the requested compositional shift; we evaluate aggregate behavior
separately.

\subsection{Downstream multistep planning witnesses}
\label{app:multistep-planning-witnesses}

\begin{figure}[!htbp]
  \centering
  \includegraphics[width=\linewidth]{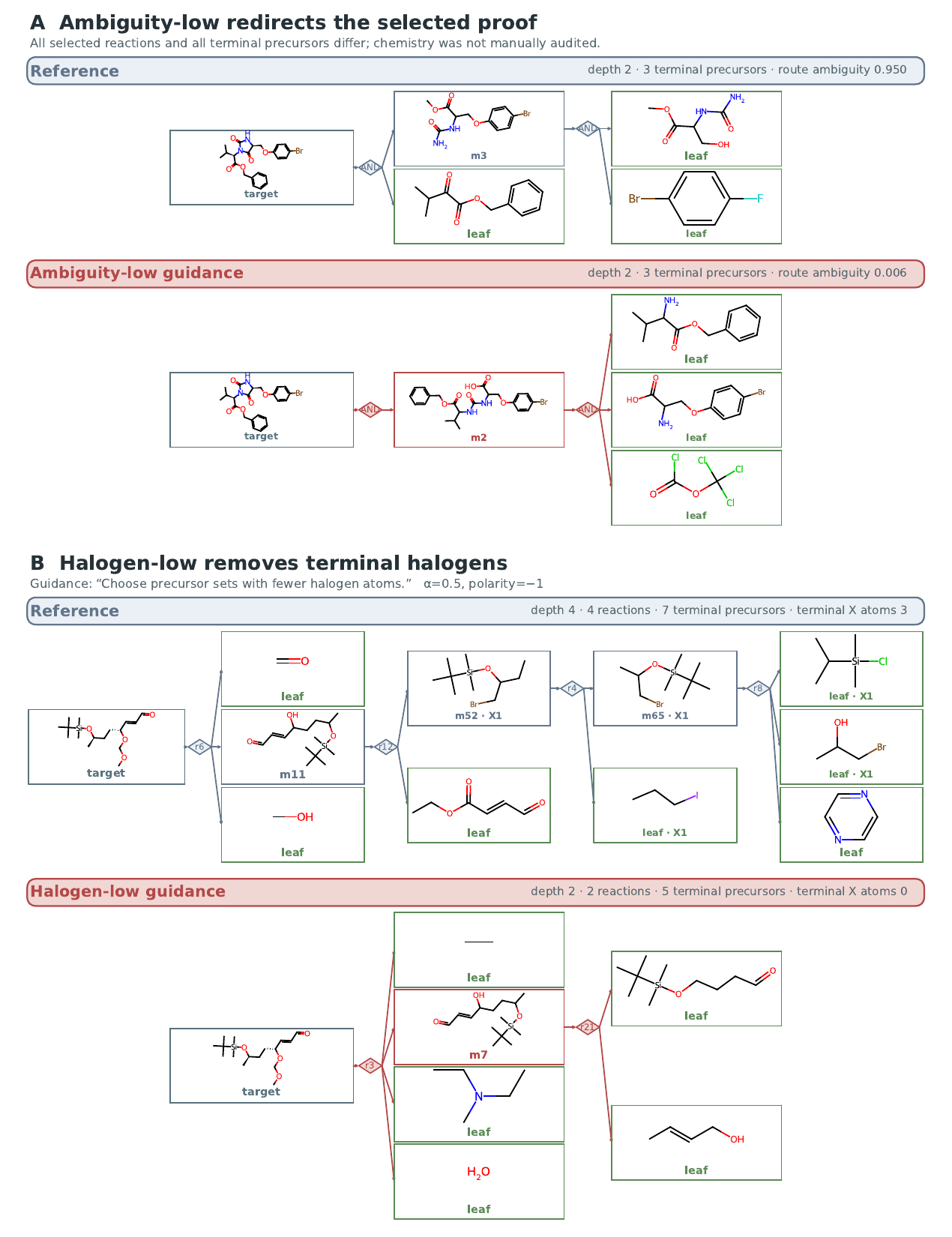}
  \caption{Selected proofs for two qualitative planning witnesses.  (A)
  Ambiguity-low redirects the route at unchanged depth; chemical
  feasibility remains unverified. (B) Halogen-low redirects the selected plan;
  both searches find a complete plan, and both selected plans pass the
  structural audit. Cards are molecule OR nodes, diamonds are reaction
  AND nodes, and $Xn$ marks $n$ halogen atoms.}
  \label{fig:multistep-selected-proofs}
\end{figure}

Figure~\ref{fig:multistep-selected-proofs} illustrates how local instruction
following can propagate through a multistep search: changing precursor
choices redirects later expansions and the resulting terminal precursors.
The two paired witnesses establish this possibility; they do not estimate
average planning gains, cross-seed stability, or experimental feasibility.

\paragraph{Search method.}
We use a custom value-guided AND--OR tree-search planner inspired by
Retro*~\citep{chen2020retrostar}. Molecule nodes represent alternative
reactions (OR), and reaction nodes require all their precursors (AND).
A frozen value model estimates the remaining search effort for unexpanded
molecules. Reaction costs sum precursor costs, and molecule costs take the
minimum over candidate reactions. The planner selects an unresolved molecule
in the current lowest-cost partial plan and expands it with RIGS, then
updates these costs back toward the target. Guided searches apply the same
instruction at every expansion. Search ends when all leaves of a complete
route belong to the building-block inventory, no expandable branch remains,
or the search reaches 500 expansions per target.

\paragraph{Route-level measures.}
Route-level reaction-class ambiguity is $\max_j(1-p_j)$, where $p_j$ is
the ASKCOS classifier's top-class confidence for reaction $j$
\citep{tu2025askcos}; lower values indicate more confident classification.
Proof depth is the longest chain of
reaction steps. Selected reactions and terminal precursors count reaction
nodes and leaf precursors, respectively; terminal halogen count sums the
halogen atoms over those leaves. A solved proof reaches the planner's
terminal-precursor criterion at every leaf.

\paragraph{Ambiguity-low selected proof.}
For target \texttt{pistachio100:038},
\path{CC(C)[C@H](C(=O)OCc1ccccc1)N1C(=O)N[C@@H](COc2ccc(Br)cc2)C1=O},
guidance reduces route-level reaction-class ambiguity while changing every
selected reaction and terminal precursor, despite equal proof depth. This is
a screen-selected computational witness: automatic
RDKit parsing and route-continuity checks pass, but we have not manually
audited its chemistry.

\paragraph{Halogen-low planning case.}
For target index~20,
\path{COCO[C@@H](/C=C/C=O)CC[C@@H](C)O[Si](C)(C)C(C)(C)C},
we compare Reference and guided searches using RIGS-GraphDiT-65M.
The guided arm uses ``Choose precursor sets with fewer halogen atoms''
with $\alpha=0.5$ and polarity $-1$.
Guidance changes the root and every later selected reaction,
yielding halogen-free terminal precursors
(Figure~\ref{fig:multistep-selected-proofs}B).
The local preference thus affects the whole selected plan, including its depth.


\subsection{Direct empirical sampled-mass reallocation}
\label{app:empirical-mass-reallocation}

Top-$k$ recovery measures whether a route set appears in the returned list;
empirical mass measures how often the sampler visits it. We reclassify the
frozen Common Top100 trajectories into five mutually exclusive outcomes: preferred,
disfavored, neutral Top100, valid outside Top100, or invalid.  For semantic
group $g$, let $\widehat r_g^{(b)}$ be the canonical precursor set decoded
from draw $b$ (or an invalid sentinel), and let $\mathcal R_g^+$ and
$\mathcal R_g^-$ be the preferred and avoided precursor sets in its frozen
evaluation inventory.  The preferred empirical mass in the first $B$ draws is
\begin{equation}
  M_+(g;B)
  = \frac{1}{B}\sum_{b=1}^{B}
    \mathbf{1}\!\left[\widehat r_g^{(b)}\in\mathcal R_g^+\right],
  \label{eq:empirical-preferred-mass}
\end{equation}
with $M_-(g;B)$ defined using $\mathcal R_g^-$.  These sets contain the
canonical routes indexed by $\mathcal P_c$ and $\mathcal N_c$, respectively,
in the evaluation group.  Repeated occurrences count repeatedly;
we neither deduplicate outputs nor backfill invalid draws.  These quantities
are literal frequencies under the frozen 100-draw protocol, not exact model
probabilities. We report results at $B=100$, using the equal-axis,
case-balanced aggregation and paired bootstrap in Appendix~\ref{app:test-protocol}.
The signed mass gap is $M_+-M_-$.

\begin{table}[!htbp]
  \caption{Direct empirical sampled-mass reallocation at $B=100$.  Each cell is
  the equal-axis, case-balanced Matched--Shuffled effect in percentage points,
  followed by its 95\% paired-bootstrap interval. We reverse the sign of
  disfavored mass, so positive $-\Delta M_-$ is favorable. The ten rows are the
  selected Common Top100 checkpoints registered in
  Table~\ref{tab:common-top100-coordinates}.}
  \label{tab:empirical-mass-reallocation}
  \centering
  \begin{tabular}{llrrrr}
    \toprule
    Capacity & Support & $\Delta M_+$ & $-\Delta M_-$
      & $\Delta(M_+-M_-)$ & $\Delta$Validity \\
    \midrule
    8M  & USPTO-MR & \effectci{+0.30}{+0.21,+0.39} & \effectci{+0.38}{+0.30,+0.46} & \effectci{+0.67}{+0.53,+0.82} & \effectci{-0.10}{-0.33,+0.13} \\
    8M  & Top15    & \effectci{+0.94}{+0.82,+1.05} & \effectci{+1.04}{+0.91,+1.17} & \effectci{+1.98}{+1.78,+2.17} & \effectci{-0.06}{-0.21,+0.09} \\
    8M  & Top30    & \effectci{+2.27}{+2.10,+2.44} & \effectci{+2.26}{+2.09,+2.43} & \effectci{+4.53}{+4.24,+4.82} & \effectci{-0.11}{-0.43,+0.22} \\
    8M  & Top45    & \effectci{+2.67}{+2.48,+2.85} & \effectci{+2.69}{+2.50,+2.89} & \effectci{+5.36}{+5.04,+5.68} & \effectci{+0.15}{-0.18,+0.48} \\
    8M  & Top60    & \effectci{+1.26}{+1.15,+1.37} & \effectci{+1.20}{+1.09,+1.32} & \effectci{+2.46}{+2.28,+2.64} &  \effectci{0.00}{-0.20,+0.20} \\
    \midrule
    65M & USPTO-MR & \effectci{+0.12}{+0.04,+0.19} & \effectci{+0.16}{+0.10,+0.22} & \effectci{+0.27}{+0.17,+0.38} & \effectci{+0.06}{-0.13,+0.24} \\
    65M & Top15    & \effectci{+3.28}{+3.01,+3.55} & \effectci{+3.26}{+3.00,+3.53} & \effectci{+6.54}{+6.11,+6.98} & \effectci{+0.14}{-0.25,+0.53} \\
    65M & Top30    & \effectci{+5.18}{+4.88,+5.49} & \effectci{+5.06}{+4.74,+5.38} & \effectci{+10.24}{+9.71,+10.77} & \effectci{+0.11}{-0.20,+0.44} \\
    65M & Top45    & \effectci{+4.07}{+3.82,+4.32} & \effectci{+4.02}{+3.77,+4.27} & \effectci{+8.09}{+7.67,+8.53} & \effectci{+0.09}{-0.30,+0.48} \\
    65M & Top60    & \effectci{+0.26}{+0.18,+0.35} & \effectci{+0.21}{+0.13,+0.28} & \effectci{+0.47}{+0.35,+0.58} & \effectci{+0.10}{-0.06,+0.27} \\
    \bottomrule
  \end{tabular}
\end{table}

Matched text reallocates repeated draws toward preferred routes and away from
avoided routes at all ten coordinates, with paired intervals excluding zero
in both directions (Table~\ref{tab:empirical-mass-reallocation}). Validity
contrasts remain unresolved, and invalid draws stay in the denominator.
The effect therefore extends beyond changing which route crosses a rank cutoff.

%

\section{External and heterogeneous robustness}
\label{app:external-robustness}
\label{app:complete-results}


\subsection{Pistachio Top30: control on recorded reaction alternatives}
\label{app:pistachio-replication}

We report a 65M Guide(RDKit5) model trained on Pistachio Top30.
We select the evaluation cohort from the native Top30 RDKit5 test groups
and retain their preference labels. It contains 1,035 products and 2,560
signed groups balanced across the five RDKit axes and both polarities.
All three arms use the same fixed cohort and labels.
We check the cohort's products and eligible precursor sets against the
Pistachio Top100 test inventory of 3,516 products and 17,556 recorded
candidate routes. In this data construction, we add template-generated
candidates only to training and validation. The evaluation-inventory count differs from the
17,549 accepted graph routes in Table~\ref{tab:base-data-matrix}, which reports
the model-bound graph view.

Reference, Matched, and different-axis Shuffled share the reaction-center
top-5 proposal schedule and sampling seeds, using 100 raw draws per group
and 50 steps.  Root and trajectory seeds are 20260721 and 42000.
Intervals use 10,000 paired case-cluster bootstrap replicates with equal
axis weights (seed 1042).  Table~\ref{tab:pistachio-full-arms} gives the
recorded checkpoint epoch and residual scale alongside all three arms.

\begin{table}[!htbp]
  \caption{\textbf{Pistachio 65M Top30 five-axis results on recorded reaction alternatives.}
  Stage-B checkpoint epoch $e=10$; residual scale $\alpha=0.50$.
  All entries are percentages. Hit$_+@100$ measures preferred-set availability;
  we compute validity over raw draws. P@$k$/N@$k$ use frequency-ranked
  unique valid outputs.}
  \label{tab:pistachio-full-arms}
  \centering
  \begin{tabular*}{\linewidth}{@{\extracolsep{\fill}}lr@{\hspace{12pt}}rrrr@{\hspace{14pt}}rrrr@{\hspace{12pt}}r@{}}
    \toprule
    & & \multicolumn{4}{c}{Preferred recovery $\uparrow$}
      & \multicolumn{4}{c}{Disfavored recovery $\downarrow$} & \\
    \cmidrule(lr){3-6}\cmidrule(lr){7-10}
    Arm & Hit$_+@100$ & P@1 & P@3 & P@5 & P@10
      & N@1 & N@3 & N@5 & N@10 & Validity \\
    \midrule
    Reference & 35.59 & 3.48 & 7.89 & 11.52 & 18.52 & 3.83 & 7.77 & 10.51 & 18.63 & 73.82 \\
    \addlinespace[2pt]
    Matched & 35.63 & 4.45 & 10.47 & 14.80 & 21.29 & 0.66 & 1.99 & 3.20 & 5.04 & 69.39 \\
    \addlinespace[2pt]
    Shuffled & 23.91 & 2.93 & 5.90 & 7.97 & 12.62 & 2.42 & 6.13 & 8.83 & 12.23 & 69.26 \\
    \bottomrule
  \end{tabular*}
\end{table}

Control also appears on recorded reaction alternatives: paired 95\% intervals
resolve preferred-route promotion and avoided-route suppression at Top-10
against Shuffled. Against Reference, P@1 and P@10 improve and N@10 falls,
while the preferred-set Hit@100 interval includes zero. This supports a
change in which available routes the model returns first, with a validity cost
visible in Table~\ref{tab:pistachio-full-arms}.

\subsection{Applicability to another metric suite}
\label{app:metric4-generalization}
\label{app:axis4-top100-protocol}
\label{app:axis4-top100-complete}

Metric4 tests the same two-stage framework with a different supervision
suite: synthetic-complexity delta, procurement burden, halogen count, and
ring count. We train separate models with Metric4 supervision,
so the experiment assesses applicability under metric-specific training.
All ten settings share a strict-Top100 inventory of 982 products and
2,048 signed groups. Each arm uses RC Top-5 proposals, 100 raw draws per
group, and 50 sampling steps; Matched and Shuffled share sampling seeds.
We select checkpoints and guidance scales on validation and keep them fixed for testing.
Estimates average groups within each case and then average cases;
95\% intervals use 10,000 paired case-cluster bootstrap resamples.

Table~\ref{tab:metric4-generalization-summary} retains all five supports at
both capacities. Seven of the eight synthetic-support settings show
preferred Top-1 promotion and avoided Top-10 suppression against Shuffled,
and preferred Top-1 gains over Reference, with intervals excluding zero.
The P@1 instruction effect remains unresolved for USPTO-MR and 65M Top60.
Validity costs relative to Reference remain visible in the last column;
Matched--Shuffled validity intervals include zero at all ten settings.
These results support applicability beyond RDKit5, with gains that depend
on the support and deployment setting.
For the Top30 settings in Table~\ref{tab:main-dataset-guidance}, the
Matched--Reference reduction in N@10 is additionally
$8.04$ percentage points (95\% CI: $[6.52,9.60]$) at 8M and
$27.52$ points ($[25.15,29.91]$) at 65M, using the same case-balanced
aggregation and paired bootstrap.

\begin{table}[!htbp]
  \caption{Instruction control under Metric4 supervision.
  Entries are percentage-point effects with 95\% confidence intervals.
  Matched--Shuffled tests instruction identity; Matched--Reference measures
  gains and validity costs over unguided generation.
  Positive $\Delta$P and $-\Delta$N are favorable; negative
  $\Delta$Validity indicates a loss.
  $(e,\alpha)$ gives the validation-selected Stage-B epoch and guidance scale.}
  \label{tab:metric4-generalization-summary}
  \label{tab:axis4-top100-coordinates}
  \centering
  \appto\papertablestyle{\setlength{\tabcolsep}{3.5pt}}
  \begin{tabular}{@{}llcccc@{}}
    \toprule
    & & \multicolumn{2}{c}{Matched--Shuffled}
      & \multicolumn{2}{c}{Matched--Reference} \\
    \cmidrule(lr){3-4}\cmidrule(lr){5-6}
    Support & $(e,\alpha)$ & $\Delta$P@1 & $-\Delta$N@10
      & $\Delta$P@1 & $\Delta$Validity \\
    \midrule
    \multicolumn{6}{@{}l}{\textbf{GraphDiT-8M}} \\
    USPTO-MR & $(14,.50)$ & \effectci{+0.76}{-0.05,+1.63} & \effectci{+2.62}{+1.48,+3.77}
      & \effectci{+1.35}{+0.56,+2.19} & \effectci{-1.46}{-1.72,-1.20} \\
    Top15 & $(12,.25)$ & \effectci{+3.46}{+2.14,+4.84} & \effectci{+7.03}{+5.50,+8.61}
      & \effectci{+3.06}{+1.55,+4.58} & \effectci{-2.88}{-3.10,-2.65} \\
    Top30 & $(6,.25)$ & \effectci{+5.37}{+3.69,+7.08} & \effectci{+7.41}{+5.91,+8.96}
      & \effectci{+4.07}{+2.34,+5.78} & \effectci{-2.76}{-2.97,-2.55} \\
    Top45 & $(14,.50)$ & \effectci{+12.17}{+10.08,+14.21} & \effectci{+19.96}{+17.64,+22.30}
      & \effectci{+9.04}{+6.98,+11.05} & \effectci{-3.62}{-3.90,-3.34} \\
    Top60 & $(14,.25)$ & \effectci{+6.19}{+4.35,+8.02} & \effectci{+10.77}{+8.94,+12.65}
      & \effectci{+4.86}{+3.06,+6.67} & \effectci{-1.70}{-1.94,-1.47} \\
    \midrule
    \multicolumn{6}{@{}l}{\textbf{GraphDiT-65M}} \\
    USPTO-MR & $(12,.50)$ & \effectci{+0.23}{-0.59,+1.04} & \effectci{+1.81}{+0.74,+2.90}
      & \effectci{+0.82}{+0.08,+1.58} & \effectci{+1.11}{+0.78,+1.46} \\
    Top15 & $(12,.25)$ & \effectci{+7.71}{+5.98,+9.47} & \effectci{+11.35}{+9.60,+13.21}
      & \effectci{+7.10}{+5.50,+8.76} & \effectci{-2.02}{-2.23,-1.80} \\
    Top30 & $(10,.50)$ & \effectci{+15.78}{+13.70,+17.90} & \effectci{+21.92}{+19.50,+24.34}
      & \effectci{+13.44}{+11.35,+15.48} & \effectci{-0.02}{-0.31,+0.28} \\
    Top45 & $(14,.50)$ & \effectci{+15.81}{+13.57,+18.05} & \effectci{+23.85}{+21.16,+26.60}
      & \effectci{+11.25}{+9.06,+13.47} & \effectci{-5.09}{-5.46,-4.70} \\
    Top60 & $(2,.05)$ & \effectci{-1.15}{-2.55,+0.20} & \effectci{+0.76}{-0.36,+1.88}
      & \effectci{-0.84}{-2.27,+0.61} & \effectci{+0.08}{-0.07,+0.23} \\
    \bottomrule
  \end{tabular}
\end{table}

\section{Data, checkpoint, and evaluation provenance}
\label{app:data-provenance}
\label{app:data}

\subsection{Identifiers, data units, and reporting scope}
\label{app:data-identifiers}

Model identifiers follow \textrm{Dataset--Role--Scale--Support}, where
\textrm{Base} denotes a no-instruction generator and \textrm{Guide(Struct)},
\textrm{Guide(Metric4)}, or \textrm{Guide(RDKit5)} identifies the adapter's
supervision suite.  Scale denotes the 8M or 65M backbone and Support its
training candidate view. We report checkpoint, epoch, residual scale, and
sampling budget separately. Reference, Matched, and Shuffled denote
evaluation conditions, while Null reuses Reference at $\alpha=0$.

Tables in this section distinguish the units in
Table~\ref{tab:data-counting-units}.  In compact cells, \textbf{P/R} means
usable products (or cases) / graph-executable routes, and \textbf{G/R} means
signed semantic groups / graph-bound route keys or rows.  The tables report
model-bound supervision and base-generator training views.  Test G/R denotes
the native execution-view size. We describe sampled rollout cohorts,
including the shared Top100 evaluations, separately.

\begin{table}[!htbp]
  \caption{Counting units used in the data tables.}
  \label{tab:data-counting-units}
  \centering
  \begin{tabular}{lp{0.69\columnwidth}}
    \toprule
    Unit & Definition \\
    \midrule
    Product/case
      & One split-local canonical target product and its candidate pool. \\
    \midrule
    Graph route
      & One candidate precursor set that survives canonicalization, mapping,
        graph-capacity, and route-binding checks. \\
    \midrule
    Semantic group
      & One case--instruction--polarity comparison over several routes. \\
    \midrule
    Paired family
      & The two opposite-polarity semantic groups for one case and axis. \\
    \midrule
    Candidate occurrence
      & A route appearing inside a semantic group; not a distinct base route. \\
    \bottomrule
  \end{tabular}
\end{table}

\subsection{Construction of USPTO-MR}
\label{app:uspto-mr-construction}

We construct USPTO-MR from USPTO-Full~\citep{lowe2017uspto} to capture
multiple observed single-step retrosynthetic alternatives for the same
product. We remove atom-mapping numbers for molecular identity comparison,
canonicalize product and precursor SMILES while preserving stereochemistry,
and group reactions by their canonical products. We collapse duplicate
precursor combinations within each product group irrespective of fragment
ordering and retain products with at least two distinct observed alternatives.
We screen reactions for SMILES validity, atom-mapping
consistency, identity transformations, and graph-size constraints. Following
canonical normalization and deduplication, USPTO-MR contains 32,584 unique
products and 77,110 product--precursor pairs. To prevent product overlap
across splits, we assign all reactions associated with a canonical product
to the same split. For products appearing in multiple original
splits, we prioritize test over validation over training, resulting in
19,022 training, 6,098 validation, and 7,464 test products. Each alternative
corresponds to a precursor combination recorded in the source dataset.

\subsection{Base-generator data matrix}
\label{app:base-data-matrix}

All synthetic supports use the same library of 270,794 retrosynthesis
templates extracted from USPTO-Full and released with DESP~\citep{yu2024desp},
shared across disjoint product splits.
Table~\ref{tab:main-data-summary} summarizes USPTO-MR and its synthetic
expansions after preprocessing.
Table~\ref{tab:base-data-matrix} reports the exact model-bound,
graph-executable views rather than upstream candidate rows.  For USPTO,
the 8M and 65M backbones reuse the same rows at each support.  USPTO
Top15/30/45/60 are synthetic-only nested prefix views; USPTO-MR is a separate
natural-data baseline.  Pistachio Top30 expands train and validation,
while its registered test split contains only recorded reactions; the
reported Pistachio model uses the 65M backbone.

\begin{table}[!htbp]
  \caption{\textbf{USPTO-MR and synthetic expansions.} Totals after preprocessing;
  full split counts in Table~\ref{tab:base-data-matrix}.}
  \label{tab:main-data-summary}
  \centering
  \begin{tabular}{lrrr}
    \toprule
    Support & Products & \shortstack{Precursor\\sets}
      & \shortstack{Avg./\\product} \\
    \midrule
    USPTO-MR & 32,584 & 77,110 & 2.37 \\
    Top15 & 31,912 & 470,308 & 14.74 \\
    Top30 & 31,912 & 922,977 & 28.92 \\
    Top45 & 31,912 & 1,360,737 & 42.64 \\
    Top60 & 31,912 & 1,788,168 & 56.03 \\
    \bottomrule
  \end{tabular}
\end{table}

\begin{table*}[!htbp]
  \caption{Base-generator data sizes.  Each cell is usable products /
  accepted graph routes (P/R). \emph{Both} means \textrm{8M} and \textrm{65M}
  use identical data rows; the two backbones still have separate
  checkpoints.  USPTO Top60 reports the frozen model-training view, not the
  larger later annotation inventory. Pistachio's accepted test graph view
  (17,549 routes) is separate from the recorded-candidate evaluation inventory
  (17,556 routes; Appendix~\ref{app:pistachio-replication}).}
  \label{tab:base-data-matrix}
  \centering
  \begin{tabular}{lllrrrr}
    \toprule
    Dataset & Support & Scale(s) & Train P/R & Validation P/R & Test P/R
      & Total P/R \\
    \midrule
    USPTO & USPTO-MR & Both
      & \countpair{19,022}{41,488} & \countpair{6,098}{14,740} & \countpair{7,464}{20,882}
      & \countpair{32,584}{77,110} \\
    USPTO & Top15 & Both
      & \countpair{18,595}{274,018} & \countpair{5,975}{88,142} & \countpair{7,342}{108,148}
      & \countpair{31,912}{470,308} \\
    USPTO & Top30 & Both
      & \countpair{18,595}{537,777} & \countpair{5,975}{172,969} & \countpair{7,342}{212,231}
      & \countpair{31,912}{922,977} \\
    USPTO & Top45 & Both
      & \countpair{18,595}{793,099} & \countpair{5,975}{254,984} & \countpair{7,342}{312,654}
      & \countpair{31,912}{1,360,737} \\
    USPTO & Top60 & Both
      & \countpair{18,595}{1,042,205} & \countpair{5,975}{335,110} & \countpair{7,342}{410,853}
      & \countpair{31,912}{1,788,168} \\
    \midrule
    Pistachio & Top30 & 65M
      & \countpair{36,931}{1,189,822} & \countpair{391}{13,142} & \countpair{3,516}{17,549}
      & \countpair{40,838}{1,220,513} \\
    \bottomrule
  \end{tabular}
\end{table*}

We assign products with evaluation precedence, keeping train, validation,
and test disjoint. The USPTO-MR graph cache removes one
duplicate precursor-set route from the 77,111-row source.  The Top$K$ rows are
therefore not raw template counts: they are the routes the corresponding base
model can actually sample during optimization.

\subsection{Synthetic precursor construction and ranking}
\label{app:candidate-ranking}

Candidate generation builds on the template-based reaction validation in
AOT*~\citep{song2025aotstar}. Our implementation uses product-side fingerprint
screening, exact SMARTS matching, and cached RDChiral template execution.
A fixed heuristic selector balances quality and
precursor diversity; its scores are not calibrated reaction-success
probabilities. The accompanying USPTO package supplies the constructed
supports and their ordering, so reproduction starts from these fixed data.

\paragraph{Instantiating complete precursor sets.}
We apply each screened template to the \emph{complete} product molecule
with RDChiral~\citep{coley2019rdchiral}. Its product-side SMARTS identifies
the local reaction environment; atom mappings anchor the precursor-side
bond edits and template-specified atoms (e.g., leaving groups) to that
match. RDChiral carries over molecular context outside the matched pattern
from the product, restores non-reacted bonds, and handles
stereochemical consistency. Each valid outcome therefore contains complete
precursor molecules, whose disconnected components form one precursor set.
We remove atom-map labels, canonicalize each component as isomeric SMILES,
sort the components while preserving multiplicity, and deduplicate
identical sets across template matches, retaining their template
provenance. We discard invalid outcomes and the unchanged-product singleton.
The quality--diversity selector below then ranks these complete
precursor sets to define the Top$K$ supports.

\paragraph{Quality components.}
For product $x$ and precursor set $r$, the quality score $Q_x(r)$ combines
four heuristic signals: \emph{template support}, favoring candidates
proposed by multiple distinct templates; \emph{template specificity},
favoring more detailed product-side reaction patterns; \emph{precursor
count}, favoring two-component sets; and \emph{largest-fragment reduction},
favoring a smaller largest precursor relative to the product. We take a
weighted average with weights 0.35, 0.30, 0.15, and 0.20, respectively,
renormalized over the available signals.

\paragraph{Diversity and greedy selection.}
For the selected set $S$, diversity is
$D(r\mid S)=1-\max_{u\in S}\operatorname{Tanimoto}(F(r),F(u))$, with
$D(r\mid\varnothing)=1$. $F$ is a 2,048-bit, radius-two Morgan fingerprint
of the full disconnected precursor set.
At each step, the selector maximizes
\begin{equation}
  s_x(r\mid S)=0.60Q_x(r)+0.40D(r\mid S)
  \label{eq:candidate-selection-score}
\end{equation}
over the remaining eligible candidates, then adds the chosen candidate to
$S$ and updates the similarities. The selector breaks ties by higher
$Q_x$, then higher $D$, then lexicographically smaller canonical precursor
SMILES. The first retained generating-template identifier is the candidate's
primary template. A candidate is eligible while fewer than three previously
selected candidates have that primary template; candidates without a primary
identifier remain eligible. If no remaining candidate satisfies this quota,
the selector relaxes it for that step.

The selector stops after $K$ candidates or when no candidates remain. The
selection order defines $\operatorname{rank}_x(r)$ in
Section~\ref{sec:method-support}; each stored \texttt{selection\_score} is
the value of Eq.~\ref{eq:candidate-selection-score} when the selector chose
that candidate. For a fixed input pool and configuration, changing $K$ only
changes the stopping point, yielding nested prefixes. The shared graph
checks and deduplication in $\operatorname{Filter}$ then determine the
usable Base training support, whose size can be below $K$.

\subsection{Guidance suites and construction}
\label{app:guidance-suite-construction}

The three guidance suites are different supervision sources and are not
interchangeable aliases.  Structure guidance is case-specific, whereas Metric4
and RDKit5 use fixed metric registries with opposite signed directions.
Table~\ref{tab:guidance-family-construction} gives the construction method,
software provenance, and two verbatim canonical examples from the
materialized text pools.
The accompanying data schema documents unannotated candidate records
(item C2). The supplementary training example (item C3) illustrates the
normalized Structure view: we retain Positive/Negative candidates, omit
Neutral candidates, and renormalize annotation probabilities over the
retained candidates.
Structure annotation details appear in
Appendix~\ref{app:structure-annotation-recipe}.

\paragraph{Metric values.}
For Metric4, complexity reduction is the target SCScore minus the mean
precursor SCScore \citep{coley2018scscore}. Procurement burden combines
precursor prices with penalties for missing quotes and longer lead times,
using \href{https://github.com/ASKCOS/askcos-data}{ASKCOS buyables data}
\citep{tu2025askcos}.

\paragraph{Metric preferences and target weights.}
Within each product's pool, we normalize RDKit counts using z-scores and
complexity reduction and procurement burden using robust z-scores.
For normalized values $\widetilde d_i$ and instruction direction $s_c$
($+1$ for higher, $-1$ for lower), the upper and lower quartile tails of
$s_c\widetilde d_i$ define preferred candidates $\mathcal P_c$ and avoided
candidates $\mathcal N_c$, including boundary ties. We exclude middle or
missing-value candidates and groups without a valid contrast. The Stage-A
target is a softmax over the retained candidates:
\[
  q_i^c=\frac{\exp(s_c\widetilde d_i)}
  {\sum_{j\in\mathcal P_c\cup\mathcal N_c}\exp(s_c\widetilde d_j)},
  \qquad i\in\mathcal P_c\cup\mathcal N_c.
\]
Appendices~\ref{app:guidance-data-matrix} and~\ref{app:stage-b-loss-details}
describe support filtering and the Stage-B metric teacher, respectively.

\begin{table*}[!htbp]
  \caption{Guidance types, construction methods, and representative canonical
  instructions. We compute metric values before writing instructions and do
  not ask the language model to estimate Metric4 or RDKit5 values.}
  \label{tab:guidance-family-construction}
  \centering
  \begin{tabular}{p{0.10\textwidth}p{0.24\textwidth}p{0.33\textwidth}p{0.23\textwidth}}
    \toprule
    Suite & Signal & Construction and software & Two canonical examples \\
    \midrule
    Structure
      & Product-local structural motifs, connectivity, fragment count, and
        retain/avoid criteria.
      & RDKit extracts candidate evidence; a frozen LLM listwise annotator
        proposes at most five relative criteria and route preferences.
        Deterministic filtering forms prefer/avoid views; RXNMapper
        \citep{schwaller2021rxnmapper} and RDKit bind surviving routes to
        executable graphs.
      & \emph{Prefer precursors that already contain the imidazole ring
        directly attached to the propyl chain.}\newline
        \emph{Avoid precursors containing heteroatoms other than bromine and
        oxygen.} \\
    \midrule
    Metric4
      & \path{mosyn.synthetic_complexity_delta},
        \path{mosyn.procurement_burden},
        \path{rdkit.halogen_atom_count}, and
        \path{rdkit.ring_count}.
      & MOSYN joins deterministic route metrics; SCScore supplies molecular
        complexity and ASKCOS buyables data supplies procurement inputs.
        RDKit computes the two structural counts.  Pool-normalized signed
        tails create the two polarities.
      & \emph{Choose disconnections that deliver a larger reduction in
        synthetic complexity.}\newline
        \emph{Choose precursor sets with a lighter procurement burden.} \\
    \midrule
    RDKit5
      & RDKit halogen-atom, carbonyl-group, total-ring, aromatic-ring, and
        heterocycle counts.
      & RDKit atom/ring APIs and fixed SMARTS compute each descriptor over the
        complete precursor set.  Pool-normalized, tie-safe signed tails
        produce labels; deterministic rule-authored templates produce wording.
      & \emph{Choose precursor sets with fewer halogen atoms.}\newline
        \emph{Choose precursor sets with more aromatic rings.} \\
    \bottomrule
  \end{tabular}
\end{table*}

\subsection{Guidance-data matrix}
\label{app:guidance-data-matrix}
\label{app:rdkit5-data-matrix}
\label{app:structure-data-matrix}
\label{app:metric4-data-matrix}

Table~\ref{tab:guidance-data-matrix} combines the three supervision suites.
USPTO 8M and 65M models share each support-specific semantic view; their
frozen route-semantic embedding widths differ. We sample the Structure
validation and test cohorts once within their registered splits and reuse
them across capacities.

After support restriction and graph filtering, RDKit5 and Metric4 retain
only guidance groups containing both preferred and avoided candidates.
Group counts therefore need not increase monotonically with Top$K$.

\begin{table*}[!htbp]
  \caption{Guidance-data sizes for the reported models. Numeric cells are
  semantic groups / graph-bound route keys or rows (G/R). Each USPTO row
  applies to both 8M and 65M; the Pistachio row applies to 65M only.}
  \label{tab:guidance-data-matrix}
  \label{tab:rdkit5-data-matrix}
  \label{tab:structure-data-matrix}
  \label{tab:metric4-data-matrix}
  \centering
  \begin{tabular}{llrrr}
    \toprule
    Dataset & Support & Train G/R & Validation G/R & Test G/R \\
    \midrule
    \multicolumn{5}{l}{\textbf{RDKit5}} \\
    USPTO & USPTO-MR
      & 60,404 / 41,488 & 20,610 / 14,740 & 25,598 / 20,882 \\
    USPTO & Top15
      & 82,912 / 261,905 & 26,644 / 84,436 & 32,370 / 103,112 \\
    USPTO & Top30
      & 130,880 / 528,659 & 42,116 / 170,286 & 51,522 / 208,613 \\
    USPTO & Top45
      & 99,548 / 768,621 & 31,970 / 247,495 & 39,110 / 303,070 \\
    USPTO & Top60
      & 134,226 / 1,046,897 & 43,100 / 336,212 & 52,564 / 410,175 \\
    Pistachio & Top30
      & 180,460 / 1,034,142 & 1,820 / 10,994 & 13,542 / 14,738 \\
    \midrule
    \multicolumn{5}{l}{\textbf{Structure}} \\
    USPTO & Top30
      & 54,012 / 537,777 & 6,752 / 65,473 & 6,752 / 64,422 \\
    \midrule
    \multicolumn{5}{l}{\textbf{Metric4}} \\
    USPTO & USPTO-MR
      & 90,854 / 41,488 & 29,918 / 14,740 & 36,964 / 20,882 \\
    USPTO & Top15
      & 121,628 / 274,018 & 39,084 / 88,142 & 47,850 / 108,148 \\
    USPTO & Top30
      & 126,094 / 537,777 & 40,550 / 172,969 & 49,738 / 212,231 \\
    USPTO & Top45
      & 131,156 / 793,099 & 42,262 / 254,984 & 51,668 / 312,654 \\
    USPTO & Top60
      & 142,810 / 1,065,712 & 45,784 / 341,862 & 55,956 / 417,568 \\
    \bottomrule
  \end{tabular}
\end{table*}

For RDKit5 USPTO Top15 and Top45, \emph{R} counts retained subset-specific
route keys after relabeling; for Pistachio, it counts referenced unique or
endpoint routes in the exact product-condition view. These are graph-bound
training identities, not candidate occurrences multiplied by semantic groups.
Metric4 counts come from the support-specific execution manifests, including
the complete frozen Top60 view. Appendix~\ref{app:metric4-generalization}
summarizes all ten settings on the shared strict-Top100 cohort.

\subsection{Frozen evaluation cohorts}
\label{app:frozen-evaluation-cohorts}

%
Complete execution views and formal rollout cohorts are distinct.
Structure uses 6,752 validation groups from 2,262 cases and 6,752 test
groups from 2,229 cases, with a variable number of groups per structural
axis. We select the checkpoint and residual scale on validation and keep
them fixed for testing. Appendix~\ref{app:test-protocol} specifies the
sampling protocol. We describe the RDKit5 Common Top100 cohort below;
Appendix~\ref{app:metric4-generalization} describes the Metric4 cohort.

\paragraph{RDKit5 strict-Top100 common-support scaling cohort.}
The primary support-scaling experiment uses one shared USPTO Top100 test
universe across all training supports. The strict-100 filter precedes family
sampling; Table~\ref{tab:top100-common-support-cohort} distinguishes the raw
candidate inventory from the routes that survive graph binding.
Each of the five axes contributes 256 paired families and 512 signed groups.
All selected cases have exactly 100 raw candidates, and the frozen RC Top-5
table resolves every selected case. Thus ``strict-100'' fixes the raw
inventory size even when fewer routes survive graph binding. Generation
accounting appears in Appendix~\ref{app:common-top100-protocol}.

\begin{table*}[!htbp]
  \caption{Size accounting for the strict-Top100 common-support test cohort.
  We recompute every column on the named case universe: raw routes count the
  frozen candidate inventory, graph-resolved route keys count routes retained
  by the frozen graph cache, and paired families and signed groups count the
  corresponding five-axis semantic records.}
  \label{tab:top100-common-support-cohort}
  \centering
  \begin{tabular}{@{}>{\raggedright\arraybackslash}p{0.30\linewidth}rrrrr@{}}
    \toprule
    Case universe / view & Cases & \shortstack{Raw route\\records} & \shortstack{Graph-resolved\\route keys}
      & \shortstack{Paired\\families} & \shortstack{Signed\\groups} \\
    \midrule
    Up-to-100 source test universe & 7,342 & 733,017 & 666,086
      & 29,320 & 58,640 \\
    Five-axis graph-bound execution view & 7,269 & 725,812 & 659,056
      & 29,314 & 58,628 \\
    Exact-100 eligible universe & 7,320 & 732,000 & 665,146
      & 29,262 & 58,524 \\
    Strict-Top100 formal cohort & 1,201 & 120,100 & 107,801
      & 1,280 & 2,560 \\
    \bottomrule
  \end{tabular}
\end{table*}


\subsection{Base-generator checkpoint provenance}
\label{app:base-generator-selection}

For the USPTO Structure experiments
(Appendix~\ref{app:structure-embedding-ablation}), we select the 8M and 65M
Base checkpoints by validation deduplicated Top-3 accuracy, at steps
73,728 and 88,105, respectively. Checkpoint selection does not use guidance
test outputs.


\subsection{Sampling and statistical protocol}
\label{app:test-protocol}

\paragraph{Sampling and returned lists.}
The primary and native-cohort three-arm evaluations use ordered RC Top-5
proposals, 100 raw trajectories per group and physical arm, and 50 flow steps.
Matched and Shuffled share reaction-center proposals, root choices, and
trajectory seeds. We canonicalize and deduplicate valid outputs, then rank
them by frequency, with canonical precursor SMILES breaking ties. Invalid
draws remain in the sampling budget. We do not backfill draws or pad short
returned lists. Metric Null is the $\alpha=0$ view of Reference and adds no
physical generation.  The separately named selector and chemistry diagnostics
specify their own readouts.

\paragraph{Shuffled instructions.}
For RDKit5 and Metric4, Shuffled samples a different axis within the same
suite and split, followed by an available polarity and text pool, each
uniformly, and uses canonical wording. These mappings (seed~1042) remain
fixed across model settings on the same cohort. Structure selects a donor
deterministically from a different product and semantic axis within the
same split. Neither policy requires matching polarity. Evaluation retains
the original instruction's preferred and avoided sets.

%
\paragraph{Supplementary cohorts.}
Appendix~\ref{app:frozen-evaluation-cohorts} specifies the Structure
validation and test cohorts, which have a variable number of groups per
structural axis. We sample them without replacement before inference
and keep them fixed across model scale, checkpoint, and residual strength.
Appendix~\ref{app:metric4-generalization} specifies the Metric4 cohort
and its case-balanced evaluation.
The RDKit5 initialization, loss, and proposal--selection studies specify
their auxiliary cohorts in
Appendices~\ref{app:stagea-initialization-ablation},
\ref{app:stageb-loss-ablation}, and~\ref{app:budgeted-proposal-selection}.

\paragraph{Aggregation and uncertainty.}
RDKit5 Common Top100 rates, mean counts, and paired effects use equal-axis,
case-balanced aggregation: we average signed groups within each product--axis
pair, products within each axis, and then the five axes equally. Pooled
recovery counts distinct product--candidate pairs. For Structure, absolute
metrics average groups,
whereas paired effects average differences between arms within each case
and then average cases. Confidence intervals use 10,000 shared case-cluster
bootstrap resamples; Structure uses bootstrap seed 42.
Metric4 retains the case-balanced aggregation specified in
Appendix~\ref{app:axis4-top100-complete}.  Bootstrap uncertainty describes
variation across cases, separately from Stage-B training-seed variation.
\section{Architecture, objectives, and implementation}
\label{app:architecture-implementation}
\label{app:framework-architecture}


%
%

This section specifies the frozen inputs, Stage-A scoring, and Stage-B
adapter and loss definitions. For each product--instruction group $g$,
$\mathcal C_g$ denotes the candidate pool after filtering and deduplication,
retaining its construction order. Appendix~\ref{app:implementation} lists
the training and inference settings.

\subsection{Frozen Stage-A representations}
\label{app:stage-a-inputs}

\paragraph{Encoders.}
Frozen Qwen3-Embedding-8B \citep{zhang2025qwen3embedding} supplies
4,096-dimensional instruction-token features. Reaction-pair features come
from pretrained reaction-pair encoders (Top30 or Top60; 256-D) or the frozen
GraphDiT Base model trained on the corresponding support (Native, 256-D or
512-D). The Top30 reaction-pair encoder was pretrained with Structure
supervision; the Top60 version uses a separate pretrained checkpoint.
We obtain Native features by pooling the complete product--precursor pair
at $t=0$ without instructions. We L2-normalize route embeddings and keep
the language and route encoders frozen during projector fitting.
Table~\ref{tab:route-encoder-lineage} maps experiments
to representations.

\begin{table*}[!htbp]
  \caption{Frozen route encoders in the reported experiments.
  Native features use the Base model trained on the corresponding support.}
  \label{tab:route-encoder-lineage}
  \centering
  \begin{tabular*}{\linewidth}{@{}>{\raggedright\arraybackslash}p{.49\linewidth}@{\extracolsep{\fill}}>{\raggedright\arraybackslash}p{.34\linewidth}l@{}}
    \toprule
    Guided experiments & Route encoder & Dimension \\
    \midrule
    USPTO 8M Top30: RDKit5, Metric4, and Structure
      & Pretrained reaction-pair encoder (Top30) & 256 \\
    USPTO 8M Top15/Top45/Top60: RDKit5 and Metric4
      & Pretrained reaction-pair encoder (Top60) & 256 \\
    USPTO-MR 8M: RDKit5 and Metric4 & Native & 256 \\
    USPTO 65M: all RDKit5/Metric4 supports and Top30 Structure
      & Native & 512 \\
    Pistachio 65M Top30: RDKit5 & Native & 512 \\
    \bottomrule
  \end{tabular*}
\end{table*}

\paragraph{Comparison and compute scope.}
The pretrained reaction-pair encoders use descriptor statistics fitted on
their respective Top30 or Top60 training routes; the 8M Top15/Top45 runs
reuse the Top60 encoder and its statistics.
Guided scaling therefore varies route representations as well as candidate
support, while Reference coverage is independent of these embeddings.
\emph{Generator-free} describes projector fitting on cached features: native
cache construction runs the frozen GraphDiT backbone, and the reported
Stage-A budget excludes cache construction and historical semantic-teacher
pretraining.

\subsection{Structure annotation and quality checks}
\label{app:structure-annotation-recipe}

\paragraph{Annotation.}
The USPTO Structure annotator uses the recorded API alias
\texttt{deepseek-v4-pro} (temperature 0.2; we did not record an immutable
model revision). Given target and candidate precursor SMILES with RDKit evidence,
it proposes up to five relative structural \emph{Prefer}/\emph{Avoid}
instructions and labels candidates positive, negative, neutral, or excluded.
Instructions omit candidate IDs, SMILES, and numeric descriptor values.
We request rewards of $+1,-1,0,0$, respectively. To form the listwise target,
we apply a temperature-one softmax to returned rewards over non-excluded
candidates, then renormalize over the positive and negative candidates
that remain after filtering. We supply the saved annotation system prompt
as supplementary item~P1
(\path{prompts/uspto_structure_v2.txt}).

\paragraph{Quality checks.}
Automated checks validate candidate coverage and label--reward consistency,
and reject malformed annotations and detected copying of candidate literals.
Mapping and graph-construction checks remove inconsistent or unrepresentable
candidates; surviving groups must retain both positive and negative candidates.
These checks establish executable preference contrasts. Chemical label
correctness and distinctness of instruction axes remain LLM judgments without
exhaustive independent expert adjudication.

\subsection{Stage-A grounding and exact route scoring}
\label{app:stage-a-details}
\label{app:stage-a-prior}
Let $H_c=E_T(c)$ denote the instruction-token features,
$Z_c=P_\phi(H_c)$ their projected tokens, and
$v_i=E_R(x,r_i)$ the normalized reaction-pair embedding.
Stage~A averages the projected token features and normalizes
the result to obtain the instruction representation $z_c$:
\begin{equation}
  \begin{aligned}
    z_c
    &=\operatorname{Norm}\!\left(
      \operatorname{MaskedMean}(Z_c)\right),\\
    \widehat q_{\phi,i}^c
    &=\frac{b_{g,i}\exp(\tau z_c^\top v_i)}
      {\sum_j b_{g,j}\exp(\tau z_c^\top v_j)}.
  \end{aligned}
  \label{eq:grounding-score}
\end{equation}
Here $\tau$ scales the similarity logits, MaskedMean averages over
non-padding instruction tokens, and Norm denotes L2 normalization.
The instruction-independent candidate weights come from Structure's
heuristic selection scores (Appendix~\ref{app:candidate-ranking}) or
RDKit5's recorded route counts (one if absent). We normalize these weights
over retained candidates, using uniform weights when all are zero, then
floor each normalized value at $10^{-8}$ to obtain $b_{g,i}$.

The projector output and route embeddings have width 256 at 8M and 512 at
65M.  Using the embeddings and listwise distribution in
Eq.~(\ref{eq:grounding-score}), the exact pairwise term compares raw
similarity logits rather than prior-adjusted logits:
\begin{equation}
  \begin{aligned}
    u_i(c)&=\tau z_c^\top v_i,\\
    \mathcal L_{\mathrm{pair}}
      &=\frac{1}{|\mathcal P_c||\mathcal N_c|}
        \sum_{i\in\mathcal P_c}\sum_{j\in\mathcal N_c}
        [m_{\mathrm{pair}}-u_i(c)+u_j(c)]_+ .
  \end{aligned}
  \label{eq:method-stagea-score-compact}
\end{equation}
The denominator averages all preferred--avoided pairs within a group.
Transfer copies only the validation-selected projector parameters and resets
optimizer state and training counters; the encoders remain frozen.

\subsection{Stage-B residual adapter path}
\label{app:stage-b-details}

\begin{figure*}[!htbp]
  \centering
  \includegraphics[width=\textwidth]{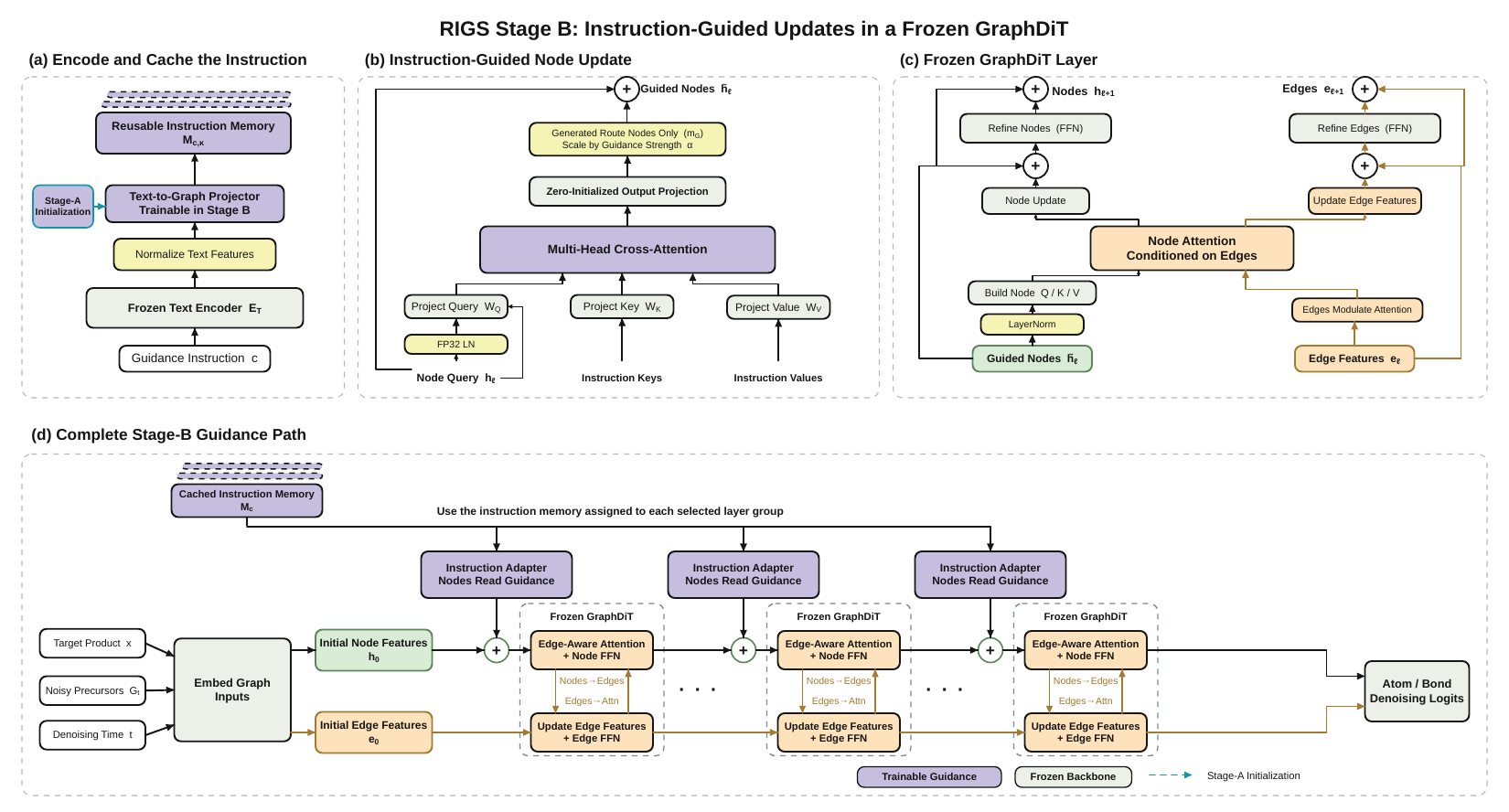}
  \caption{\textbf{Stage-B components and assembly.}
  (a) A frozen language encoder and a trainable Stage-A-initialized projector
  construct reusable instruction-token memory.  (b) Node queries attend to
  that memory through a trainable adapter whose output projection starts at
  zero; masking, residual scaling, and the node-state skip are explicit.
  (c) The frozen GraphDiT block propagates
  the guided node update through its
  existing node--edge coupling. (d) We reuse the same components at
  selected pre-block injection sites. The adapters inject guidance directly
  only into node states; edge states respond inside the frozen backbone.}
  \label{fig:appendix-stage-b-components}
\end{figure*}

Let $\mathcal I$ index the adapted layers and $\mathcal G$ index the
projector-sharing groups. The map $\gamma:\mathcal I\to\mathcal G$ assigns
each adapted layer to a group. Each group $\kappa\in\mathcal G$ has a
projector $P_{\kappa,\theta}$ initialized from $P_{\phi^\star}$.
We write the group's instruction-token memory and the layer-wise memory
from Section~\ref{sec:method-stage-b} as
\begin{equation}
  M_{c,\kappa}=P_{\kappa,\theta}(H_c),
  \qquad M_c^{(\ell)}=M_{c,\gamma(\ell)}.
  \label{eq:method-instruction-memory}
\end{equation}
During inference, we cache the collection
$M_c=(M_{c,\kappa})_{\kappa\in\mathcal G}$. Passing $M_c$ in place of $c$ evaluates the same conditional
model with precomputed instruction features.
The full masked adapter update is
\begin{equation}
  \begin{aligned}
  \Delta h_\ell(c)
    &=A_{\ell,\theta}(h_\ell,M_{c,\gamma(\ell)}),\\
  A_{\ell,\theta}(h,M)
    &:=m_G\odot\operatorname{Attn}_{\ell,\theta}
      (\operatorname{LN}(h),M,M),\\
  \widetilde h_\ell&=h_\ell+\alpha\Delta h_\ell(c),
  &h_{\ell+1}&=B_\ell(\widetilde h_\ell).
  \end{aligned}
  \label{eq:method-token-xattn}
\end{equation}
Here $\operatorname{Attn}_{\ell,\theta}$ is multi-head node-to-text attention
and $m_G$ masks generative graph positions. We omit the frozen block's
edge-state, product, and time arguments from the notation.
Using one-based layer indices, the 8M model shares one projector between
layers $\{2,4\}$ and another between $\{6,8\}$. The 65M model additionally
uses groups $\{10,12\}$ and $\{14,16\}$. Each listed layer has its own
cross-attention adapter.
Figure~\ref{fig:appendix-stage-b-components} shows the components and their
pre-block assembly.

\subsection{Stage-B loss components and metric teacher}
\label{app:stage-b-loss-details}

For candidate $i$ in group $g$, let $D^{\mathrm{student}}_{g,i}$ and
$D^{\mathrm{ref}}_{g,i}$ be the guided and reference denoising losses on the
same noisy precursor graph. Define
$d_{g,i}=D^{\mathrm{ref}}_{g,i}-D^{\mathrm{student}}_{g,i}$; positive
$d_{g,i}$ indicates improved denoising under guidance.
The four terms in Eq.~(\ref{eq:method-stageb-core}) are defined below.

\paragraph{Positive DFM ($\mathcal L_{\mathrm{pos}}$).}
$D^{\mathrm{student}}_{g,i}$ is the full categorical flow-matching loss over
all valid node and edge positions.  The positive term keeps only preferred
candidates $i\in\mathcal P_c$, renormalizes the cached teacher probabilities
within that subset, and averages their full losses across eligible groups.  It
is an absolute likelihood anchor: ranking alone could satisfy a relative
comparison by damaging avoided routes without improving preferred ones.

\paragraph{Raw route ranking ($\mathcal L_{\mathrm{rank}}$).}
Within each eligible group, we compare every preferred--avoided pair $(i,j)$
using the uncalibrated gains $d_{g,i}$ and $d_{g,j}$. We apply a one-sided
Huber penalty to the margin violation
$[m_{\mathrm{rank}}-(d_{g,i}-d_{g,j})]_+$, then average the penalty over
pairs and groups without rescaling scores within products.

\paragraph{Rollout Reference KL ($\mathcal L_{\mathrm{roll}}$).}
During training, we periodically replay noisy precursor graphs visited
by the guided sampler through both Student and frozen Reference.
The loss averages
$\operatorname{KL}(p_{\mathrm{ref}}\Vert p_{\mathrm{student}})$ over every
valid node and edge category at those replay states.  Unlike a teacher-forced
penalty, it controls drift on states that the guided model actually reaches.

\paragraph{Full-reference KL guard ($\mathcal L_{\mathrm{guard}}$).}
On shared noisy precursor graphs used for training, the full-state KL includes
every valid node and edge position, including within-group disagreement positions. We average
candidate KLs within each group, using endpoint probabilities as weights, to obtain $K_g$.
Writing $\langle\cdot\rangle_g$ for the normalized, group-weighted training
average, the unthresholded diagnostic and the guard objective are
\begin{equation}
  \mathcal L_{\mathrm{fullKL}}=\langle K_g\rangle_g,
  \qquad
  \mathcal L_{\mathrm{guard}}
    =\bigl\langle[K_g-\delta_{\mathrm{KL}}]_+\bigr\rangle_g.
  \label{eq:method-groupwise-guard}
\end{equation}
We apply the threshold before averaging: groups below the tolerance
contribute zero.
The guard constrains drift on noisy precursor graphs from training
candidates; rollout KL covers those reached by the sampler.

The first two terms implement instruction steering and the latter two preserve
the base generator.  Structure guidance uses this core directly.  For metric
axis $a$, let $x_{g,i}^{(a)}$ be the deterministic route metric and let
$s_g^{(a)}=+1$ request larger values and $s_g^{(a)}=-1$ request smaller values.
Pool z-scoring subtracts the group mean and divides by its population standard
deviation; constant-valued groups are ineligible.  The deterministic teacher
and Student improvement distribution are
\begin{equation}
  \begin{aligned}
  z_{g,i}^{(a)}
    &=\operatorname{PoolZScore}(\{x_{g,j}^{(a)}\}_j)_i,
    &u_{g,i}^{(a)}&=s_g^{(a)}z_{g,i}^{(a)},\\
  q_{g,i}^{(a)}
    &=\operatorname{softmax}_i(u_{g,i}^{(a)}/T_q),
    &p_{g,i}&=\operatorname{softmax}_i(d_{g,i}/T_s).
  \end{aligned}
  \label{eq:method-metric-teacher}
\end{equation}
The metric branch optimizes
\begin{equation}
  \mathcal L_{\mathrm{teach}}=\operatorname{KL}(q_g^{(a)}\Vert p_g),
  \qquad
  \mathcal L_{B,\mathrm{metric}}
  =\mathcal L_{B,\mathrm{core}}
   +\lambda_{\mathrm{teach}}\mathcal L_{\mathrm{teach}}.
  \label{eq:method-stageb-metric}
\end{equation}
$T_q$ is the teacher temperature and $T_s$ is the Student-score temperature;
larger values flatten the corresponding distribution.  The teacher is an
offline deterministic transformation of route metrics, not another neural
network, and is absent at inference.

\subsection{Notation}
\label{app:rigs-notation}

Table~\ref{tab:rigs-notation} summarizes the notation used in the method and
Algorithms~\ref{alg:rigs-stage-a} and~\ref{alg:rigs-stage-b}.
Within a fixed semantic group, we suppress the group subscript in
$x,r_i,c,q^c,\mathcal P_c$, and $\mathcal N_c$; superscript $\star$ denotes a
validation-selected parameter set. In tables and plots, $\alpha$ abbreviates
$\alpha_{\mathrm{eval}}$. Attention diagrams retain the standard $Q/K/V$
meanings; $\kappa$ denotes a projector-sharing group.
Support size $K$, raw-draw budget $B$, and returned-list cutoff $k$ are
distinct. Contrast order and sign conventions follow the captions.

\begin{table}[!htbp]
  \caption{Recurring notation across support construction, two-stage training,
  and evaluation.}
  \label{tab:rigs-notation}
  \centering
  \begin{tabular}{
    >{\raggedright\arraybackslash}p{0.30\linewidth}
    >{\raggedright\arraybackslash}p{0.64\linewidth}}
    \toprule
    Symbol & Definition \\
    \midrule
    \multicolumn{2}{@{}l}{\textbf{Support and supervision}} \\
    $x,\ r_i$ & Product graph and candidate precursor-set graph. \\
    $g=(x,c,\mathcal C_g,\mathcal Y_g)$
      & Product, instruction, indexed pool, and supervision record. \\
    $\mathcal D_{\mathrm{tr}},\mathcal D_{\mathrm{val}}$
      & Labeled groups for training and validation. \\
    $K,\ K_x,\ \mathcal A_K(x)$
      & Nominal support size, realized size, and retained ordered
        template-expansion prefix. \\
    $\mathcal P_c,\ \mathcal N_c,\ q^c$
      & Preferred/avoided index sets; listwise target distribution. \\
    \midrule
    \multicolumn{2}{@{}l}{\textbf{Stage A: grounding}} \\
    $E_T,H_c;\ E_R,v_i$
      & Frozen text/route encoders; token/route features. \\
    $\phi,\phi^\star,P_\phi$
      & Projector parameters, selected parameters, and projector. \\
    $Z_c,z_c,\widehat q_\phi^c,b_{g,i},\tau$
      & Projected tokens, pooled embedding, predicted pool distribution,
        endpoint prior, and similarity-logit scale. \\
    $\mathcal L_A,\mathcal L_{\mathrm{pair}};
      \lambda_{\mathrm{list}},\lambda_{\mathrm{pair}}$
      & Stage-A loss, pairwise term, and listwise/pairwise weights. \\
    \midrule
    \multicolumn{2}{@{}l}{\textbf{Stage B: residual steering}} \\
    $\theta,\theta^\star;\ P_{\kappa,\theta},A_{\ell,\theta}$
      & Trainable/selected parameters; projector; node-to-text adapter. \\
    $\mathcal I,\mathcal G,\kappa,\gamma;\ M_{c,\kappa}$
      & Injection layers; projector groups, index and assignment;
        projected token memory. \\
    $M_c^{(\ell)},M_c$
      & Per-layer and grouped memories (Eq.~\ref{eq:method-instruction-memory}). \\
    $h_\ell,B_\ell;\ \alpha_{\mathrm{train}},\alpha_{\mathrm{eval}}$
      & Node state, frozen block, and train/eval residual scales. \\
    $d_{g,i}=D^{\mathrm{ref}}_{g,i}-D^{\mathrm{student}}_{g,i}$
      & Guidance-induced reduction in complete-route denoising cost. \\
    $\mathcal L_{\mathrm{pos}},\mathcal L_{\mathrm{rank}},
      \mathcal L_{\mathrm{roll}},\mathcal L_{\mathrm{guard}}$
      & Preferred-endpoint, ranking, rollout-KL and reference-guard
        terms of $\mathcal L_{B,\mathrm{core}}$. \\
    $\mathcal L_{\mathrm{teach}},\mathcal L_{B,\mathrm{metric}}$
      & Metric-teacher KL and metric-augmented Stage-B loss. \\
    \midrule
    \multicolumn{2}{@{}l}{\textbf{Sampling and evaluation}} \\
    $B,S;\ \rho,G_t$
      & Raw draw budget, flow steps, RC proposal and graph at $t$. \\
    $k;\ \mathrm P@k,\mathrm N@k$
      & Returned-list cutoff and preferred/avoided set hit rates. \\
    $\Delta,\Delta\Delta$
      & Named contrast; difference of contrasts. \\
    \bottomrule
  \end{tabular}
\end{table}


\subsection{Implementation and reproducibility details}
\label{app:implementation}

\paragraph{Objective coefficients.}
In the notation of Eq.~(\ref{eq:method-stagea-objective}), Stage~A uses
$\lambda_{\mathrm{list}}=1.0$, $\lambda_{\mathrm{pair}}=0.5$, pair margin
$m_{\mathrm{pair}}=0.1$, and fixed logit scale $\tau=10$.
Stage~B uses
$\lambda_{\mathrm{pos}}=0.10$, $\lambda_{\mathrm{rank}}=0.05$,
$\lambda_{\mathrm{roll}}=1.00$, $\lambda_{\mathrm{guard}}=0.05$, and
$\delta_{\mathrm{KL}}=0.02$.  Structure guidance uses this core objective
  directly, whereas the metric branches set $\lambda_{\mathrm{teach}}=0.05$ in
  Eq.~(\ref{eq:method-stageb-metric}) and use $T_q=T_s=1$ in
  Eq.~(\ref{eq:method-metric-teacher}). We set candidate-score calibration to
\texttt{none}; Appendix~\ref{app:stage-b-loss-details} specifies the exact
score and teacher reductions.

\paragraph{Training settings.}
Stage~A and Stage~B use learning rates $10^{-4}$ and $3\times10^{-5}$,
respectively. For RDKit5 Common Top100, both model sizes train for 30/14
epochs (Stage~A/Stage~B), with effective batches of 512 semantic groups for
USPTO-MR and 32 for synthetic supports. Training uses BF16 mixed precision
and seed 42 (43 and 44 for Stage-B robustness repeats), with
$\alpha_{\mathrm{train}}=1$ fixed throughout Stage~B.
Table~\ref{tab:training-coordinates} summarizes Structure training and
selected checkpoints; RDKit5 Common Top100 and Metric4
evaluation settings appear in Tables~\ref{tab:common-top100-coordinates}
and~\ref{tab:axis4-top100-coordinates}.

\begin{table*}[!htbp]
  \caption{Structure Top30 training and validation-frozen checkpoints.
  Training epochs are Stage~A/Stage~B; the selected Stage-A epoch initializes
  Stage~B, whose test setting is $(e,\alpha_{\mathrm{eval}})$.
  Both capacities use Matched-P@1-first selection.}
  \label{tab:training-coordinates}
  \centering
  \begin{tabular}{lccc}
    \toprule
    Suite / scale & \shortstack{Training epochs\\Stage A / B}
      & \shortstack{Stage-A\\selected epoch}
      & \shortstack{Stage-B test\\$(e,\alpha_{\mathrm{eval}})$} \\
    \midrule
    Structure / 8M  & 30 / 30 & 28 & $(22,.50)$ \\
    Structure / 65M & 20 / 30 & 20 & $(26,1)$ \\
    \bottomrule
  \end{tabular}
\end{table*}

\paragraph{Inference implementation.}
Inference scales only the cross-attention output residual by
$\alpha_{\mathrm{eval}}$, not backbone states, Reference logits, or loss
weights. Appendix~\ref{app:test-protocol} specifies sampling, readout, and
uncertainty.

\paragraph{CTMC sampling.}
\label{app:ctmc-sampling}
We retain the Base proposal schedule, graph initialization, and CTMC sampler
\citep{wang2026retrosynthesis}. Given product $x$, instruction $c$, and
reaction-center proposal $\rho$, the guided model $f_{\theta,\alpha}$
predicts a factorized distribution $\widehat p_t$ over terminal precursor
graphs from the current graph $G_t$ at time $t\in[0,1]$.
The guided transition rates are
\begin{equation}
  \widehat R_t(G_t,G')
  =\mathbb E_{\widetilde G_1\sim\widehat p_t}
    \left[R_t(G_t,G'\mid G_0,\widetilde G_1)\right],
  \label{eq:method-guided-rate}
\end{equation}
where $R_t$ is the Base sampler's fixed conditional rate and
$\widetilde G_1$ is a possible terminal graph marginalized under the model
prediction, not a ground-truth input. The unchanged first-order update is
\begin{equation}
  \Pr(G_{t+\Delta t}=G'\mid G_t,x,\rho,c)
  \approx \mathbf{1}\{G'=G_t\}
    +\Delta t\,\widehat R_t(G_t,G').
  \label{eq:method-guided-sampling}
\end{equation}
Here $\Delta t=1/S$, with $S=50$ steps in the reported evaluations.

\FloatBarrier
\section{Limitations and Future Work}
\label{sec:experiment-limitations}

Our evaluation focuses primarily on single-step retrosynthesis. The multistep
case study illustrates how RIGS can guide a planner, but we have not
systematically evaluated its benefits for multistep planning. Future work will
integrate RIGS into LLM-guided planning and explore post-training to improve
output validity and instruction following.

\end{document}